\documentclass[journal, twoside]{IEEEtran}

\usepackage{decar-common}
\usepackage{decar-dynamics}
\usepackage{decar-lie}
\usepackage{decar-post}

\newcommand{\subparagraph}{}
\usepackage{titlesec}
\titlespacing{\subsubsection}{0pt}{6pt}{0pt}

\usepackage{booktabs}

\usepackage{xfrac}

\newlength{\subfigheight}
\newsavebox{\subfigbox}

\DeclareMathOperator{\blkdiag}{blkdiag}
\DeclareMathOperator{\Adjop}{Ad}

\makeatletter
\AtBeginDocument{
  \check@mathfonts
}
\makeatother

\usepackage{tabularx}   

\Crefformat{figure}{#2Fig.~#1#3}
\Crefmultiformat{figure}{Fig.~#2#1#3}{ and~#2#1#3}{, #2#1#3}{, and~#2#1#3}

\definecolor{dark_grey}{rgb}{0.45, 0.45, 0.45}
\definecolor{light_grey}{rgb}{0.75, 0.75, 0.75}

\newcommand{\tabentry}[3]{\num{#1} $\cdot$ \colour{dark_grey}{\num{#2}} $\cdot$ \colour{light_grey}{\num{#3}}}

\DeclareMathOperator{\meter}{m}

\newcommand{\update}[1]{\colour{black}{#1}}
\newcommand{\secondupdate}[1]{\colour{black}{#1}}

\begin{document}

%
%
%
%
%
%
%
\def \myJournal {IEEE Transactions on Robotics}
\def \myDoi {10.1109/TRO.2022.3229842}
\def \myPaperSiteName {IEEE Xplore}
\def \myPaperSiteLink {https://ieeexplore.ieee.org/document/10026622}
\def \myYear {2023}
\def \myPaperCitation{T. Hitchcox and J. R. Forbes, ``Improving Self-Consistency
in Underwater Mapping Through Laser-Based Loop Closure,'' \textit{IEEE
Transactions on Robotics}, vol. 39, no. 3, pp.  1873-1892, 2023.}


\begin{figure*}[t]

\thispagestyle{empty}
\begin{center}
\begin{minipage}{6in}
\centering
This paper has been accepted for publication in \emph{\myJournal}. 
\vspace{1em}

This is the author's version of an article that has, or will be, published in this journal or conference. Changes were, or will be, made to this version by the publisher prior to publication.
\vspace{2em}

\begin{tabular}{rl}
DOI: & \myDoi\\
\myPaperSiteName: & \texttt{\myPaperSiteLink}
\end{tabular}

\vspace{2em}
Please cite this paper as:

\myPaperCitation

\vspace{15cm}
\copyright \myYear \hspace{4pt}IEEE. Personal use of this material is permitted. Permission from IEEE must be obtained for all other uses, in any current or future media, including reprinting/republishing this material for advertising or promotional purposes, creating new collective works, for resale or redistribution to servers or lists, or reuse of any copyrighted component of this work in other works.

\end{minipage}
\end{center}
\end{figure*}
\newpage
\clearpage
\pagenumbering{arabic} 


\fontdimen16\textfont2=\fontdimen17\textfont2
\fontdimen13\textfont2=5pt

\title{Improving Self-Consistency in Underwater Mapping Through Laser-Based Loop Closure (Extended)}

\author{Thomas~Hitchcox,~\IEEEmembership{Graduate Student Member,~IEEE,}
        and~James~Richard~Forbes,~\IEEEmembership{Member,~IEEE}%

\thanks{Manuscript received 30 May 2022; revised 21 September 2022; accepted 22
November, 2022.  This work was supported in part by the Natural Sciences and
Engineering Research Council of Canada and in part by Voyis Imaging Inc. through
the Collaborative Research and Development program.  The work of Thomas Hitchcox
was supported by the McGill Engineering Doctoral Award program.  This paper was
recommended for publication by Associate Editor Maurice Fallon and Editor
Francois Chaumette upon evaluation of the reviewers' comments.}

\thanks{T.~Hitchcox (corresponding author) and J.~R.~Forbes are with the
Department of Mechanical Engineering, McGill University, Montreal, QC H3A~0C3,
Canada.  \texttt{thomas.hitchcox@mail.mcgill.ca,
james.richard.forbes@mcgill.ca}.}}%

\markboth{IEEE Transactions on Robotics.  Preprint Version.  Accepted November,
2022}{Hitchcox and Forbes: Improving Self-Consistency in Underwater Mapping
through Laser-Based Loop Closure (Extended)}%

\maketitle

\begin{abstract}
    Accurate, self-consistent bathymetric maps are needed to monitor changes in
    subsea environments and infrastructure. These maps are increasingly
    collected by underwater vehicles, and mapping requires an accurate vehicle
    navigation solution. Commercial off-the-shelf (COTS) navigation solutions
    for underwater vehicles often rely on external acoustic sensors for
    localization, however \secondupdate{survey-grade acoustic} sensors are
    expensive to deploy and limit the range of the vehicle. Techniques from the
    field of simultaneous localization and mapping, particularly loop closures,
    can improve the quality of the navigation solution over dead-reckoning, but
    are difficult to integrate into COTS navigation systems. This work presents
    a method to improve the self-consistency of bathymetric maps by smoothly
    integrating loop-closure measurements into the state estimate produced by a
    commercial subsea navigation system. Integration is done using a
    white-noise-on-acceleration motion prior, without access to raw sensor
    measurements or proprietary models. Improvements in map self-consistency are
    shown for both simulated and experimental datasets, including a 3D scan of
    an underwater shipwreck in Wiarton, Ontario, Canada.
\end{abstract}


\section{Introduction}
\label{sec:intro}

\IEEEPARstart{A}{ccurate}, self-consistent bathymetric maps are critical for
assessing the health of subsea environments and infrastructure.  Increasingly,
these maps are collected by autonomous underwater vehicles (AUVs) using a
variety of on-board sensors, including cameras
\cite{Kim2013,Suresh2019,Rahman2019}, sonar
\cite{Roman2007,Barkby2011a,Ozog2016}, and laser scanners
\cite{Palomer2019,Hitchcox2020}.  Since the map is constructed using the
estimated AUV trajectory, long-term navigation accuracy is a prerequisite for
building accurate maps.

The standard navigation solution for commercial AUVs is a
commercial off-the-shelf (COTS) inertial navigation system (INS), with acoustic
aiding from a Doppler velocity log (DVL).  The dead-reckoned precision of these
systems is measured by drift rate as a percent of distance traveled, with
high-quality DVL-INS systems achieving a drift rate of as low as
\SI{0.01}{\percent}.  However, without localizing measurements the precision of
the state estimate will deteriorate without bound, impacting long-term accuracy.

Since GPS signals attenuate rapidly in water, AUV localization is primarily done
using acoustics \cite{Paull2014}.  For example, long baseline (LBL) arrays are
acoustic beacons installed on the seafloor that trilaterate the position of an
AUV, much like an ``acoustic GPS.''  \secondupdate{Short baseline (SBL) and}
ultrashort baseline (USBL) systems are affixed to a surface vessel, and measure
the acoustic range and bearing of an underwater vehicle.  These sensors are
frequently deployed in a commercial setting, and have been used to aid AUV
navigation in the literature, for example \cite{Jakuba2008}.

\begin{figure}[t]
	\sbox\subfigbox{%
	  \resizebox{\dimexpr\columnwidth}{!}{%
		\includegraphics[height=4cm]{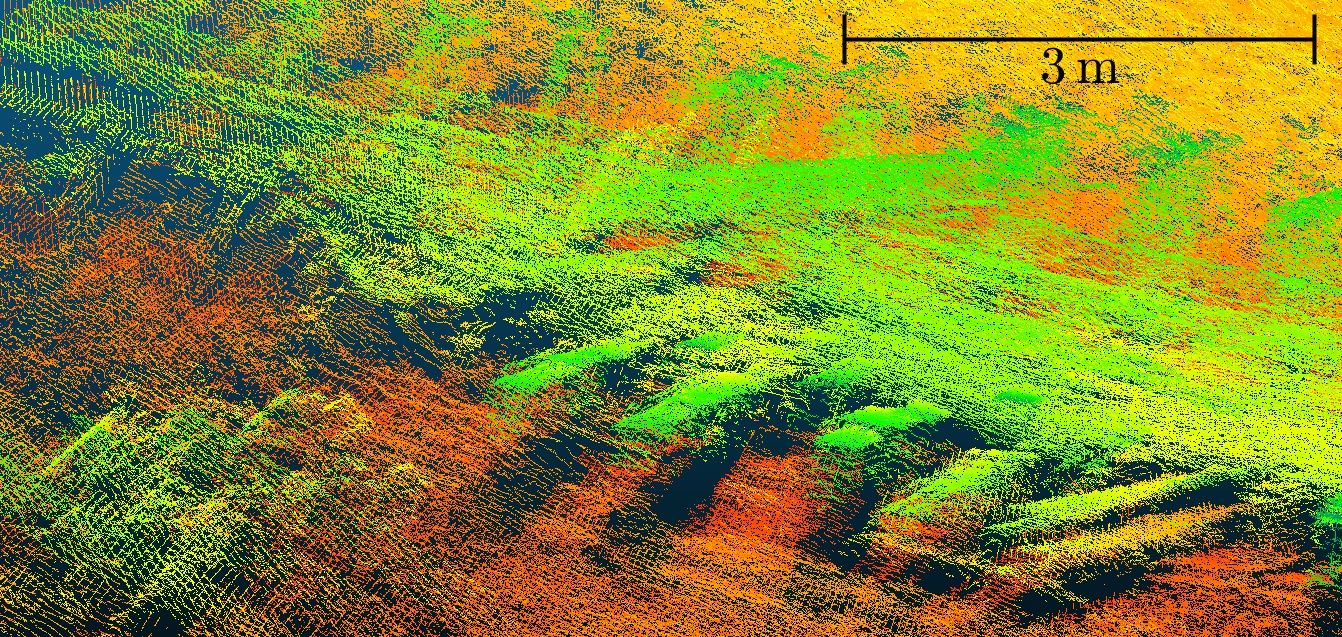}%
	  }%
	}
	\setlength{\subfigheight}{\ht\subfigbox}
	\centering
    \includegraphics[height=\subfigheight]{figs/summary_pic_iso_prior_small.jpg}
    \par \smallskip
    \includegraphics[height=\subfigheight]{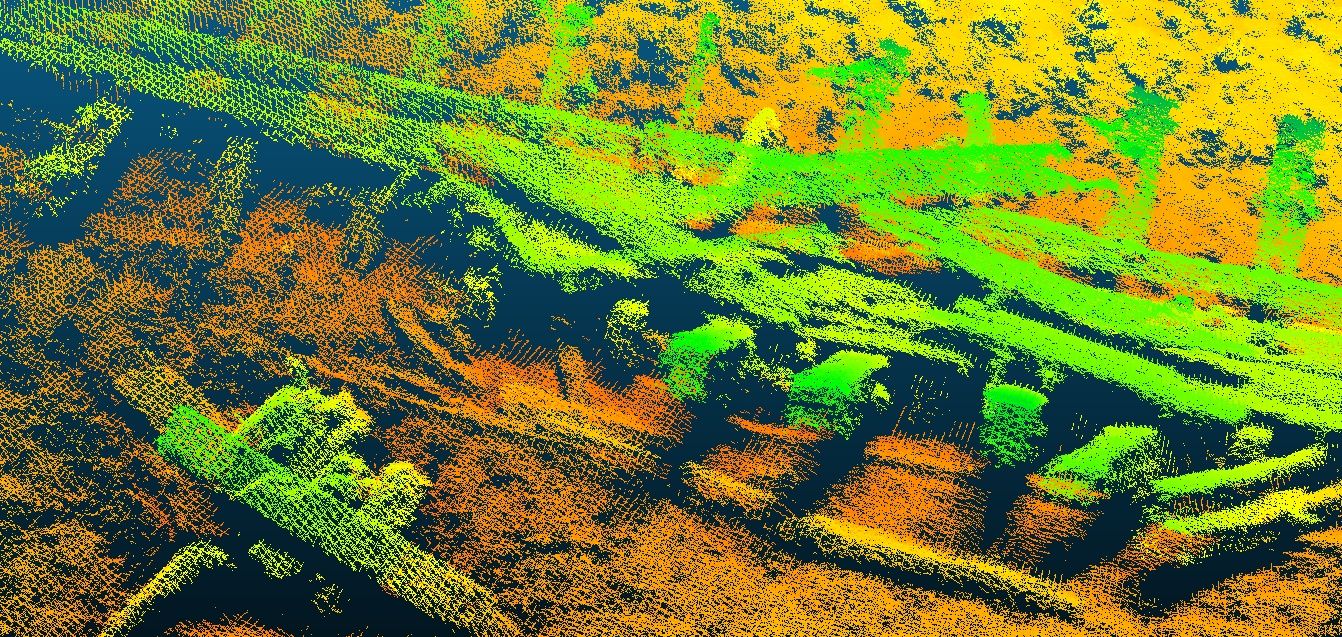}
	\caption{A point cloud map of an actual shipwreck collected in Wiarton,
	Ontario, Canada, \update{where colour represents relative depth.}  The top
	map is generated using the state estimate produced by a commercial
	off-the-shelf (COTS) Doppler velocity log-aided inertial measurement system
	(DVL-INS).  The bottom map is generated using the proposed method, which
	conditions the DVL-INS estimate on loop-closure measurements \textit{without
	access to the raw DVL-INS sensor measurements.} Note the improvement in map
	\textit{self-consistency} when loop-closure measurements are included.}
    \label{fig:summary}
\end{figure}

Acoustic positioning systems enable accurate and precise AUV trajectory
estimates, however they are expensive to deploy and limit the mission domain of
the vehicle.  For example, LBL systems are time-consuming to install and
calibrate, while \secondupdate{SBL and} USBL systems require the presence of a
large surface vessel.  In addition, acoustic positioning systems produce
measurements with limited precision, which may lead to small irregularities in a
composite map built from several overlapping measurements of the same area.
This in turn may make it difficult to assess relative distances and deformation,
or other measurements critical to subsea safety.

\subsection{Motivation}
\label{sec:motivation}

\textit{Loop closures} play a central role in many simultaneous localization and
mapping (SLAM) algorithms, whereby a vehicle returns to and is able to recognize
a previously explored region of the map.  Loop-closure measurements effectively
``reset'' any navigation drift accumulated throughout the loop
\cite{Cadena2016}, resulting in navigation solutions that are both more accurate
and more precise than dead-reckoning, without the need for external localizing
measurements.  Multiple loop closures over time lead to bounded navigation drift
and a more \textit{self-consistent} map estimate, \update{whereby the resulting
map is free of irregularities and ``double vision'' effects produced by poorly
aligned measurements,} \secondupdate{an example of which is shown in
\Cref{fig:summary}.}  This is not to be confused with the term
\textit{consistent}, which in the context of state estimation describes a
solution for which the covariance bounds accurately reflect the error in the
mean state estimate \cite[Sec.~5.4.2]{Bar-Shalom2004}.

\begin{figure}[!t]
	\sbox\subfigbox{%
	  \resizebox{\dimexpr\columnwidth-1em}{!}{%
		\includegraphics[height=4cm]{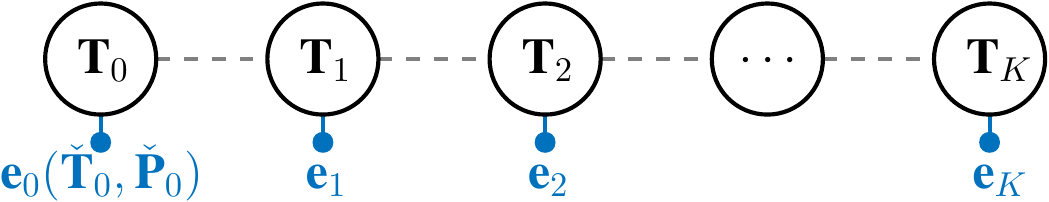}%
	  }%
	}
	\setlength{\subfigheight}{\ht\subfigbox}
	\centering
	\includegraphics[height=\subfigheight]{figs/factor_graph_ins-eps-converted-to.pdf}
	\caption{A factor graph depicting the output from a commercial DVL-INS.
	Only the estimated trajectory $\mbfcheck{T}_{0:K}$ and incomplete marginal
	covariance matrices $\mbfcheck{P}_{0:K}$ are available, leading to the
	formation of prior factors $\mbf{e}_{0:K}$.  Without factors linking
	adjacent nodes, loop-closure corrections cannot propagate throughout the
	graph, and the trajectory cannot be updated.}
	\label{fig:factorgraphins}
\end{figure}

Previous applications of SLAM for underwater mapping leverage loop-closure
measurements to improve map self-consistency.  However, these applications have
largely been implemented on research platforms with access to raw sensor
measurements and full knowledge of the state estimation algorithm.  In contrast,
commercial ``strapdown'' DVL-INS systems for subsea navigation produce a state
\textit{estimate}, and due to their proprietary nature rarely provide access to 
\begin{packed_enum}
    \item \textit{raw sensor measurements}, including interoceptive measurements
	$\mbf{u}_k$, such as from an IMU, and exteroceptive measurements
	$\mbf{y}_\ell$, such as from a DVL;
    \item a \textit{process model} of the form 
	\begin{equation}
		\mbftilde{x}_k = \mbf{f}_{k-1}(\mbf{x}_{k-1}, \mbf{u}_{k-1}), 
	\end{equation}
	which describes how the vehicle moves throughout time;
	\item \textit{sensor models} of the form 
	\begin{equation}
		\mbftilde{y}_\ell = \mbf{g}_\ell(\mbf{x}_\ell, \mbf{v}_\ell),
	\end{equation}
	which allow for predicted measurements; and
	\item \textit{sensor noise and bias specifications}, for example 
    \begin{subequations}
        \begin{align}
			\mbf{u}(t) =& \ \mbfbar{u}(t) + \mbs{\beta}(t) + \mbf{w}(t), \\
			\mbsdot{\beta}(t) \sim& \ \mathcal{N}(\mbf{0}, \mbc{Q}_{\dot{\beta}} \delta (t-t^\prime)), \\
			\mbf{w}(t) \sim& \ \mathcal{N}(\mbf{0}, \mbc{Q}_w \delta(t-t^\prime)),
		\end{align}
    \end{subequations}
	where $\mbf{u}$ is known to be corrupted by time-varying random walk bias
	$\mbs{\beta}$ and Gaussian white noise $\mbf{w}$, characterized by power
	spectral densities $\mbc{Q}_{\dot{\beta}}$ and $\mbc{Q}_w$, respectively. 
\end{packed_enum}
Commercial DVL-INS systems, \update{for example the Sonardyne SPRINT-Nav 500
\cite{sprintnavdatasheet},} are effectively ``black boxes,'' and their lack of
transparency makes it difficult to incorporate loop-closure measurements using
conventional state estimation tools \cite{Kuemmerle2011,Kaess2012}, as
illustrated by the factor graph \cite{Dellaert2017} in
\Cref{fig:factorgraphins}.  

\subsection{Prior Work}
\label{sec:priorwork}

The field of simultaneous localization and mapping has found ample application
in the domain of subsea robotics.  For example, \cite{Roman2007} produced a
self-consistent bathymetric map of two hydrothermal vents by aligning point
cloud submaps generated using multibeam sonar.  A distributed particle mapping
algorithm was described in \cite{Barkby2011a}, where particle weighting was
determined based on the innovation between multibeam sonar measurements and the
existing map.  However, the map resolution was limited by the selection of a
grid cell size.  \update{This limitation was later addressed in
\cite{Barkby2012}, which adopted Gaussian processes as a map representation.}
The submap alignment approach was followed by \cite{Palomer2016}, which
demonstrated improvements in submap simplification and point cloud alignment in
a harbour scanning application.  Harbour scanning and surveillance was also the
subject of \cite{Johannsson2010}, which used a feature-based approach to align
point clouds collected from imaging sonar.  Recent research has focused on more
structured environments, for example ship hull inspection
\cite{Kim2013,Kim2015,Ozog2016,Teixeira2016,Li2018} and subsea infrastructure
\cite{Palomer2019}.  

\update{These studies generally had access to the information enumerated in
\Cref{sec:motivation}, and as a result were able to incorporate loop-closure
measurements using conventional state estimation techniques.  For example,
\cite{Roman2007} applied loop-closure measurements within an extended Kalman
filtering framework, and enjoyed access to raw navigation sensor measurements as
well as a vehicle process model.  Individual particles in \cite{Barkby2011a} and
\cite{Barkby2012} were propagated forward using DVL measurements and a
constant-velocity motion model.  The research platform used in the related ship
hull inspection studies \cite{Kim2013,Kim2015,Ozog2016,Teixeira2016,Li2018}
produced raw DVL, IMU, and depth sensor measurements, \secondupdate{while the
platform in \cite{Rahman2019} had access to a variety of raw sensor measurements
including stereo vision and profiling sonar.}  These applications used
conventional pose-graph SLAM to incorporate loop-closure measurements produced
by various exteroceptive sensors.}

\subsection{Contribution}
\label{sec:contribution}

\update{\secondupdate{This work describes a novel approach to underwater mapping
using a high-resolution laser line scanner and the output of a commercial
DVL-INS navigation system.  First, this work develops a robust laser-based
front-end algorithm to produce high-precision loop-closure measurements by
aligning point cloud scans collected in challenging underwater environments.
Next, this work shows how to cleanly fuse loop-closure measurements into the
output of a survey-grade COTS DVL-INS system via factor graph optimization.  As
these commercial systems are typically ``black boxes'' which only provide a
navigation \textit{estimate}, the proposed approach shows how to systematically
incorporate loop-closure measurements \textit{without access to raw sensor
measurements or other information typically required in state estimation tasks}.
In contrast to previous approaches, the proposed methodology also enforces a
\textit{smoothness} requirement on the posterior trajectory estimate.  This
eliminates discontinuities often encountered in dead-reckoned trajectory
estimates, and is critical for accurate feature detection in laser submaps.  In
summary, the proposed methodology describes a robust and comprehensive system
for high-precision, self-consistent underwater mapping using COTS navigation
systems.  Improvements to both map self-consistency and the relative accuracy of
the trajectory estimate are rigorously evaluated in simulation and on an actual
underwater mapping dataset.  

}
}

\subsection{Paper Organization}
\label{sec:organization}

This paper is organized as follows.  \Cref{sec:preliminaries} contains
preliminary information on conventions used, state estimation on matrix Lie
groups, and batch state estimation.  \Cref{sec:methodology} introduces the
methodology, including the formulation of loop-closure measurements from laser
scan data and the construction of the batch optimization problem.
\Cref{sec:results} contains results on simulated and field datasets.  The paper
concludes in \Cref{sec:conclusion} with a review of the findings and
opportunities for future work.

\section{Preliminaries}
\label{sec:preliminaries}

\subsection{Reference Frames and Navigation Conventions}
\label{sec:rframes}

This section discusses the conventions for attitude and displacement used in
this paper.  A three-dimensional dextral reference frame $\rframe{a}$ is
composed of three orthonormal physical basis vectors.  The position of physical
point $\update{z}$ relative to point $\update{w}$, denoted by
$\ura{r}^{\hspace{-0.2em}\update{zw}}$, is resolved in reference frame
$\rframe{a}$ as ${\mbf{r}_a^{\update{zw}}}$ and in reference frame $\rframe{b}$
as $\mbf{r}_b^{zw}$.  These these quantities are related via ${\mbf{r}^{zw}_a =
\mbf{C}_{ab} \mbf{r}^{zw}_b}$, where $\mbf{C}_{\update{ab}}$ is a direction
cosine matrix, ${\mbf{C} \in SO(3) = \{ \mbf{C} \in \rnums^{3\times3} \, | \,
\mbf{C}\mbf{C}^\trans = \eye, \det \mbf{C} = +1 \}}$ \cite{Barfoot2017}.
\update{Time-varying quantities are indicated by the subscript $(\cdot)_k$, for
example $\mbf{r}^{z_kw}_a$ describes the position of moving point $z$ at time
$t_k$.}  In this work, point $z$ is affixed to the vehicle, while $w$ denotes
the stationary point in the world.  Body frame $\rframe{b}$ rotates with the
vehicle, while local geodetic frame $\rframe{a}$ remains stationary.  Both
$\rframe{b}$ and $\rframe{a}$ are north-east-down (NED), in agreement with
maritime convention.

\subsection{Matrix Lie Groups}
\label{sec:liegroups}

The attitude and position of a vehicle \update{at time $t_k$}, collectively
referred to as the vehicle's ``pose,'' may be conveniently represented in 3D
space as an element of matrix Lie group $SE(3)$ \cite[Sec.~7.1.1]{Barfoot2017},
\begin{equation}
    \mbf{T}^{z_{\update{k}}w}_{ab_{\update{k}}} = \begin{bmatrix}
        \mbf{C}_{ab_{\update{k}}} & \mbf{r}^{z_{\update{k}}w}_a \\ \mbf{0} & 1 
    \end{bmatrix}
    \in SE(3),
    \label{eqn:se3pose}
\end{equation}
with ${SE(3) = \{ \mbf{T} \in \rnums^{4\times 4} \, | \, \mbf{C} \in SO(3),
\mbf{r} \in \rnums^3 \}}$.  Associated with every matrix Lie group is a matrix
Lie algebra, defined as the tangent space at the group identify \cite{Sola2018}.
For $SE(3)$, this is $\mathfrak{se}(3) \triangleq T_\eye SE(3)$.  For estimation
problems involving matrix Lie groups, the matrix Lie algebra is a convenient
space to represent perturbations and uncertainty.  An element of
$\mathfrak{se}(3)$ is given by \cite[Sec.~2.3]{Arsenault2019}
\begin{equation}
    \mbs{\xi}^\wedge = \begin{bmatrix}
        \mbs{\phi} \\ \mbs{\rho}
    \end{bmatrix}^\wedge = 
        \begin{bmatrix}
        0 & -\phi_3 & \phi_2 & \rho_1 \\ 
        \phi_3 & 0 & -\phi_1 & \rho_2 \\ 
        -\phi_2 & \phi_1 & 0 & \rho_3 \\ 
        0 & 0 & 0 & 0
    \end{bmatrix} \in \mathfrak{se}(3),
\end{equation}
where ${(\cdot)^\wedge : \rnums^6 \to \mathfrak{se}(3)}$ is an isometric
operator.  The inverse of this operator is ${(\cdot)^\vee : \mathfrak{se}(3) \to
\rnums^6}$, such that ${(\mbs{\xi}^\wedge)^\vee = \mbs{\xi}}$.  A Lie group and
Lie algebra are related through the exponential map, which for matrix Lie groups
is the matrix exponential, 
\vspace{-1pt}
\begin{equation}
    \mbf{T} = \exp(\mbs{\xi}^\wedge).
    \vspace{-1pt}
\end{equation}
The matrix logarithm is used to return to the Lie algebra via 
\vspace{-1pt}
\begin{equation}
    \mbs{\xi}^\wedge = \log(\mbf{T}).
    \vspace{-1pt}
\end{equation}
Elements of the matrix Lie algebra are combined according to the
Baker-Campbell-Hausdorff (BCH) equation,
\vspace{-1pt}
\begin{equation}
    \mbs{\gamma}^\wedge = \log \left( \exp(\mbs{\xi}^\wedge) \exp(\mbs{\eta}^\wedge) \right).
    \vspace{-1pt}
\end{equation}
An approximation to the BCH equation for ${\mbs{\xi} \gg \mbs{\eta}}$ is 
\vspace{-1pt}
\begin{equation}
    \mbs{\gamma}^\wedge \approx (\mbs{\xi} + \mbf{J}^\textrm{r}(\mbs{\xi})\inv \mbs{\eta})^\wedge,
    \vspace{-1pt}
\end{equation}
where $\mbf{J}^\textrm{r}$ is the right Jacobian of $SE(3)$
\cite[Sec.~7.1.5]{Barfoot2017}.

Errors on matrix Lie groups are defined multiplicatively.  This work uses the
left-invariant error definition,
\vspace{-3pt}
\begin{equation}
    \delta \mbf{T} = \mbf{T}\inv \mbftilde{T},
    \label{eqn:lefterror}
    \vspace{-3pt}
\end{equation}
where $\mbf{T}$ is the current state estimate and $\mbftilde{T}$ is a state
estimate generated from sensor measurements or prior information.  The
corresponding perturbation scheme is
\vspace{-1pt}
\begin{equation}
    \mbf{T} = \mbfbar{T} \exp(-\delta \mbs{\xi}^\wedge),
    \label{eqn:perturbation}
    \vspace{-1pt}
\end{equation}
with perturbation ${\delta \mbs{\xi} \sim \mathcal{N}(\mbf{0}, \mbf{P})}$,
${\mbf{P} = \expect{\delta \mbs{\xi} \, \delta \mbs{\xi}^\trans} \in \rnums^{6
\times 6}}$.  Note the negative sign in \eqref{eqn:perturbation} ensures
consistency with the left-invariant error definition \eqref{eqn:lefterror}.  The
state estimate is therefore defined by mean estimate $\mbfbar{T}$ and covariance
$\mbf{P}$. 

This work makes frequent use of the adjoint matrix $\Adj(\mbf{T})$, which maps
perturbations about the group identity to other group elements \cite{Sola2018}.
Formally, 
\vspace{-3pt}
\begin{equation}
    \Adj(\mbf{T}) \delta \mbs{\xi} \triangleq \left( \mbf{T} \delta \mbs{\xi}^\wedge \mbf{T}\inv \right)^\vee.
    \vspace{-3pt}
\end{equation}
For $SE(3)$, the adjoint matrix is \cite{Arsenault2019}
\begin{equation}
    \Adj(\mbf{T}) = \begin{bmatrix}
        \mbf{C} & \mbf{0} \\ \mbf{r}^\times \mbf{C} & \mbf{C}
    \end{bmatrix},
\end{equation}
where $(\cdot)^\times$ is the skew-symmetric operator
\cite[Sec.~7.1.2]{Barfoot2017}.  The adjoint matrix is represented in the matrix
Lie algebra as 
\begin{equation}
    \left( \adj(\mbs{\xi}^\wedge_1) \mbs{\xi}_2 \right)^\wedge \triangleq \left[ \mbs{\xi}^\wedge_1, \mbs{\xi}^\wedge_2 \right] = \mbs{\xi}^\wedge_1 \mbs{\xi}^\wedge_2 - \mbs{\xi}^\wedge_2 \mbs{\xi}^\wedge_1,
\end{equation}
where $\left[ \cdot, \cdot \right]$ is the Lie bracket
\cite[Sec.~10.2.6]{Chirikjian2011}.  For $\mathfrak{se}(3)$, 
\begin{equation}
    \adj(\mbs{\xi}^\wedge) = \begin{bmatrix}
        \mbs{\phi}^\times & \mbf{0} \\
        \mbs{\rho}^\times & \mbs{\phi}^\times
    \end{bmatrix}.
\end{equation}

\subsection{Gaussian Processes}
\label{sec:gaussianprocesses}

A continuous-time Gaussian process (GP) may be viewed as a distribution over
functions, 
\begin{equation}
    \mbf{f}(t) \sim \mathcal{GP} \left( \mbs{\mu}(t), \mbs{\Sigma}(t, t^\prime) \right),
\end{equation}
where $\mbs{\mu}(t)$ is the mean function, and $\mbs{\Sigma}(t,t^\prime)$ is the
covariance function \cite[Sec.~2.2]{Williams2006}\cite[Sec.~2.3]{Barfoot2017}.
For any finite collection of time steps $t_{0:K}$, $\mbf{f}(t_{0:K})$ follows a
joint Gaussian distribution.  The covariance function determines how individual
function samples $\mbf{f}_i(t)$ covary over time.  For example, a GP for which
the covariance over time is large will be smoother than a GP for which the
covariance over time is small.  This work uses the zero-mean white noise GP,
given by \cite[Sec.~2.3]{Barfoot2017}
\vspace{-6pt}
\begin{equation}
    \mbf{w}(t) \sim \mathcal{GP}(\mbf{0}, \mbc{Q} \delta(t-t^\prime)),
    \label{eqn:whitenoisegp}
\end{equation}
where $\mbc{Q}$ is a power spectral density matrix and $\delta(\cdot)$ is the
Dirac delta function.

\subsection{The White-Noise-On-Acceleration Motion Prior}
\label{sec:wnoa}

The white-noise-on-acceleration (WNOA) motion prior may be summarized by the
following set of stochastic differential equations \cite{Anderson2015},
\vspace{-6pt}
\begin{subequations}
    \begin{align}
        \mbfdot{T}(t) =& \ \mbf{T}(t) \mbs{\varpi}_b(t)^\wedge,
        \label{eqn:ctkin} \\
        \mbsdot{\varpi}_b(t) \sim& \ \mathcal{GP}(\mbf{0}, \mbc{Q} \delta (t - t^\prime)).
        \label{eqn:wnoanoise}
    \end{align}
\end{subequations}
\Cref{eqn:ctkin} describes the continuous-time state kinematics for $SE(3)$,
with ${\mbs{\varpi}_b}$ the \textit{generalized velocity}, such that ${T
\mbs{\varpi}^\wedge_b \in \mathfrak{se}(3) }$, with $T$ a time increment.  The
subscript $(\cdot)_b$ has been included to emphasize that $\mbs{\varpi}$ is a
body-frame quantity.  The time rate of change of $\mbs{\varpi}$ is distributed
according to the zero-mean white noise Gaussian process in
\eqref{eqn:wnoanoise}, with power spectral density $\mbc{Q}$.  Note that
$\mbc{Q}$ is a hyperparameter that needs to be tuned.  This motion prior helps
to enforce smoothness is the posterior state estimate, as
${\expect{\mbsdot{\varpi}} = \mbf{0}}$.  In discrete time, this implies
${\expect{\mbs{\varpi}_k} = \mbs{\varpi}_{k-1}}$.  The white noise assumption
also preserves sparsity in the upcoming batch problem \cite{Barfoot2014a}.  

The WNOA assumption is reasonable in the context of subsea navigation, as AUV
kinematics evolve slowly over time.  With the inclusion of $\mbs{\varpi}$, the
augmented navigation state becomes the ordered pair
\begin{equation}
    \mbf{X} = \left( \mbf{T}, \mbs{\varpi} \right) \in SE(3) \times \rnums^6.
    \label{eqn:navstate}
\end{equation}

\subsection{Batch State Estimation}
\label{sec:batchestimation}

Given a set of exteroceptive measurements ${ \{ \mbf{y}_\ell \}^L_{\ell=1}}$,
interoceptive measurements ${ \{ \mbf{u}_k \}^{K-1}_{k=0}}$, and prior estimate
${\mbf{Y}_0 \! =\! \mbfbar{Y}_0 \exp (-\delta \mbs{\eta}_0^\wedge)}$,
${\mbf{S}_0 \! = \! \expect{\delta \mbs{\eta}_0 \, \delta \mbs{\eta}_0^\trans}}$, the standard
approach to batch estimation is to produce a maximum a posteriori (MAP)
solution, given by 
\begin{equation}
    \mbfhat{X} = \argmax_{\mbf{X}} \, p \! \left( \mbf{X} \, | \, \mbf{y}_{1:L}, \mbf{u}_{0:K-1}, \mbf{Y}_0 \right).
    \label{eqn:mapestimate}
\end{equation}
Under the Markov assumption, the joint probability in \eqref{eqn:mapestimate}
may be factored as 
\begin{equation}
    \mbfhat{X} = \argmax_{\mbf{X}}
    \prod^L_{\ell=1} p \hspace{-0.2em} \left( \mbf{y}_\ell \hspace{0.05em} \big| \hspace{0.05em} \mbf{X}_\ell \right) \hspace{-0.2em} \prod^{K}_{k=1} \hspace{-0.2em} p \hspace{-0.2em} \left( \mbf{X}_k \hspace{0.05em} \big| \hspace{0.05em} \mbf{X}_{k-1}, \mbf{u}_{k-1} \right) p \hspace{-0.2em} \left( \mbf{X}_0 \hspace{0.05em} \big| \hspace{0.05em} \mbf{Y}_0 \right).
    \label{eqn:mapfactored}
\end{equation}
Taking the negative log likelihood of \eqref{eqn:mapfactored} results in a
nonlinear least-squares problem of the form 
\begin{equation}
    \mbfhat{X} = \argmin_{\mbf{X}} J(\mbf{X}),
    \label{eqn:batchcost}
\end{equation}
where the objective function $J(\mbf{X})$ is given by
\begin{align*}
    J(\mbf{X}) =& \ \frac{1}{2} \sum^L_{\ell=1} \norm{ \mbf{e}_\ell (\mbfbar{y}_\ell, \mbf{g}_\ell(\mbfbar{X}_\ell, \mbf{0}))}^2_{\mbf{R}^{-1}_\ell}
    + \frac{1}{2} \norm{\mbf{e}_0(\mbfbar{Y}_0, \mbfbar{X}_0)}^2_{\mbf{S}^{-1}_0} \\
    & \ + \frac{1}{2} \sum^K_{k=1} \norm{\mbf{e}_k(\mbf{f}_{k-1}(\mbfbar{X}_{k-1}, \mbfbar{u}_{k-1}), \mbfbar{X}_k)}^2_{\mbf{Q}^{-1}_k}.
    \numberthis
    \label{eqn:batchsum}
\end{align*}
In \eqref{eqn:batchsum}, $\mbf{e}_k$, $\mbf{e}_\ell$, and $\mbf{e}_0$ are the
interoceptive, exteroceptive, and prior errors, respectively, while
$\mbf{f}_{k-1}$ and $\mbf{g}_\ell$ represent the nonlinear process and
measurement models, respectively.  The notation
${\norm{\mbf{e}}^2_{\mbs{\Sigma}\inv} = \mbf{e}^\trans \mbs{\Sigma}\inv
\mbf{e}}$ denotes the squared Mahalanobis distance, and $\mbf{Q}_k$ and
$\mbf{R}_\ell$ represent the discrete-time covariance on the interoceptive and
exteroceptive errors, respectively.  To minimize \eqref{eqn:batchcost},
\eqref{eqn:batchsum} is repeatedly linearized about the current state estimate
$\mbfbar{X}$, and the local minimizing solution found using, for example,
Gauss-Newton or Levenberg-Marquardt.

\section{Methodology}
\label{sec:methodology}

\begin{figure}[b]
	\sbox\subfigbox{%
	  \resizebox{\dimexpr\columnwidth-1em}{!}{%
		\includegraphics[height=4cm]{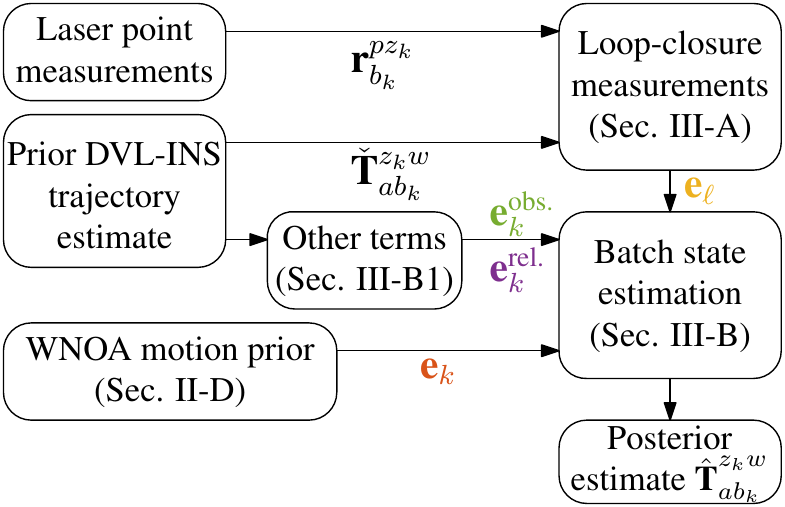}%
	  }%
	}
	\setlength{\subfigheight}{\ht\subfigbox}
	\centering
	\includegraphics[height=\subfigheight]{figs/system_overview-eps-converted-to.pdf}
	\caption{\update{A visual overview of \Cref{sec:methodology}.  Note the colour of
	the error terms is consistent with the factor graph of
	\Cref{fig:factorgraph}.}}
	\label{fig:systemoverview}
\end{figure}

\update{This section describes the primary contributions of this paper, namely
the formulation of laser-based loop-closure measurements and the smooth
incorporation of these measurements into a COTS DVL-INS trajectory estimate.  An
overview of the upcoming methodology is shown in \Cref{fig:systemoverview}.}

\subsection{Loop Closures from Subsea Point Cloud Scans}
\label{sec:loopclosures}

To correct for drift in the DVL-INS trajectory estimate, loop-closure
measurements are obtained by aligning sections of the point cloud scan collected
using a Voyis Insight Pro underwater laser scanner.  The raw laser profiles are
first filtered and registered to the trajectory estimate to produce a 3D point
cloud.  Loop-closure opportunities are identified at path crossings, and
alignment is performed using a multi-step point cloud alignment algorithm.  

\begin{figure}[b]
    \vspace{-3pt}
	\sbox\subfigbox{%
	  \resizebox{\dimexpr\columnwidth-1em}{!}{%
		\includegraphics[height=4cm]{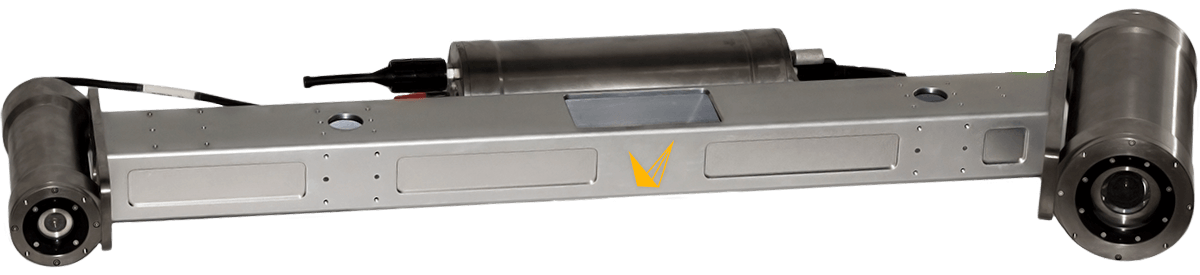}%
	  }%
	}
	\setlength{\subfigheight}{\ht\subfigbox}
	\centering
	\includegraphics[height=\subfigheight]{figs/voyis-uls-500-pro_nogreen.png}
	\caption{An Insight Pro underwater laser scanner developed by Voyis Imaging
     Inc.  The beam emitter is on the left, while the camera is on the right.
     3D point clouds are produced by triangulating the laser beam.  The baseline
     between the emitter and the camera is approximately \SI{1}{\meter}. }
	\label{fig:voyispro}
\end{figure}

\subsubsection{Point Cloud Generation}
\label{sec:pointcloudgeneration}

The Voyis Insight Pro underwater laser scanner, pictured in \Cref{fig:voyispro}, records 2D
profile measurements of the seabed at a frequency of \SI{20}{\Hz}.  To construct
a 3D point cloud, individual laser profiles are registered to the prior DVL-INS
trajectory estimate $\mbfcheck{T}^{z_kw}_{ab_k}$ via 
\vspace{-4pt}
\begin{equation}
    \begin{bmatrix}
        \mbf{r}^{pw}_a \\ 1
    \end{bmatrix} = \mbfcheck{T}^{z_kw}_{ab_k} \mbf{T}^{sz}_{b\ell} \begin{bmatrix}
        \mbf{r}^{ps_k}_{\ell_k} \\ 1
    \end{bmatrix},
    \label{eqn:registerprofiles}
\end{equation}
where ${\mbf{r}^{ps_k}_{\ell_k} \in \rnums^3}$ is a laser measurement of point
$p$ at time $t_k$ resolved in the sensor frame, and ${\mbf{T}^{sz}_{b\ell} \in
SE(3)}$ is a static extrinsics matrix.  Where necessary, the DVL-INS trajectory
is interpolated according to \cite[Sec.~2.4]{Eade2014}
\vspace{-2pt}
\begin{subequations}
    \begin{align}
        \mbfcheck{T}_j =& \ \mbfcheck{T}_i \exp \left( \alpha \log \left( \mbfcheck{T}\inv_i \mbfcheck{T}_k \right) \right), \\
        \alpha =& \ \frac{t_j - t_i}{t_k - t_i},
    \end{align}
\end{subequations}
where ${\mbfcheck{T}_j = \mbfcheck{T}^{z_jw}_{ab_j}}$, and ${t_i < t_j < t_k}$.  The
result of these operations is a filtered point cloud $\mathcal{P}$ resolved in
the local geodetic frame, ${\mathcal{P} = \left\{ \mbf{r}^{p_iw}_a
\right\}^N_{i=1}}$.

\subsubsection{Point Cloud Alignment}
\label{sec:pcalignment}

\begin{figure}[htb]
	\sbox\subfigbox{%
	  \resizebox{\dimexpr\columnwidth-1em}{!}{%
		\includegraphics[height=4cm]{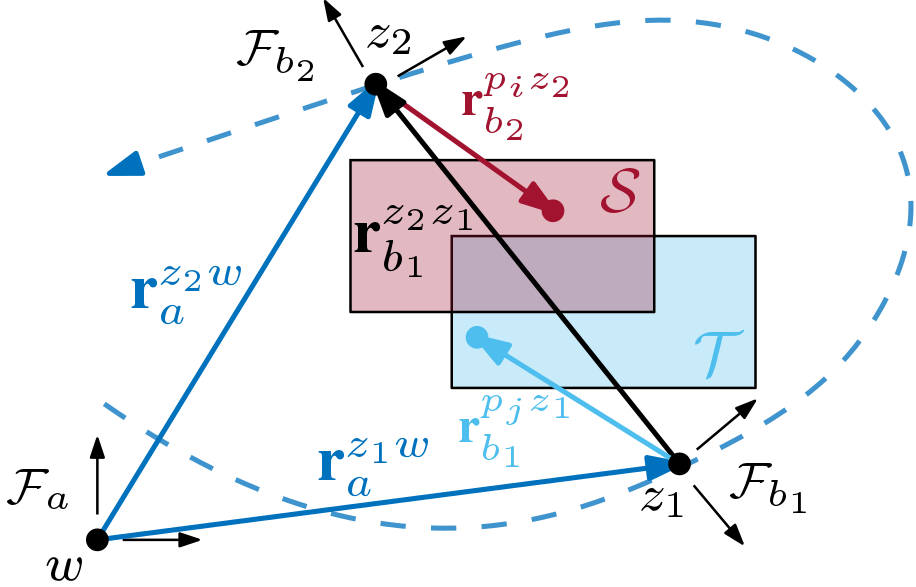}%
	  }%
	}
	\setlength{\subfigheight}{\ht\subfigbox}
	\centering
	\includegraphics[height=\subfigheight]{figs/point_cloud_alignment.png}
	\caption{Generating a loop-closure measurement by aligning source submap
	$\mathcal{S}$ to target submap $\mathcal{T}$.  The vehicle trajectory
	appears as a dashed line.  Submaps ${\mathcal{T} = \{ \mbf{r}^{p_jz_1}_{b_1}
	\}^M_{j=1}}$ and ${\mathcal{S} = \{ \mbf{r}^{p_iz_2}_{b_2} \}^N_{i=1}}$ are
	constructed from the point measurements at vehicle poses
	$\mbf{T}^{z_1w}_{ab_1}$ and $\mbf{T}^{z_2w}_{ab_2}$, respectively.  Point
	cloud alignment produces the loop-closure measurement
	$\mbf{T}^{z_2z_1}_{b_1b_2}$.}
	\label{fig:pointcloudalignment}
\end{figure}

The objective of point cloud alignment is to combine partially overlapping scans
of the same 3D object or scene.  In the context of SLAM, point cloud alignment
is often performed to estimate the relative pose between two or more
observations, for example to reduce odometry drift \cite{Ye2019a} or to bound
navigation drift over time by closing large loops in the trajectory
\cite{Lin2019}.  More formally, the problem of point cloud alignment may be
expressed as 
\begin{equation}
    \mbf{T}^\star_{12} = \argmin_{\mbf{T}\in SE(3)} \frac{1}{2} \sum^{N}_{i=1} \sum^{M}_{j=1} b_{ij} \cdot w_{ij} \cdot \left\Vert \mbf{e}_{ij} \left( \mbf{T}_{12}, \mbf{r}^{p_iz_2}_{b_2}, \mbf{r}^{p_jz_1}_{b_1} \right) \right\Vert^2_{\mbs{\Sigma}\inv_{ij}},
    \label{eqn:pcalignment}
\end{equation}
where the pose ${\mbf{T}^\star_{12} = \left( \mbf{T}^{z_2z_1}_{b_1b_2}
\right)^\star}$ optimally aligns source cloud ${\mathcal{S} =
\{\mbf{r}^{p_iz_2}_{b_2} \}^{N}_{i=1}}$ to target cloud ${\mathcal{T} =
\{\mbf{r}^{p_jz_1}_{b_1} \}^{M}_{j=1}}$.  The Boolean value ${b_{ij} = \{ 0,1
\}}$ assumes a value of 1 if $(p_i,p_j)$ represents an inlier correspondence,
while ${w_{ij} \in [0,1]}$ is a correspondence weight, often computed using a
robust cost function \cite{Babin2019}.  $\mbf{T}^\star_{12}$ is optimal in the
sense that it minimizes the sum of squared weighted errors, often a combination
of point-to-point and point-to-plane errors \cite{Chen1992,Low2004}, with
associated error covariance ${\mbs{\Sigma}_{ij}(\mbf{R}_i, \mbf{R}_j)}$.
$\mbf{R}_i$ and $\mbf{R}_j$ represent the covariance on the point measurements
$\mbf{r}^{p_iz_2}_{b_2}$ and $\mbf{r}^{p_jz_1}_{b_1}$, respectively, with
${\mbf{R}_i = \expect{\delta \mbf{r}_i \, \delta \mbf{r}_i^\trans}}$, ${\delta
\mbf{r}_i = \mbf{r}^{p_iz_2}_{b_2} - \mbfbar{r}^{p_iz_2}_{b_2}}$.  \update{A
depiction of the point cloud alignment problem is shown in
\Cref{fig:pointcloudalignment}.}

In this work, loop-closure locations are identified at simple path crossings on
the $(x,y)$ plane, at time stamps $t_{\ell_1}$ and $t_{\ell_2}$, with
${t_{\ell_1} < t_{\ell_2}}$.  Ordinarily, cross-covariance information would be
used to determine the search region, and thus the required size of the submaps
to construct, using a squared Mahalanobis distance test \cite{Neira2001},
however this information is absent from the DVL-INS trajectory estimate.
Instead, given the inherently low drift rate of the DVL-INS
\cite{sprintnavdatasheet}, the source and target clouds are constructed using a
simple $(x,y)$ distance threshold, e.g.
\begin{equation}
    \mbf{r}^{pw}_a \in \mathcal{T} \, \big| \left\| \begin{bmatrix} \eye & \mbf{0} \end{bmatrix} \left( \mbf{r}^{pw}_a - {\mbfcheck{r}_a}^{\hspace{-0.4em}z_{\ell_1}w} \right) \right\|_2 \le \delta r^\star.
    \label{eqn:findlc}
\end{equation}
In this work a constant value of ${\delta r^\star = \SI{5}{\m}}$ appears to work
well, however a gradually increasing threshold related to the length of the
trajectory could also be used.  

To provide a body-frame relative pose measurement \eqref{eqn:pcalignment}, the
point measurements are first resolved in the body frames, 
\begin{equation}
    \begin{bmatrix}
        \mbf{r}^{pz_\ell}_{b_\ell} \\ 1
    \end{bmatrix} = \left( \mbfcheck{T}^{z_\ell w}_{ab_\ell} \right)\inv \begin{bmatrix}
        \mbf{r}^{pw}_{a} \\ 1
    \end{bmatrix}.
\end{equation}
The point clouds are preprocessed by downsampling to a \SI{5}{\centi \meter}
grid, \update{which reduces the amount of point data \secondupdate{by
approximately a factor of 10} while still preserving high-frequency features of
the scanned object.}  Normal vectors are then estimated using the 40 nearest
Euclidean neighbours.  To account for cases of large navigation drift between
observations, the \texttt{TEASER++} coarse alignment algorithm \cite{Yang2020}
is used to initialize an iterative closest point (ICP)-based fine alignment
algorithm.  To run \texttt{TEASER++}, FPFH feature descriptors \cite{Rusu2009a}
are computed at 3D SIFT keypoints \cite{Lowe2004}, and a set of putative
correspondences is formed from the 10 nearest neighbour matches in
{33-dimensional} FPFH space.  Keypoints and descriptors are computed using the
Point Cloud Library (PCL) v1.9 \cite{Rusu2011}.  Default values are used for all
\texttt{TEASER++} parameters.  

\secondupdate{The combination of SIFT keypoints and FPFH descriptors was
selected for this application following an alignment study on the shipwreck
field dataset introduced in \Cref{sec:fieldresults}.  In this dataset, a vehicle
makes eight passes over a small shipwreck, producing eight point cloud submaps
and 28 unique submap pairs.  27 of the 28 pairs were then aligned using
\texttt{TEASER++} and different detector/descriptor combinations, with one
submap pair excluded due to insufficient overlap.  The study includes three
keypoint detectors and two 3D feature descriptors.  The keypoint detectors are
SIFT, ISS \cite{Zhong2009}, and Harris 3D keypoints \cite{Sipiran2011}, while
the feature descriptors are FPFH and SHOT \cite{Tombari2010}.  These detectors
and descriptors were included in the study both due to their prevalence in the
point cloud alignment literature and the availability of an open-source
implementation in PCL v1.9.  SIFT and Harris 3D keypoint parameters were tuned
slightly to obtain several hundred keypoints in each submap, while default
values were used for ISS keypoints.  For fairness, both FPFH and SHOT
descriptors used the same search radius value of \SI{0.25}{\meter}.  

\newcolumntype{L}{>{\centering}m{0.04\columnwidth}}
\newcolumntype{E}{>{\centering}m{0.11\columnwidth}}
\newcolumntype{D}{>{\centering}m{0.001\columnwidth}}
\begin{table}[b]
    \caption{\secondupdate{Summary statistics from the keypoint detector and
    descriptor alignment study, reported in the format} ${\SI{50}{\percent} \,
    \cdot \, \colour{dark_grey}{\SI{90}{\percent}} \, \cdot \,
    \colour{light_grey}{\textrm{MAX}}}$.  \secondupdate{ Attitude errors $\|
    \delta \mbs{\phi} \|$ and position errors $ \| \delta \mbs{\rho} \| $ are
    computed according to \eqref{eqn:teaserposeerror}.  The lowest value in each
    column is indicated in bold font.  Si~=~SIFT, I~=~ISS, H~=~Harris3D,
    F~=~FPFH, and So~=~SHOT.}}
    \renewcommand{\arraystretch}{1.2}
    \begin{tabularx}{\columnwidth}{LLEDEDEDEDED>{\arraybackslash}E}
        \toprule
            KP & \hspace{-8pt} D & \multicolumn{5}{c}{\hspace{-32pt}$\| \delta \mbs{\phi} \| \, \left[ \deg \right]$} & & \multicolumn{5}{c}{\hspace{-128pt} $\| \delta \mbs{\rho} \| \, \left[ \meter \right]$} \\
            \hline
            \multirow{2}{*}{\begin{tabular}{@{}c@{}} Si \end{tabular}} 
            & \hspace{-8pt} F  & 0.44          & \hspace{-24pt}$\cdot$ & \hspace{-42pt} \colour{dark_grey}{1.05}          & \hspace{-50pt}$\cdot$ & \hspace{-66pt} \colour{light_grey}{\textbf{1.54}} & & \hspace{-86pt} 0.04          & \hspace{-104pt}$\cdot$ & \hspace{-130pt} \colour{dark_grey}{0.09}          & \hspace{-148pt}$\cdot$ & \hspace{-170pt} \colour{light_grey}{\textbf{0.16}} \\
            & \hspace{-8pt} So & \textbf{0.42} & \hspace{-24pt}$\cdot$ & \hspace{-42pt} \colour{dark_grey}{0.86}          & \hspace{-50pt}$\cdot$ & \hspace{-66pt} \colour{light_grey}{92.59}         & & \hspace{-86pt} \textbf{0.03} & \hspace{-104pt}$\cdot$ & \hspace{-130pt} \colour{dark_grey}{\textbf{0.07}} & \hspace{-148pt}$\cdot$ & \hspace{-170pt} \colour{light_grey}{3.33}          \\
            \hline
            \multirow{2}{*}{\begin{tabular}{@{}c@{}} I \end{tabular}} 
            & \hspace{-8pt} F  & 0.49          & \hspace{-24pt}$\cdot$ & \hspace{-42pt} \colour{dark_grey}{\textbf{0.84}} & \hspace{-50pt}$\cdot$ & \hspace{-66pt} \colour{light_grey}{52.34}         & & \hspace{-86pt} 0.04          & \hspace{-104pt}$\cdot$ & \hspace{-130pt} \colour{dark_grey}{0.09}          & \hspace{-148pt}$\cdot$ & \hspace{-170pt} \colour{light_grey}{2.31}          \\
            & \hspace{-8pt} So & 0.61          & \hspace{-24pt}$\cdot$ & \hspace{-42pt} \colour{dark_grey}{176.28}        & \hspace{-50pt}$\cdot$ & \hspace{-66pt} \colour{light_grey}{179.72}        & & \hspace{-86pt} 0.05          & \hspace{-104pt}$\cdot$ & \hspace{-130pt} \colour{dark_grey}{20.12}         & \hspace{-148pt}$\cdot$ & \hspace{-170pt} \colour{light_grey}{22.56}         \\
            \hline
            \multirow{2}{*}{\begin{tabular}{@{}c@{}} H \end{tabular}} 
            & \hspace{-8pt} F  & 0.83          & \hspace{-24pt}$\cdot$ & \hspace{-42pt} \colour{dark_grey}{21.14}         & \hspace{-50pt}$\cdot$ & \hspace{-66pt} \colour{light_grey}{178.44}        & & \hspace{-86pt} 0.08          & \hspace{-104pt}$\cdot$ & \hspace{-130pt} \colour{dark_grey}{3.05}          & \hspace{-148pt}$\cdot$ & \hspace{-170pt} \colour{light_grey}{22.48}         \\
            & \hspace{-8pt} So & 0.83          & \hspace{-24pt}$\cdot$ & \hspace{-42pt} \colour{dark_grey}{52.32}         & \hspace{-50pt}$\cdot$ & \hspace{-66pt} \colour{light_grey}{164.84}        & & \hspace{-86pt} 0.06          & \hspace{-104pt}$\cdot$ & \hspace{-130pt} \colour{dark_grey}{4.22}          & \hspace{-148pt}$\cdot$ & \hspace{-170pt} \colour{light_grey}{18.28} \color{black} \\
        \bottomrule
    \end{tabularx}
	\label{tab:coarsealignmentstudy}
\end{table}

Each submap pair was then aligned by \texttt{TEASER++} using each of the six
detector/descriptor combinations.  The results are given in
\Cref{tab:coarsealignmentstudy}, which lists summary statistics on attitude
errors $\| \delta \mbs{\phi} \|$ and position errors $\| \delta \mbs{\rho} \|$
in the format} ${\SI{50}{\percent} \, \cdot \,
\colour{dark_grey}{\SI{90}{\percent}} \, \cdot \,
\colour{light_grey}{\textrm{MAX}}}$.
\secondupdate{Pose errors were computed between each \texttt{TEASER++}
relative pose estimate $\mbftilde{T}_i$ and the ground-truth relative pose
$\mbf{T}_i$, computed from a well-initialized ICP alignment, as 
\begin{equation}
    \delta \mbs{\xi}_i = \log (\mbf{T}_i\inv \mbftilde{T}_i)^\vee.
    \label{eqn:teaserposeerror}
\end{equation}

Examining \Cref{tab:coarsealignmentstudy}, the combination of SIFT keypoints and
SHOT descriptors (second row) delivers the lowest median attitude error
(\SI{0.42}{\deg}), as well as the lowest \SI{50}{\percent} and \SI{90}{\percent}
position errors.  However, this combination produced at least one outlier
measurement from the 27 submap pairs, while the SIFT+FPFH combination (first
row) produced zero outliers.  In addition, the SIFT+FPFH combination produced
reasonable median and \SI{90}{\percent} errors.  Note the extremely large
position errors in \Cref{tab:coarsealignmentstudy} are due to failed alignments
producing a \SI{180}{\deg} ``flip'' of the (relatively flat) point cloud
submaps.  The submaps are measured at a range of approximately \SI{7}{\meter},
thus ``flipped'' alignments produce a relative body-frame position error of more
than twice this value.  

As the objective of a coarse alignment algorithm is to robustly initialize ICP
as close to ground-truth as possible, the SIFT+FPFH combination was selected for
this application.  Note that \texttt{TEASER++} was chosen for the coarse
alignment algorithm as it has been shown in extensive point cloud alignment
studies \cite{Yang2020} to outperform other coarse alignment methods, for
example FGR \cite{Zhou2016} and RANSAC \cite{Fischler1981}.}


For the fine alignment step, this work uses the Weighted Optimal Linear Attitude
and Translation Estimator (WOLATE) algorithm \cite{Qian2020} within an ICP-based
alignment scheme.  Alignment errors are formulated between each point in the
source cloud and their single nearest neighbour in the target cloud.  A
combination of point-to-point and point-to-plane errors are used, with the
surface variation $v(p_j)$ \cite{Pauly2002} of the target points determining the
type of error used for each association.  Following the study in
\cite{Hitchcox2020a}, the Fractional Root Mean Squared Distance (FRMSD) robust
cost function \cite{Phillips2007} is used for outlier rejection when aligning
structured scans, such as shipwrecks.  The algorithm terminates when the
\secondupdate{pose} differential \secondupdate{$\delta \mbs{\xi}_i$} between two
successive iterations falls below a threshold, or when a maximum number of
iterations is reached.  Following the recommendations in \cite{Pomerleau2013}
for best practices when reporting ICP algorithms, the preprocessing steps and
relevant parameters are summarized in \Cref{tab:pcparams}.

\setlength{\tabcolsep}{4pt}
\begin{table}[t]
    \centering
    \caption{ICP preprocessing and alignment parameters}
    \label{tab:pcparams}
    \renewcommand{\arraystretch}{1.2}
    \begin{tabularx}{\columnwidth}{llX}
    \toprule
    Stage & Configuration & Description \\ \hline
    Preprocessing & \texttt{VoxelGrid} & Downsample to
    \SI{5}{\centi\meter} grid \\
    & \texttt{Normals} & 40 nearest neighbours \\
    Keypoints & \texttt{3D SIFT} & PCL v1.9 implementation \\
    Descriptors & \texttt{FPFH} & PCL v1.9 implementation \\
    Coarse align. & \texttt{TEASER++} & 10 matches, default params. \\ 
    ICP data assn. & \texttt{KDTree} & Single nearest neighbour \\
    ICP error min. & \texttt{Mixed} & \texttt{Pt-Pl} if $v(p_j) < \SI{3e-2}{}$
    \\
    Outlier reject. & \texttt{FRMSD} & Default params. from \cite{Phillips2007} \\
    Termination & \texttt{Diff.} & ${\| \delta \mbs{\phi}_i \|_2 <
    \SI{1e-2}{\radian}}$, and ${\| \delta \mbs{\rho}_i \|_2 <
    \SI{1e-3}{\meter}}$ \\
    & \texttt{Counter} & 20 iterations max \\
    \bottomrule
    \end{tabularx}
\end{table}

\subsubsection{Loop-Closure Measurement Model}
\label{sec:measurementmodel}

Point cloud alignment yields the loop-closure measurement
\begin{equation}
    \mbs{\Xi}_{\ell_1 \ell_2} \triangleq \mbf{T}^{z_{\ell_2}z_{\ell_1}}_{b_{\ell_1}b_{\ell_2}} = \left( \mbf{T}^{z_{\ell_1}w}_{ab_{\ell_1}} \right)\inv \mbf{T}^{z_{\ell_2}w}_{ab_{\ell_2}} \in SE(3),
\end{equation}
and, given the perturbation scheme \eqref{eqn:perturbation}, the noise model is
\begin{subequations}
    \begin{align}
        \mbs{\Xi}_{\ell_1 \ell_2} =& \ \mbf{g}_\ell (\mbfcheck{T}_{\ell_1}, \mbfcheck{T}_{\ell_2}, \delta \mbs{\xi}_\Xi) \\
        =& \ \mbsbar{\Xi}_{\ell_1 \ell_2} \exp(-\delta \mbs{\xi}^\wedge_{\Xi}), \\
        \delta \mbs{\xi}_{\Xi} \sim& \ \mathcal{N}(\mbf{0}, \mbf{R}_\Xi),
    \end{align}
\end{subequations}
\update{where the shorthand ${\mbfcheck{T}_{\ell_i} =
\mbfcheck{T}^{z_{\ell_i}w}_{ab_{\ell_i}}, i = 1, 2}$ is used for readability}.
The covariance $\mbf{R}_\Xi$ on the loop-closure measurement may be obtained
from the point cloud alignment algorithm in a number of ways, for example the
linearization-based approach in \cite{Brossard2020}.

\subsection{Updating the Trajectory}
\label{sec:updatetrajectory}

\subsubsection{Formulating the Objective Function}
\label{sec:formulateJ}

\begin{figure}[tb]
	\sbox\subfigbox{%
	  \resizebox{\dimexpr\columnwidth-1em}{!}{%
		\includegraphics[height=4cm]{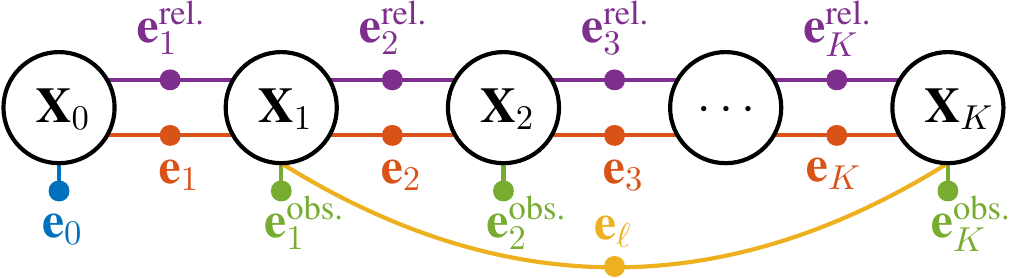}%
	  }%
	}
	\setlength{\subfigheight}{\ht\subfigbox}
	\centering
	\includegraphics[height=\subfigheight]{figs/factor_graph-eps-converted-to.pdf}
	\caption{The factor graph corresponding to the batch state estimation
	problem.  The formation of WNOA factors $\mbf{e}_k$ and relative pose
	factors $\mbf{e}^{\textrm{rel.}}_k$ allow corrections from loop-closure
	factor $\mbf{e}_\ell$ to propagate throughout the graph.}
	\label{fig:factorgraph}
\end{figure}

The objective is now to condition the prior DVL-INS trajectory estimate on the
newly available loop-closure measurements.  This is accomplished through
nonlinear batch state estimation, described in \Cref{sec:batchestimation}.
\update{This section describes how the error terms in the batch problem are
formulated, and \Cref{fig:factorgraph} shows the resulting factor graph.}

First, the prior, process, and measurement errors must be defined.  Given the
augmented navigation state \eqref{eqn:navstate}, errors must be defined for the
$SE(3)$ pose and for the generalized velocity.  Using both the left-invariant
error definition \eqref{eqn:lefterror} and the constant velocity WNOA motion
prior, the prior error is
\begin{equation}
    \mbf{e}_0 = \begin{bmatrix}
        \mbf{e}^\xi_0 \\ \mbf{e}^\varpi_0
    \end{bmatrix} = \begin{bmatrix}
        \log \left( \mbf{T}_0\inv \mbf{Y}_0 \right)^\vee \\
        \mbs{\varpi}_0 - \mbs{\psi}_0
    \end{bmatrix},
    \label{eqn:priorerror}
\end{equation}
where ${ ( \mbf{Y}_0, \mbs{\psi}_0 ) }$ is the prior estimate on the first
navigation state.  The process errors take the form
\begin{equation}
    \mbf{e}_k = \begin{bmatrix}
        \mbf{e}^\xi_k \\ \mbf{e}^\varpi_k
    \end{bmatrix} = \begin{bmatrix}
        \log \left( \mbf{T}_k\inv \mbftilde{T}_k \right)^\vee \\
        \mbs{\varpi}_k - \mbs{\varpi}_{k-1}
    \end{bmatrix},
    \label{eqn:odomerror}
\end{equation}
where the predicted pose at time $t_k$,
\begin{equation}
    \mbftilde{T}_k = \mbf{f}_{k-1}(\mbf{T}_{k-1}, \mbs{\varpi}_{k-1}) = \mbf{T}_{k-1} \exp(T \mbs{\varpi}^\wedge_{k-1}),
\end{equation}
arises from a forward Euler discretization of the continuous-time $SE(3)$
kinematics \eqref{eqn:ctkin} over an integration period of ${T = t_k -
t_{k-1}}$.  The loop-closure errors are
\begin{equation}
    \mbf{e}_\ell = \mbf{T}_{\ell_2}\inv \mbftilde{T}_{\ell_2} = \mbf{T}_{\ell_2}\inv \mbf{T}_{\ell_1} \mbs{\Xi}_{\ell_1\ell_2},
    \label{eqn:lcerror}
\end{equation}
where $\mbf{T}_{\ell_1}$ and $\mbf{T}_{\ell_2}$ are the two poses involved in
loop-closure measurement $\ell$.  Additionally, it was discovered in testing
that including a relative pose constraint between each subsequent pair of poses
helped the loop-closure correction to propagate throughout the trajectory.  The
relative pose errors take the same form as the loop-closure errors, 
\begin{equation}
    \mbf{e}^{\textrm{rel.}}_k = \mbf{T}_k\inv \mbftilde{T}_k = \mbf{T}_k\inv \mbf{T}_{k-1} \mbs{\Xi}_{k-1,k},
    \label{eqn:relposeerror}
    \vspace{-1pt}
\end{equation}
where the relative pose measurements are taken directly from the initializing
solution, 
\vspace{-3pt}
\begin{equation}
    \mbs{\Xi}_{k-1,k} = \mbfcheck{T}\inv_{k-1} \mbfcheck{T}_k.
    \vspace{-1pt}
\end{equation}
Finally, since roll, pitch, and depth are directly observable AUV states
\cite{Barkby2011a}, errors are included of the form 
\begin{equation}
    \mbf{e}^{\textrm{obs.}}_k = \begin{bmatrix}
        e^{\phi_1}_k \\ e^{\phi_2}_k \\ e^{\textrm{z}}_k
    \end{bmatrix} = \mbf{D} \mbf{E}_k \log \left( \mbf{T}_k\inv \mbfcheck{T}_k \right)^\vee,
    \label{eqn:obserror}
    \vspace{-3pt}
\end{equation}
where 
\begin{equation}
    \mbf{D} = \begin{bmatrix}
        \eye & \mbf{0} & \mbf{0} \\ 
        \mbf{0} & \mbf{0} & 1
    \end{bmatrix} \in \rnums^{3\times 6}, 
    \quad \mbf{E}_k = \begin{bmatrix}
        \eye & \\ & \mbf{C}_{ab_k} \mbf{J}^\ell(\mbf{e}^\phi_k)
    \end{bmatrix},
\end{equation}
where $\mbf{J}^\ell$ is the left Jacobian of $SO(3)$
\cite[Sec.~7.1.3]{Barfoot2017}, and
\vspace{-3pt}
\begin{equation}
    \mbf{e}^\xi_k = \begin{bmatrix} (\mbf{e}^\phi_k)^\trans & (\mbf{e}^\rho_k)^\trans \end{bmatrix}^\trans.
\end{equation}
The least-squares objective function \eqref{eqn:batchsum} is then augmented as 
\vspace{-4pt}
\begin{equation}
    J_{\textrm{aug.}}(\mbf{X}) = J(\mbf{X}) + \frac{1}{2} \sum_{k=1}^K \left( \big \Vert \mbf{e}^{\textrm{rel.}}_k \big \Vert^2_{\mbf{R}\inv_{\textrm{rel.}}} +  \big \Vert \mbf{e}^{\textrm{obs.}}_k \big \Vert^2_{\mbf{R}\inv_{\textrm{obs.}}} \right), 
    \label{eqn:jprime}
\end{equation}
where $\mbf{R}_{\textrm{rel.}}$ and $\mbf{R}_{\textrm{obs.}}$ are considered to
be additional hyperparameters.  Together, the relative pose errors
\eqref{eqn:relposeerror} promote loop-closure propagation, while the WNOA errors
\eqref{eqn:odomerror} promote smoothing.  The hyperparameters $\mbc{Q}$ and
$\mbf{R}_{\textrm{rel.}}$ may be tuned to control the smoothness of the
posterior solution, while $\mbf{R}_{\textrm{obs.}}$ is tuned to ensure the
posterior does not stray too far in observable dimensions.

The batch estimation problem is visualized in the factor graph of
\Cref{fig:factorgraph}.  Note that, in contrast to the initial factor graph in
\Cref{fig:factorgraphins}, there are now factors linking adjacent nodes.  This
will allow corrections from the loop-closure measurements to propagate
throughout the pose graph, as required. 

\subsubsection{Minimizing the Objective Function}
\label{sec:minimizeJ}
To minimize \eqref{eqn:jprime}, the estimation errors are repeatedly linearized
about the current navigation state estimate $\mbfbar{X}$.  Perturbing the
navigation state as 
\begin{subequations}
    \begin{align}
        \mbf{T} =& \ \mbfbar{T} \exp(-\delta \mbs{\xi}^\wedge), 
        \label{eqn:perturbT} \\
        \mbs{\varpi} =& \ \mbsbar{\varpi} + \delta \mbs{\varpi}, 
        \label{eqn:perturbvarpi} \\
        \delta \mbf{x} =& \begin{bmatrix}
            \delta \mbs{\xi}^\trans & \delta \mbs{\varpi}^\trans
        \end{bmatrix}^\trans,
    \end{align}
\end{subequations}
the prior error \eqref{eqn:priorerror} is linearized as 
\begin{equation}
    \mbf{e}_0 = \mbfbar{e}_0 + \mbf{F}_0^0 \delta \mbf{x}_0 + \mbf{M}_0 \delta \mbf{y}_0,
\end{equation}
where ${\delta \mbf{y}_0 = \begin{bmatrix} \delta \mbs{\eta}_0^\trans & \delta
\mbs{\psi}_0^\trans \end{bmatrix}^\trans}$, ${\mbf{S}_0 = \expect{\delta
\mbf{y}_0 \, \delta \mbf{y}_0^\trans}}$, and where the prior Jacobians are
\begin{subequations}
    \begin{align}
        \mbf{F}_0^0 =& \ \blkdiag(\mbf{J}^\ell ( \mbfbar{e}^\xi_0 )\inv, \eye), \\
        \mbf{M}_0   =& \ \blkdiag(-\mbf{J}^{\textrm{r}} ( \mbfbar{e}^\xi_0 )\inv, -\eye),
    \end{align}
\end{subequations}
with $\mbf{J}^\ell$ being the left Jacobian of $SE(3)$.  Note that detailed
derivations of the work appearing in this section are available in the
supplementary material in \Cref{apx:derivations}.  The discrete-time process
errors \eqref{eqn:odomerror} are linearized as 
\begin{equation}
    \mbf{e}_k = \mbfbar{e}_k + \update{\mbf{F}^k_{k-1}} \delta \mbf{x}_{k-1} + \update{\mbf{F}^k_k} \delta \mbf{x}_k,
\end{equation}
where the process error Jacobians are given by 
\begin{subequations}
    \begin{align*}
        \mbf{F}^k_{k-1} =& \begin{bmatrix}
            \mbf{U}_{k-1} & \mbf{V}_{k-1} \\
            \mbf{0} & -\eye
        \end{bmatrix}, 
        \numberthis \\
        \mbf{U}_{k-1} =& \ -\mbf{J}^\textrm{r}(\mbfbar{e}^\xi_k)\inv \Adj (\exp (-T \mbsbar{\varpi}_{k-1}^\wedge)), \\
        \mbf{V}_{k-1} =& \ T\mbf{J}^\textrm{r}(\mbfbar{e}^\xi_k)\inv \mbf{J}^\textrm{r}(T \mbsbar{\varpi}_{k-1}), \\
        \mbf{F}^k_k =& \begin{bmatrix}
            \mbf{J}^\ell(\mbfbar{e}^\xi_k)\inv & \mbf{0} \\ \mbf{0} & \eye
        \end{bmatrix}.
        \numberthis
    \end{align*}
\end{subequations}
The loop-closure errors \eqref{eqn:lcerror} are linearized as 
\begin{equation}
    \mbf{e}_\ell = \mbfbar{e}_\ell + \mbf{H}^\ell_{\ell_1} \delta \mbs{\xi}_{\ell_1} + \mbf{H}^\ell_{\ell_2} \delta \mbs{\xi}_{\ell_2} + \mbf{M}_\ell \delta \mbs{\xi}_{\Xi}, 
\end{equation}
with corresponding Jacobians 
\begin{subequations}
    \begin{align}
        \mbf{H}^\ell_{\ell_1} =& \ -\mbf{J}^{\textrm{r}} (\mbfbar{e}_\ell)\inv \Adj(\mbsbar{\Xi}_{\ell_1\ell_2}\inv), \\
        \mbf{H}^\ell_{\ell_2} =& \ \mbf{J}^\ell(\mbfbar{e}_\ell)\inv, \\
        \mbf{M}_\ell   =& \ -\mbf{J}^{\textrm{r}}(\mbfbar{e}_\ell)\inv. 
    \end{align}
\end{subequations}
The relative pose errors \eqref{eqn:relposeerror} are linearized in the same
manner.  Finally, errors on the observable states \eqref{eqn:obserror} are
linearized by approximating
\vspace{-4pt}
\begin{equation}
    \mbf{C}_{ab_k} = \mbfbar{C}_{ab_k} \exp \left( -\delta \mbs{\phi}_k^\times \right) \approx \mbfbar{C}_{ab_k} \left( \eye - \delta \mbs{\phi}_k^\times \right).
    \vspace{-2pt}
\end{equation}
Assuming ${\mbf{e}^\rho_k \to \mbf{0}}$ as the optimization proceeds, this
yields
\vspace{-1pt}
\begin{align}
    \mbf{e}^{\textrm{obs.}}_k =& \ \mbfbar{e}^{\textrm{obs.}}_k + \mbf{H}^{\textrm{obs.}}_k \delta \mbs{\xi}_k, \\
    \mbf{H}^{\textrm{obs.}}_k =& \ \mbf{D} \mbf{E}_k(\mbfbar{C}_{ab_k}, \mbfbar{e}^\phi_k) \mbf{J}^\ell (\mbfbar{e}^\xi_k)\inv.
    \vspace{-2pt}
\end{align}
The final step is to determine the covariance on the discrete-time WNOA process
errors.  This is done by discretizing the power spectral density $\mbc{Q}$ via
\cite[(4.110)]{Farrell2008}
\vspace{-2pt}
\begin{equation}
    \mbf{Q}_k = \int^{t_k}_{t_{k-1}} \mbf{A}(t_k,s) \mbf{L}(s) \mbc{Q}(s) \left( \mbf{A}(t_k,s) \mbf{L}(s) \right)^\trans \dee s,
    \label{eqn:dtq}
    \vspace{-2pt}
\end{equation}
where ${ \mbf{A}, \mbf{L} }$ characterize the continuous-time error kinematics,
which for the WNOA motion prior take the form 
\vspace{-1pt}
\begin{equation}
    \delta \mbfdot{x}(t) = \underbrace{\begin{bmatrix} 
        -\adj(\mbsbar{\varpi}^\wedge) & -\eye \\ \mbf{0} & \mbf{0}
    \end{bmatrix}}_{\mbf{A}} \delta \mbf{x}(t) + \underbrace{\begin{bmatrix}
        \mbf{0} \\ \eye
    \end{bmatrix}}_{\mbf{L}} \delta \mbf{w}(t),
\end{equation}
with ${\delta \mbf{w}(t) \sim \mathcal{GP} (\mbf{0}, \mbc{Q}(t-t^\prime))}$.
The exact solution to \eqref{eqn:dtq} may be obtained via the matrix exponential
\cite{VanLoan1978}, however to avoid this expensive operation this work makes
use of a third-order approximation in $\mbf{A}$ \cite[(4.119)]{Farrell2008},
\vspace{-3pt}
\begin{align*}
    \mbf{Q}_k &\approx T \mbs{\Upsilon} + \frac{T^2}{2} \left( \mbf{A} \mbs{\Upsilon} + \mbs{\Upsilon} \mbf{A}^\trans \right) \\
    &\hspace{-0.6em} + \frac{T^3}{6}  \left( \mbf{A}^2 \mbs{\Upsilon} + 2 \mbf{A} \mbs{\Upsilon} \mbf{A}^\trans + \mbs{\Upsilon} \left( \mbf{A}^\trans \right)^2 \right) 
    \numberthis \\
    &\hspace{-0.6em} + \frac{T^4}{24} \left( \mbf{A}^3 \mbs{\Upsilon} + 3 \mbf{A}^2 \mbs{\Upsilon} \mbf{A}^\trans + 3 \mbf{A} \mbs{\Upsilon} \left( \mbf{A}^\trans \right)^2 + \mbs{\Upsilon} \left( \mbf{A}^\trans \right)^3 \right),
    \vspace{-4pt}
\end{align*}
where ${\mbs{\Upsilon} = \mbf{L} \mbc{Q} \mbf{L}^\trans}$.  Finally, the
minimizing solution for a single iteration of Gauss-Newton is
\vspace{-1pt}
\begin{equation}
    \delta \mbf{x}_\star = \begin{bmatrix}
        \delta \mbs{\xi}_\star \\ \delta \mbs{\varpi}_\star 
    \end{bmatrix} = -\left( \mbs{\Gamma}^\trans \mbf{W} \mbs{\Gamma} \right)\inv \mbs{\Gamma}^\trans \mbf{W} \mbf{e}.
\end{equation}
Jacobian $\mbs{\Gamma}$ is given by 
\vspace{-3pt}
\begin{align}
    \mbs{\Gamma} =& \begin{bmatrix}
        \mbf{F}^\trans & \mbf{H}^\trans & {\mbf{H}^{\textrm{rel.}}}^\trans & {\mbf{H}^{\textrm{obs.}}}^\trans
    \end{bmatrix}^\trans, \\ 
    \mbf{F} =& \begin{bmatrix}
        \mbf{F}^0_0 & & & \\
        \mbf{F}^1_0 & \mbf{F}^1_1 & & \\
        & \ddots & \ddots & \\
        & & \mbf{F}^K_{K-1} & \mbf{F}^K_K
    \end{bmatrix}, \\
    \mbf{H} =& \begin{bmatrix}
        & \mbf{H}^1_{\ell_1} & & & \mbf{H}^1_{\ell_2} & \\
        & & & \vdots & & \\
        & & \mbf{H}^L_{\ell_1} & & & \mbf{H}^L_{\ell_2}
    \end{bmatrix}, 
\end{align}
\vspace{-12pt}
\begin{align}
    \mbf{H}^{\textrm{rel.}} =& \begin{bmatrix}
        \mbf{H}^{\textrm{rel.},1}_0 & \mbf{H}^{\textrm{rel.},1}_1 & & \\
        & \ddots & \ddots & \\
        & & \mbf{H}^{\textrm{rel.},K}_{K-1} & \mbf{H}^{\textrm{rel.},K}_K
    \end{bmatrix}, \\
    \mbf{H}^{\textrm{obs.}} =& \begin{bmatrix}
        \mbf{0} & \mbf{H}^{\textrm{obs.}}_1 & & \\
        & & \ddots & \\
        & & & \mbf{H}^{\textrm{obs.}}_K
    \end{bmatrix},
\end{align}
and weighting matrix ${\mbf{W} = \mbs{\Sigma}\inv}$ is described by 
\vspace{6pt}
\begin{equation}
    \mbs{\Sigma} = \blkdiag \left( \mbf{M}_0 \mbf{S}_0 \mbf{M}_0^\trans, \mbf{Q}_{1:K}, \mbf{R}_{1:L}, \mbf{R}^{\textrm{rel.}}_{1:K}, \mbf{R}^{\textrm{obs.}}_{1:K} \right),
    \vspace{6pt}
\end{equation}
where, for the loop-closure errors, 
\vspace{3pt}
\begin{equation}
    \mbf{R}_\ell = \mbf{M}_\ell \mbf{R}_\Xi \mbf{M}_\ell^\trans.
    \vspace{3pt}
\end{equation}
The column matrix of errors is simply 
\vspace{3pt}
\begin{equation}
    \mbf{e} = \begin{bmatrix}
        \mbf{e}_0^\trans & \mbf{e}_{1:K}^\trans & \mbf{e}_{1:L}^\trans & \left( \mbf{e}^{\textrm{rel.}}_{1:K} \right)^\trans & \left( \mbf{e}^{\textrm{obs.}}_{1:K} \right)^\trans
    \end{bmatrix}^\trans.
    \vspace{6pt}
\end{equation}
Finally, in accordance with the perturbation scheme
(\ref{eqn:perturbT},~\ref{eqn:perturbvarpi}), the state update is given by 
\vspace{5pt}
\begin{subequations}
    \begin{align}
        \mbf{T} \leftarrow& \ \mbf{T} \exp(-\delta \mbs{\xi}^\wedge_\star), \\
        \mbs{\varpi} \leftarrow& \ \mbs{\varpi} + \delta \mbs{\varpi}_\star.
    \end{align}
\end{subequations}
%

\secondupdate{
\subsubsection{Rejecting False Loop-Closure Measurements}
\label{sec:outlierrejection}

Measurement outliers are inevitable in real-world robotics problems, and a
robust implementation of the proposed methodology requires a method to identify
and reject false loop-closure measurements.  Many approaches exist in the
literature for rejecting loop-closure measurement outliers, for example
switchable constraints \cite{Suenderhauf2012}, expectation-maximization
\cite{Lee2013}, and graduated non-convexity \cite{Yang2020b}.  

This application uses a recently developed adaptive robust cost function (RCF)
to reject false loop-closure measurements, owing to its ability to handle
multivariate, mixed-unit error definitions, such as loop-closure errors
\eqref{eqn:lcerror}, in a statistically sound manner \cite{Hitchcox2022b}.  The
RCF assigns a weight ${w_\ell(\epsilon_\ell(\mbf{e}_\ell)) \in (0,1]}$ to
loop-closure error $\mbf{e}_\ell$ according to the Mahalanobis distance
associated with the error,
\vspace{3pt}
\begin{equation}
    \epsilon_\ell(\mbf{e}_\ell) = \| \mbf{e}_\ell \|_{\mbs{\Sigma}_\ell\inv} \in \mathbb{R}_{\ge0},
    \vspace{3pt}
\end{equation}
where, ordinarily, the covariance $\mbs{\Sigma}_\ell$ on the (relative)
loop-closure measurement would be the relative uncertainty computed between the
two vehicle poses involved in the measurement \cite{Mangelson2019}.  Since the
DVL-INS output does not contain the joint covariance information required to
properly compute $\mbs{\Sigma}_\ell$, a constant value is used here,
\vspace{3pt}
\begin{equation}
    \mbs{\Sigma}_\ell = \blkdiag(\sigma^2_{\phi_\textrm{out}} \eye, \sigma^2_{\rho_\textrm{out}} \eye),
    \vspace{3pt}
\end{equation}
with ${\sigma_{\phi_\textrm{out}} = \SI{1}{\deg}}$ and
${\sigma_{\rho_\textrm{out}} = \SI{1}{\meter}}$.  These $1\sigma$ values reflect
the low heading uncertainty of the survey-grade DVL-INS used in the field
experiments, as well as the static search bound used for the loop-closure
detection method \eqref{eqn:findlc}.

}

\section{Results}
\label{sec:results}

\subsection{\update{Assessing the Quality of the State Estimate}}
\label{sec:measuringselfconsistency}

The methodology described in \Cref{sec:methodology} conditions an existing state
estimate on newly available loop-closure measurements.  Since loop closures
provide relative constraints between poses, as shown in \Cref{fig:factorgraph},
it is expected that this approach will 
\begin{packed_enum}
    \item reduce \textit{relative pose errors} throughout the trajectory; and 
    \item produce a more self-consistent point cloud map, as measured by a
    reduction in the \textit{point disparity error} in overlapping regions.
\end{packed_enum}

\subsubsection{\update{Measuring Errors in the Estimated Trajectory}}
\label{sec:sctraj}

A pose-based relative error metric based on \cite{Kuemmerle2009} is used to
measure the accuracy of the estimated trajectory.  Let ${\mbfhat{T}_k\in SE(3)}$
represent the estimated pose at time $t_k$.  The pose at time $t_k$ relative to
the pose at time $t_\ell$ is 
\vspace{-2pt}
\begin{equation}
    \delta \mbfhat{T}_{\ell k} = \mbfhat{T}\inv_\ell \mbfhat{T}_k,
    \label{eqn:relativepose}
    \vspace{-2pt}
\end{equation}
where $\mbfhat{T}_\ell$ is taken to be the earliest pose involved in any
loop-closure measurement.  The relative pose error may then be expressed as 
\vspace{-6pt}
\begin{equation}
    \mbf{E}^{\textrm{rel.}}_k = \delta \mbf{T}\inv_{\ell k} \delta \mbfhat{T}_{\ell k} = \begin{bmatrix}
        \delta \mbf{C}_k & \delta \mbf{r}_k \\ \mbf{0} & 1
    \end{bmatrix},
    \label{eqn:relativeposemetric}
    \vspace{-1pt}
\end{equation}
where $\delta \mbf{T}_{\ell k}$ is \eqref{eqn:relativepose} evaluated using the
ground-truth trajectory.  \secondupdate{Relative pose errors on $SE(3)$ may then
be expressed in the Lie algebra as}
\vspace{-5pt}
\begin{equation}
    \secondupdate{\delta \mbf{e}^\textrm{rel.}_k = \log \left( \mbf{E}^\textrm{rel.}_k \right)^\vee = \begin{bmatrix}
        (\delta \mbs{\phi}^\textrm{rel.}_k)^\trans \ (\delta \mbs{\rho}^\textrm{rel.}_k)^\trans
    \end{bmatrix}^\trans.}
    \label{eqn:posemetric}
\end{equation}
However, since the AUV trajectories studied in this work are largely planar, the
accuracy of the trajectory estimate is reported as the norm of the relative
displacement error $\delta \mbf{r}_k$ projected on the $(x,y)$ plane, 
\vspace{-2pt}
\begin{equation}
    e^{\textrm{rel.}}_k = \norm{ \begin{bmatrix} \eye & \mbf{0} \end{bmatrix} \delta \mbf{r}_k}_2.
    \label{eqn:displacementmetric}
    \vspace{-2pt}
\end{equation}
Assessing performance using a relative metric such as
\eqref{eqn:relativeposemetric} avoids the problem of aligning the estimated and
ground-truth trajectories, which would be required if attempting to provide an
absolute performance metric \cite{Kuemmerle2009}.  In addition to being
intuitive and easy to visualize, the relative planar displacement metric
\eqref{eqn:displacementmetric} allows for direct comparison to other navigation
solutions within the subsea industry, where position drift is often reported on
the $(x,y)$ plane as a percentage of distance traveled.  For estimates
incorporating multiple loop closures along the length of the trajectory, the
relative displacement error \eqref{eqn:displacementmetric} is expected to remain
bounded over time.

\subsubsection{Measuring Self-Consistency in the Point Cloud Map}
\label{sec:scmap}

Performance is also assessed by evaluating the self-consistency in overlapping
regions of the point cloud map.  A point cloud map generated from an accurate
trajectory estimate is expected to be well-aligned, or ``crisp.'' In contrast, a
map produced using a drifting trajectory estimate will see ``double vision''
effects in overlapping regions due to poorly aligned scans.  To assess
self-consistency in the point cloud map, this work uses the \textit{point
disparity} metric from \cite{Roman2006}.  For point clouds ${\mathcal{S} =
\{\mbf{r}^{p_iw}_a \}^N_{i=1}}$ and ${\mathcal{T} = \{\mbf{r}^{p_jw}_a
\}^M_{j=1}}$, this metric is 
\vspace{-2pt}
\begin{equation}
    e^{\textrm{rel.}}_j = \norm{\mbf{r}^{p_jw}_a - \mbf{r}^{p_iw}_a}_2,
    \label{eqn:pointconsistencymetric}
    \vspace{-1pt}
\end{equation}
where point $p_j$ in $\mathcal{T}$ is the nearest Euclidean neighbour to point
$p_i$ in $\mathcal{S}$.  Note \eqref{eqn:pointconsistencymetric} is only
computed within the intersection of $\mathcal{S}$ and $\mathcal{T}$ to prevent
cloud size from biasing the metric.  The point disparity metric is relative, and
may be computed without access to ground-truth information \cite{Roman2006},
making it especially important for field trials where a ground-truth map is not
available.  \secondupdate{Note the point disparity metric is susceptible to
map-to-map error, whereby an erroneous group of two or more well-aligned submaps
would produce low disparity errors, despite separation from the true submap
group.  The results in the following studies were visually checked to ensure the
absence of this error, though extending \eqref{eqn:pointconsistencymetric} to
account for map-to-map error is an interesting avenue for future research. }

\subsection{Hyperparameter Values}
\label{sec:hyperparameters}

The methodology described in \Cref{sec:methodology} involves three sets of
hyperparameters.  These are 
\begin{packed_enum}
    \item $\mbc{Q}$, the PSD on the white noise Gaussian process;
    \item $\mbf{R}^{\textrm{rel.}}_k$, the covariance on the relative pose
    errors; and 
    \item $\mbf{R}^{\textrm{obs.}}_k$, the covariance on the roll, pitch, and
    depth errors, all of which are assumed to be observable. 
\end{packed_enum}
The hyperparameter sets take the form 
\vspace{-4pt}
\begin{subequations}
    \begin{align}
        \mbc{Q} =& \ \diag(\mathcal{Q}_{\dot{\omega}} \eye, \mathcal{Q}_{\dot{\nu}} \eye), \\
        \mbf{R}^\textrm{rel.}_k =& \ \diag(\sigma^2_\phi \eye, \sigma^2_\rho \eye), \\
        \mbf{R}^\textrm{obs.}_k =& \ \diag(\sigma^2_{\textrm{rp}} \eye, \sigma^2_{\textrm{z}}),
    \end{align}
    \label{eqn:hyperparameters}%
\end{subequations}
where $\mathcal{Q}_{\dot{\omega}}$, $\mathcal{Q}_{\dot{\nu}}$ are the power
spectral densities on the body-centric angular and linear acceleration,
respectively, and $\sigma^2_\phi$, $\sigma^2_\rho$ are the variances on the
body-centric angular and linear displacement, respectively.
$\sigma^2_{\textrm{rp}}$ is the variance on vehicle roll and pitch, and
$\sigma^2_{\textrm{z}}$ is the variance on vehicle depth.  While this
hyperparameter structure is simple, it was found to work well for both simulated
and field experiments, and generally makes physical sense.  For example,
\cite{Tang2019} scales the values of a diagonal $\mbc{Q}$ matrix to reflect the
nonholonomic constraints of an automobile.  In contrast, AUVs are highly
maneuverable, leading to the selection of isotropic hyperparameters in
\eqref{eqn:hyperparameters}. 

This section contains results from both simulated and field experiments.  The
hyperparameter values used to obtain each set of results are summarized in
\Cref{tab:hypparamtable}.  These values were hand-selected to produce good
results without extensive tuning, and were modified based on the quality of the
DVL-INS state estimate and the frequency at which DVL-INS data were available.

\newcolumntype{Y}{>{\centering\arraybackslash}X}
\begin{table}[hb]
    \centering
    \caption{Hyperparameter values used in experiments}
    \renewcommand{\arraystretch}{1.2}
    \begin{tabularx}{\columnwidth}{YYYY}
    \toprule
    Hyperparameter & Unit & Simulated & Field \\
    \hline
    $\mathcal{Q}_{\dot{\omega}}$ & \SI{}{\radian\squared\second\tothe{-3}} & \SI{1e-2}{} & \SI{1e-2}{} \\
    $\mathcal{Q}_{\dot{\nu}}$ & \SI{}{\meter\squared\second\tothe{-3}} & \SI{9e-4}{} & \SI{1e-4}{} \\
    $\sigma_\phi$ & \SI{}{\radian} & \SI{1e-3}{} & \SI{1e-3}{} \\
    $\sigma_\rho$ & \SI{}{\meter} & \SI{1e-4}{} & \SI{1e-3}{} \\ 
    $\sigma_{\textrm{rp}}$ & \SI{}{\deg} & 5 & 5 \\
    $\sigma_{\textrm{z}}$ & \SI{}{\meter} & 0.25 & 0.25 \\
	\bottomrule
    \end{tabularx}
	\label{tab:hypparamtable}
\end{table}

\subsection{Simulation Results: AUV Area Inspection}
\label{sec:simresults}
%

\begin{figure}[b]
    \vspace{-12pt}
	\centering
	\includegraphics[width=\columnwidth]{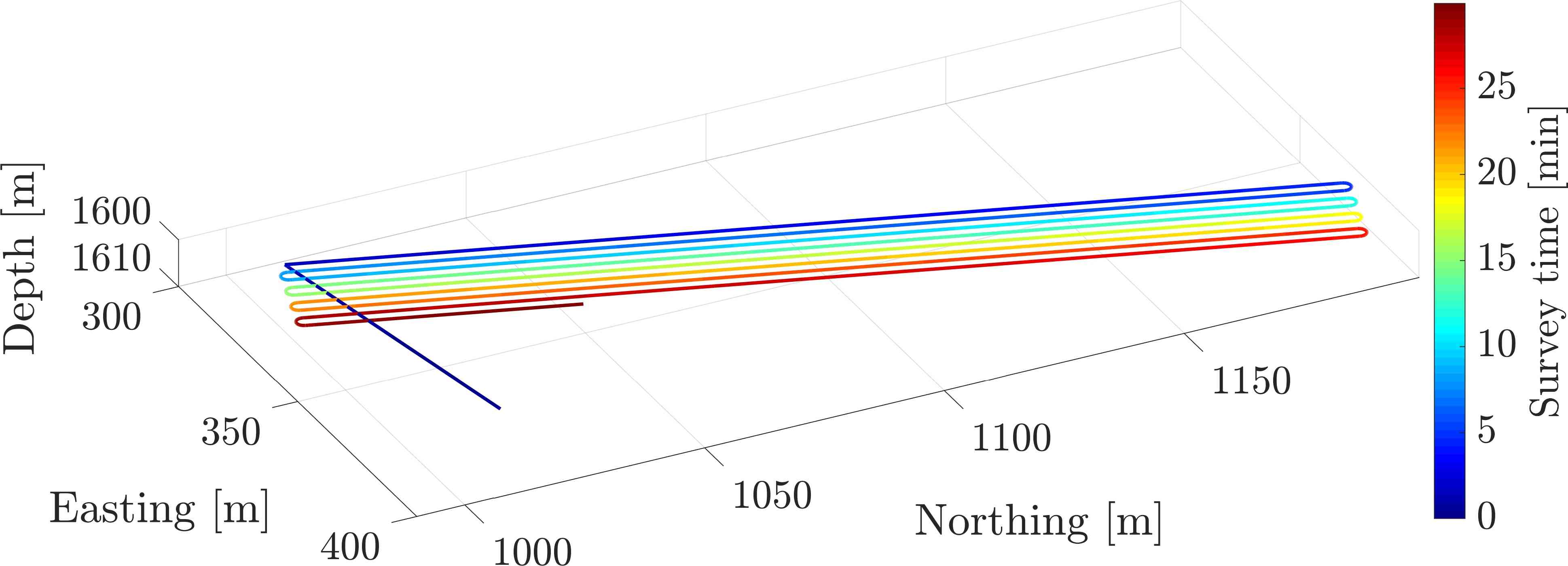}
	\caption{The simulated dataset.  A single tie-line is intersected on the
	left by multiple ``lawnmower'' inspection passes.}
	\label{fig:sonardyne_sec}
    \vspace{-4pt}
\end{figure}

\begin{figure*}[h!b]
	\sbox\subfigbox{%
	  \resizebox{\dimexpr0.89\textwidth-1em}{!}{%
		\includegraphics[height=3cm]{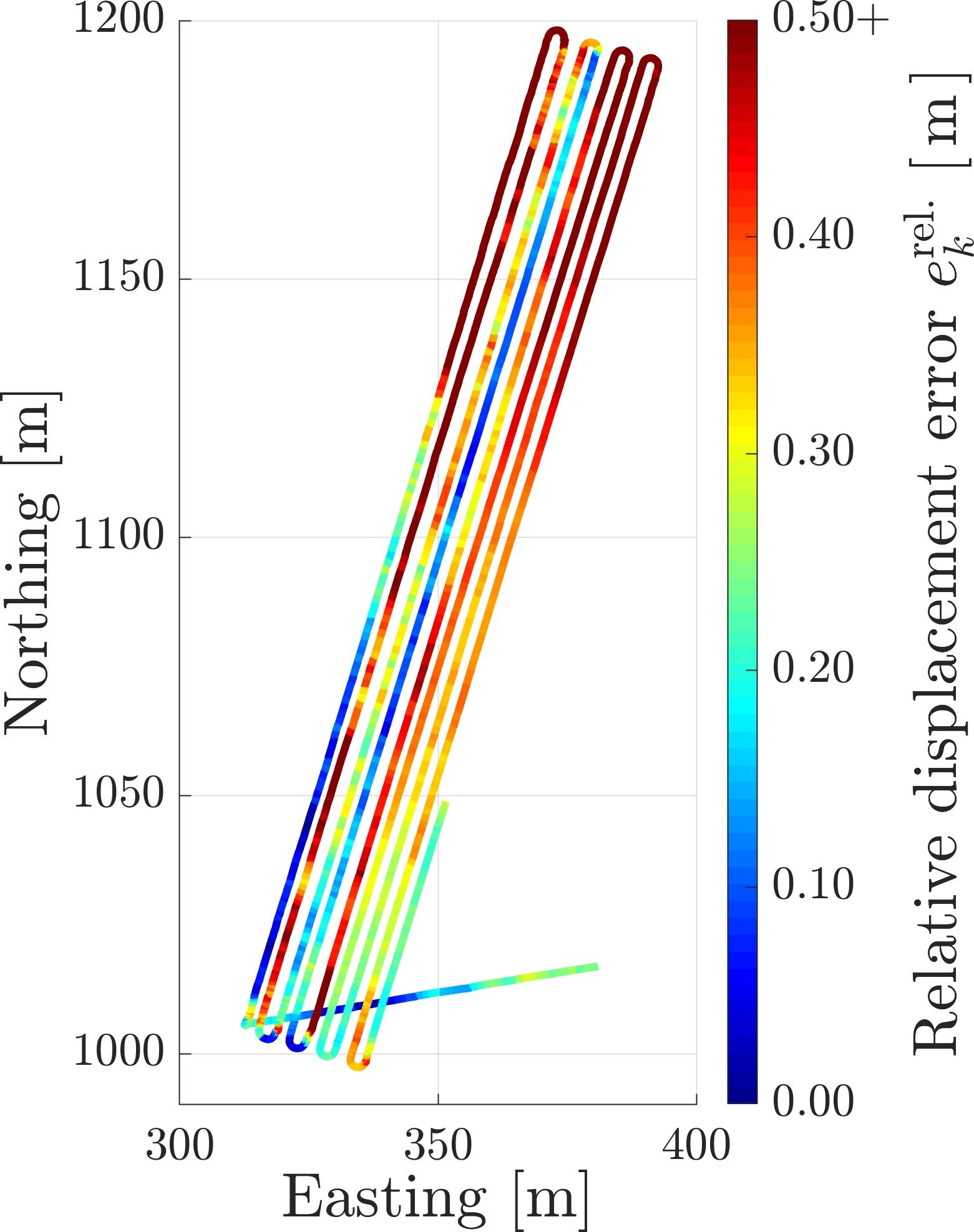}%
		\includegraphics[height=3cm]{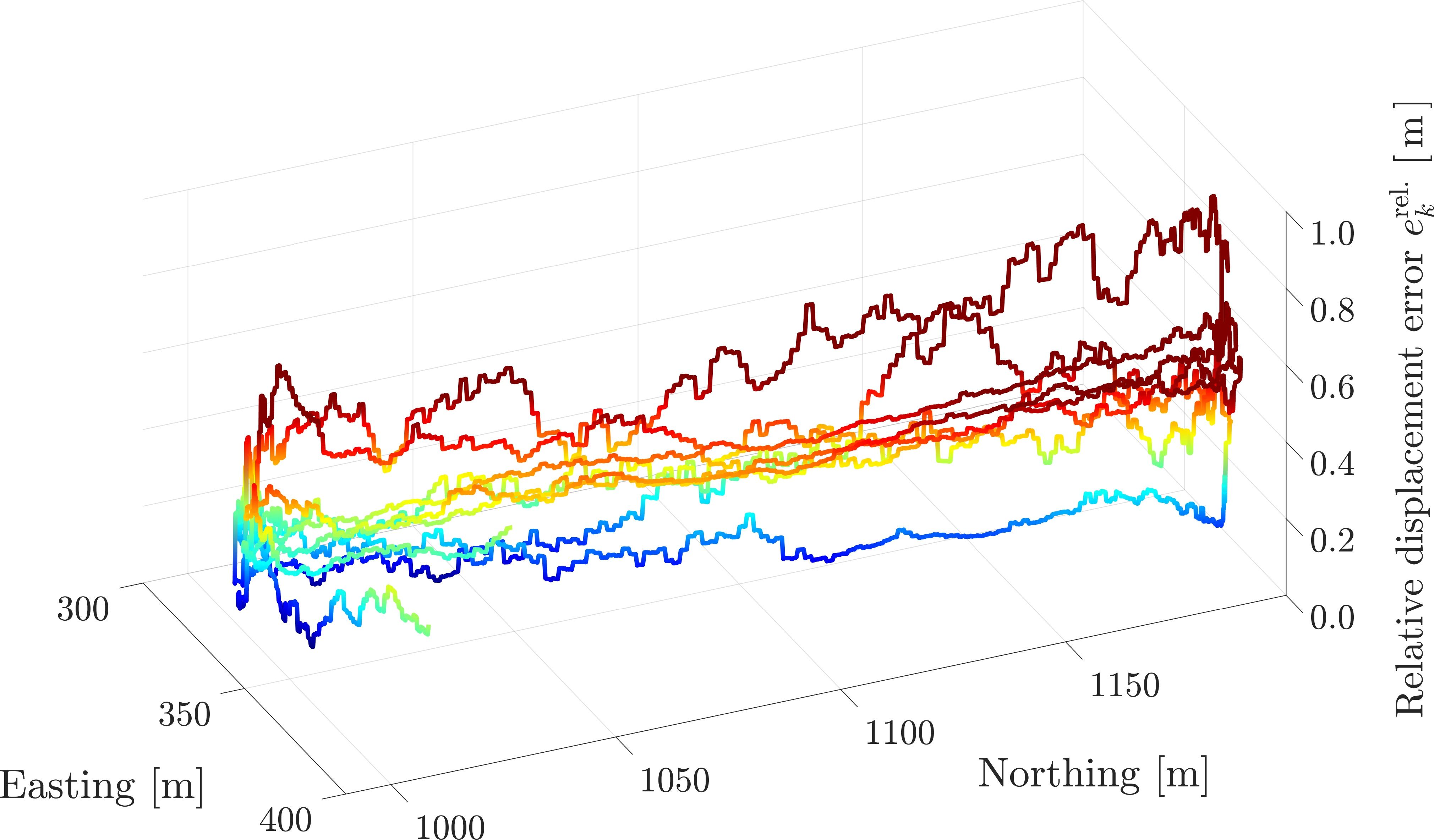}%
	  }%
	}
	\setlength{\subfigheight}{\ht\subfigbox}
	\centering
	\subcaptionbox{INS top \label{fig:sd_heat_prior_top}}{%
	  \includegraphics[height=\subfigheight]{figs/updated/small/sd_heatmap_top_prior.jpg}
	}
    \hspace{3pt}
	\subcaptionbox{INS isometric.  \update{Heatmap scale is identical to \Cref{fig:sd_heat_prior_top}.} \label{fig:sd_heat_prior_iso}}{%
	  \includegraphics[height=\subfigheight]{figs/updated/small/sd_heatmap_iso_prior.jpg}
	}
    \par \bigskip
    \subcaptionbox{INS+LC top \label{fig:sd_heat_posterior_top}}{%
	  \includegraphics[height=\subfigheight]{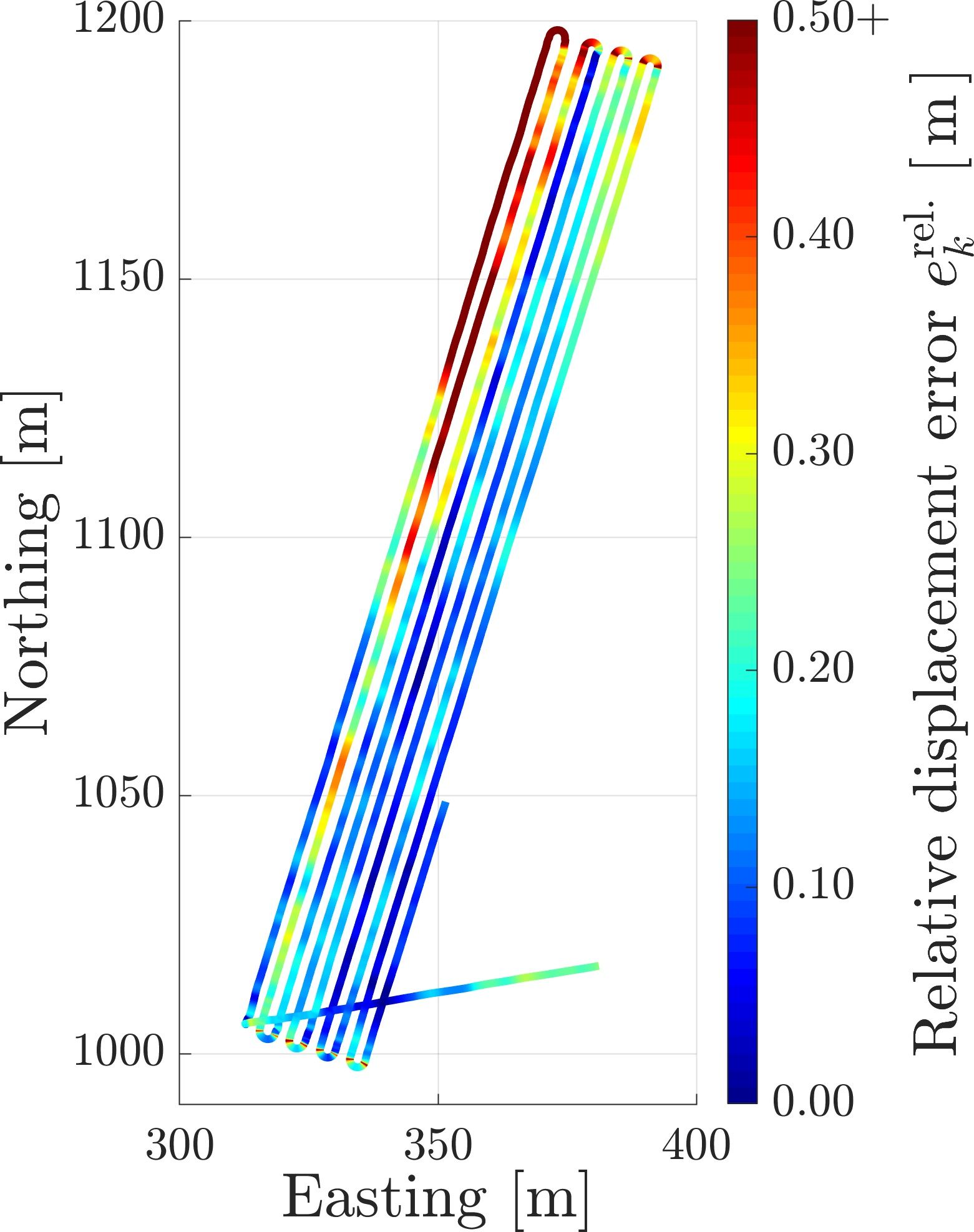}
	}
    \hspace{3pt}
	\subcaptionbox{INS+LC isometric.  \update{Heatmap scale is identical to \Cref{fig:sd_heat_posterior_top}.} \label{fig:sd_heat_posterior_iso}}{%
	  \includegraphics[height=\subfigheight]{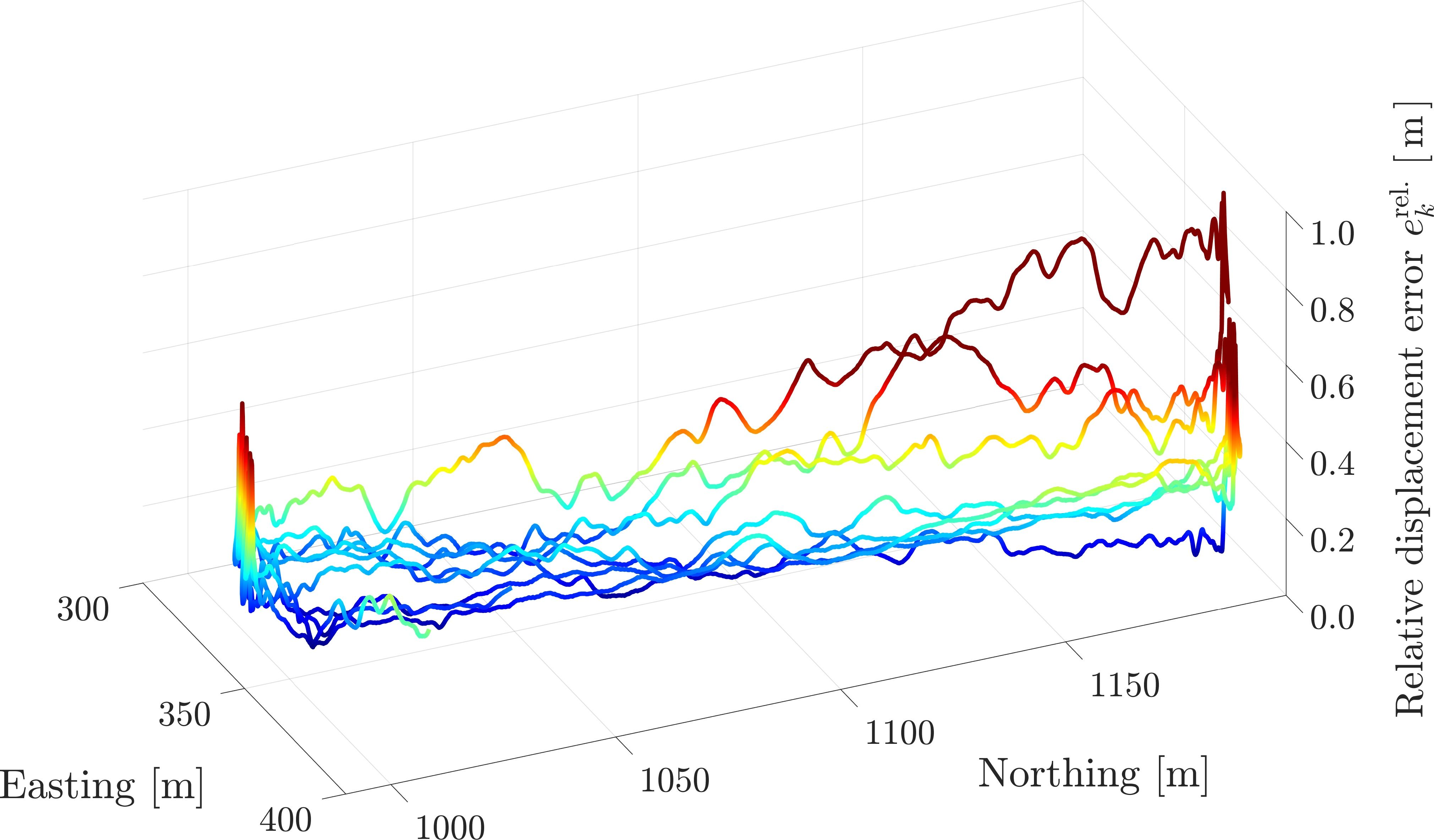}
	}
	\caption{Heatmaps representing the relative displacement error for the INS
	and INS+LC trajectories.  The left column shows a top view, while the right
	column shows an isometric view to illustrate the smoothing effects of the
	WNOA terms.  Note the series of spikes at the far ends of
	\Cref{fig:sd_heat_posterior_iso}, where the vehicle completes low-radius
	turns to initiate the next survey pass.}
    \label{fig:heatmapfig}
\end{figure*}

Simulated output from a DVL-INS and a corresponding ground-truth trajectory were
provided by industry collaborator Sonardyne.  The DVL-INS output contains
latitude and longitude, depth, and roll-pitch-yaw estimates for a simulated AUV
deployment.  At each time step, marginal variance estimates are available for
the depth and heading states, and a joint covariance estimate is available for
planar position in the local geodetic frame.  \SI{30}{\minute} of DVL-INS output
data is available at a frequency of \SI{5}{\Hz}.  Note that the DVL-INS output
has been heavily degraded by Sonardyne to better assess the ability of loop
closures to mitigate navigation drift and does not reflect the performance of
Sonardyne commercial products.

\begin{figure*}[h!b]
    \sbox\subfigbox{%
        \resizebox{\dimexpr0.995\textwidth-1em}{!}{%
            \includegraphics[height=5cm]{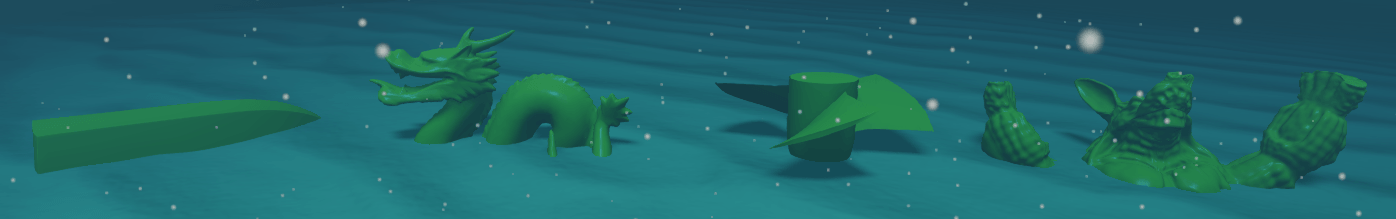}%
        }%
    }
	\setlength{\subfigheight}{\ht\subfigbox}
	\centering
    \subcaptionbox{An underwater scene generated  in the Stonefish AUV simulator
    \cite{Cieslak2019}.  From left to right: boat hull, dragon, propeller, and
    armadillo.  The hull and propellor are available in Stonefish, while the
    dragon and armadillo are from the Stanford 3D Scanning Repository
    \cite{Curless1996}.  \label{fig:sf_screengrab}}{%
		\includegraphics[height=\subfigheight]{figs/sf_grab_cropped_2_small.png}
	}
    \sbox\subfigbox{%
    \resizebox{\dimexpr0.97\textwidth-1em}{!}{%
        \includegraphics[height=3cm]{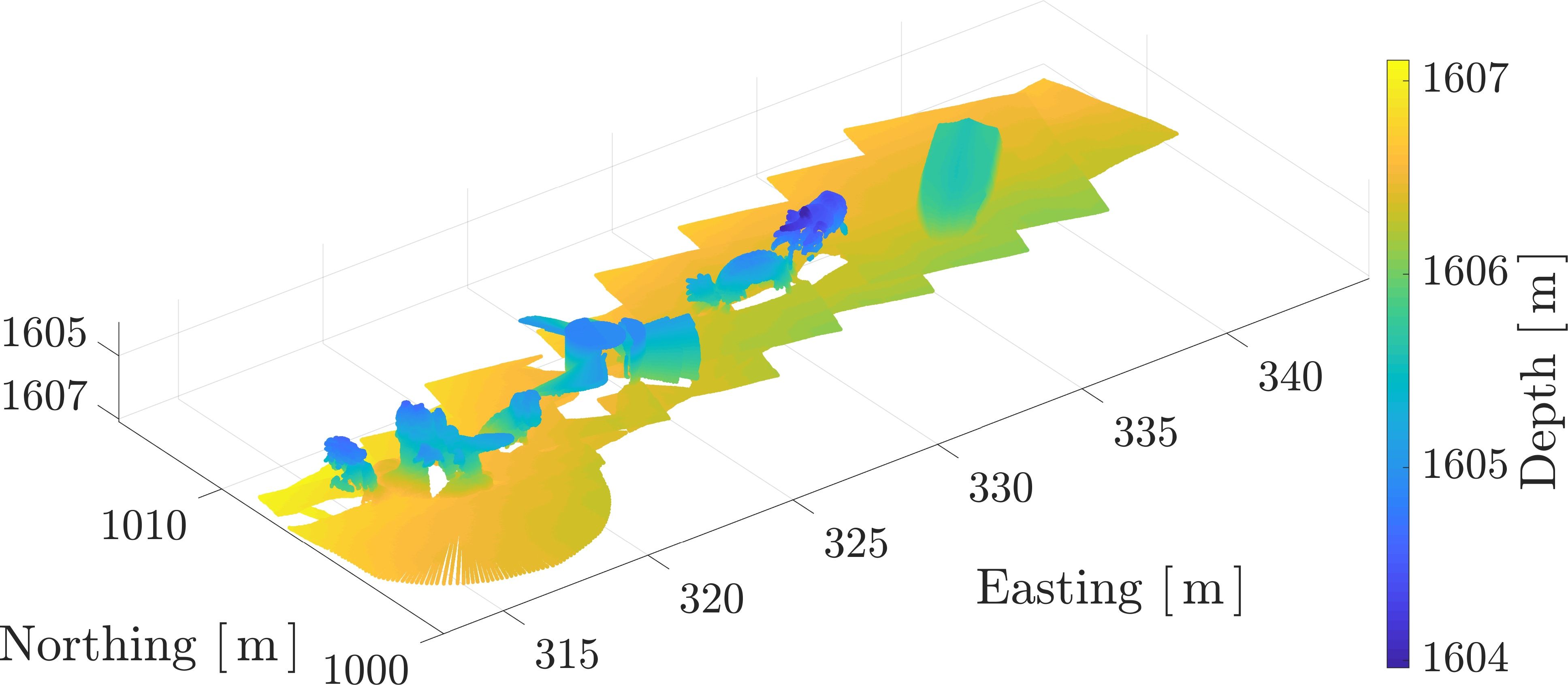}%
        \includegraphics[height=3cm]{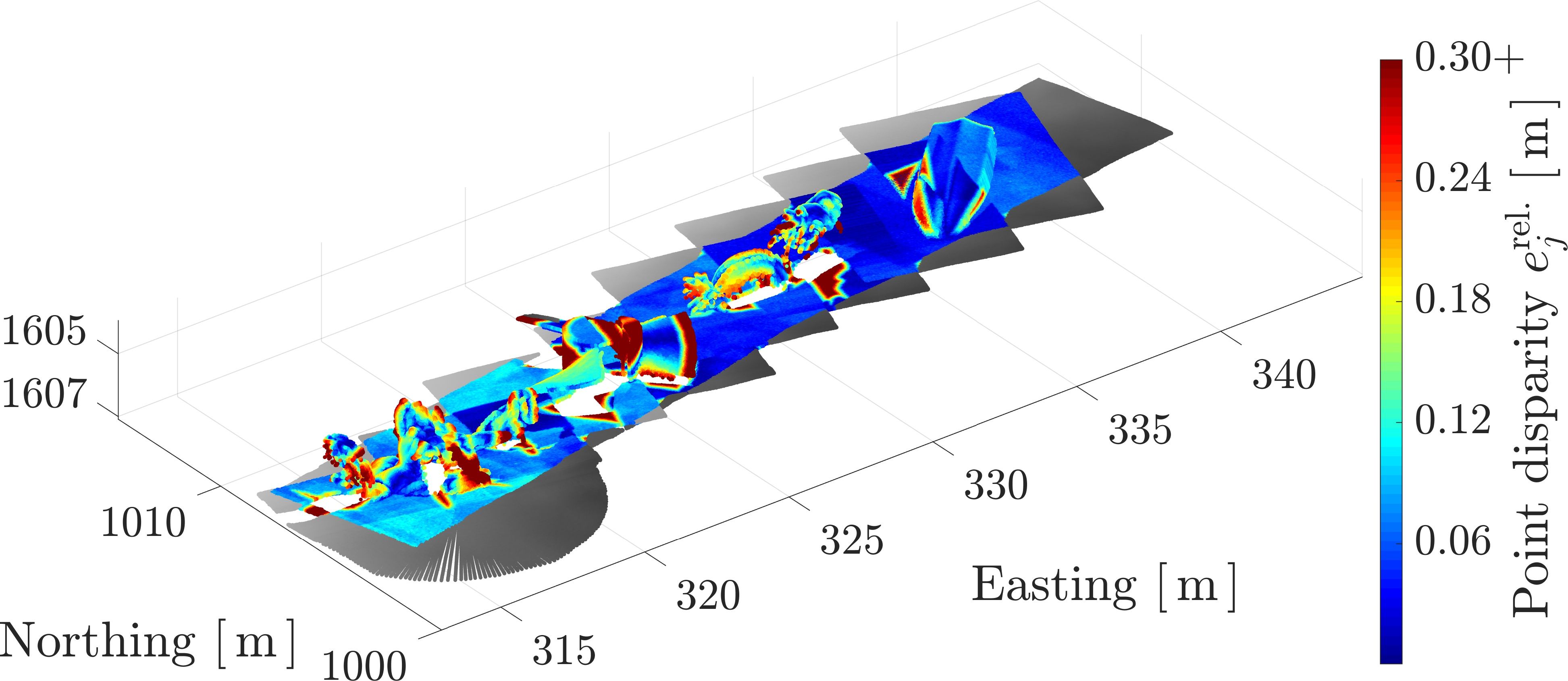}%
    }%
    }
    \setlength{\subfigheight}{\ht\subfigbox}
    \par\bigskip
    \subcaptionbox{Prior elevation (INS) \label{fig:sf_elevation_prior}}{%
        \includegraphics[height=\subfigheight]{figs/updated/small/elevation_prior_sd.jpg}
    }	
    \hspace{3pt}
    \subcaptionbox{Prior disparity (INS)
    \label{fig:sf_disp_prior}}{%
        \includegraphics[height=\subfigheight]{figs/updated/small/disp_prior_sd.jpg}
    }
    \par\medskip
    \subcaptionbox{Posterior elevation (INS+LC) \label{fig:sf_elevation_posterior}}{%
    \includegraphics[height=\subfigheight]{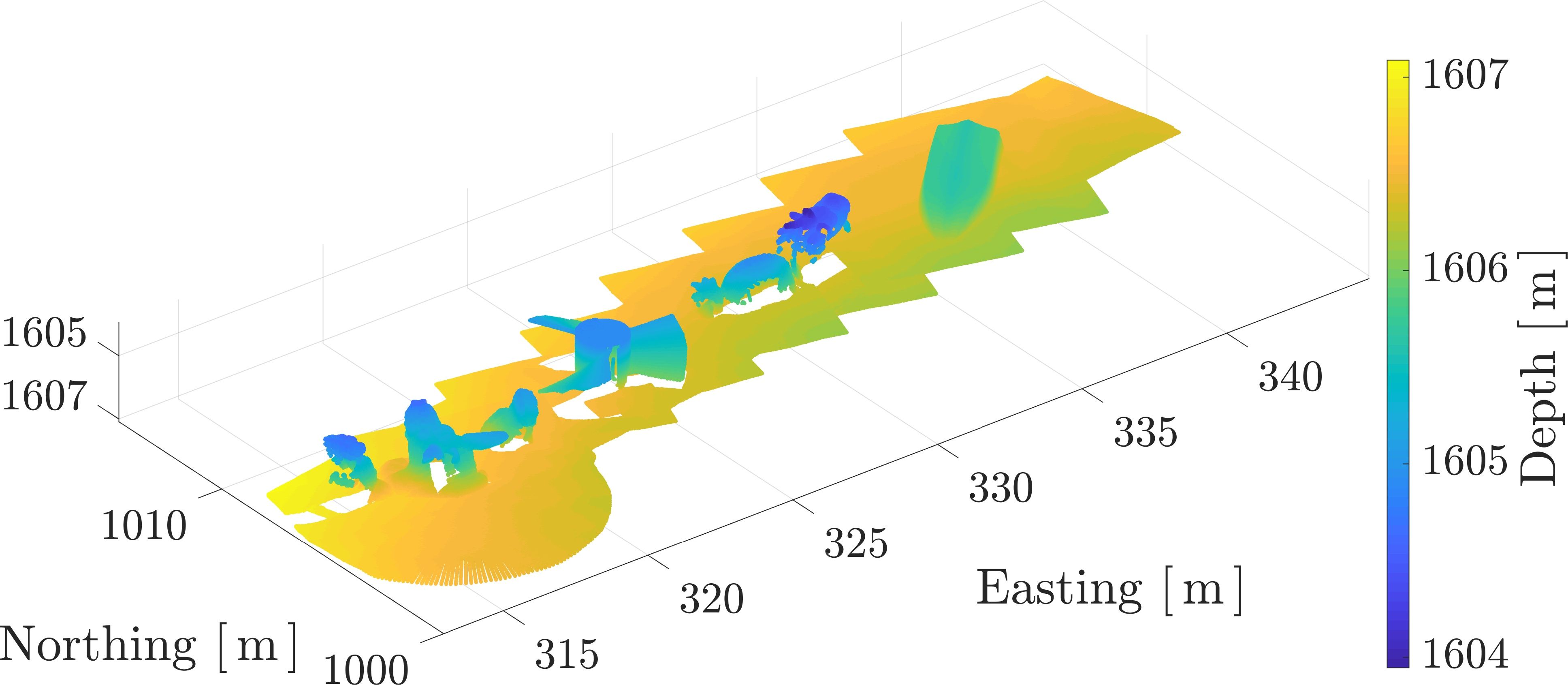}
    }
    \hspace{3pt}
    \subcaptionbox{Posterior disparity (INS+LC) \label{fig:sf_disp_posterior}}{%
    \includegraphics[height=\subfigheight]{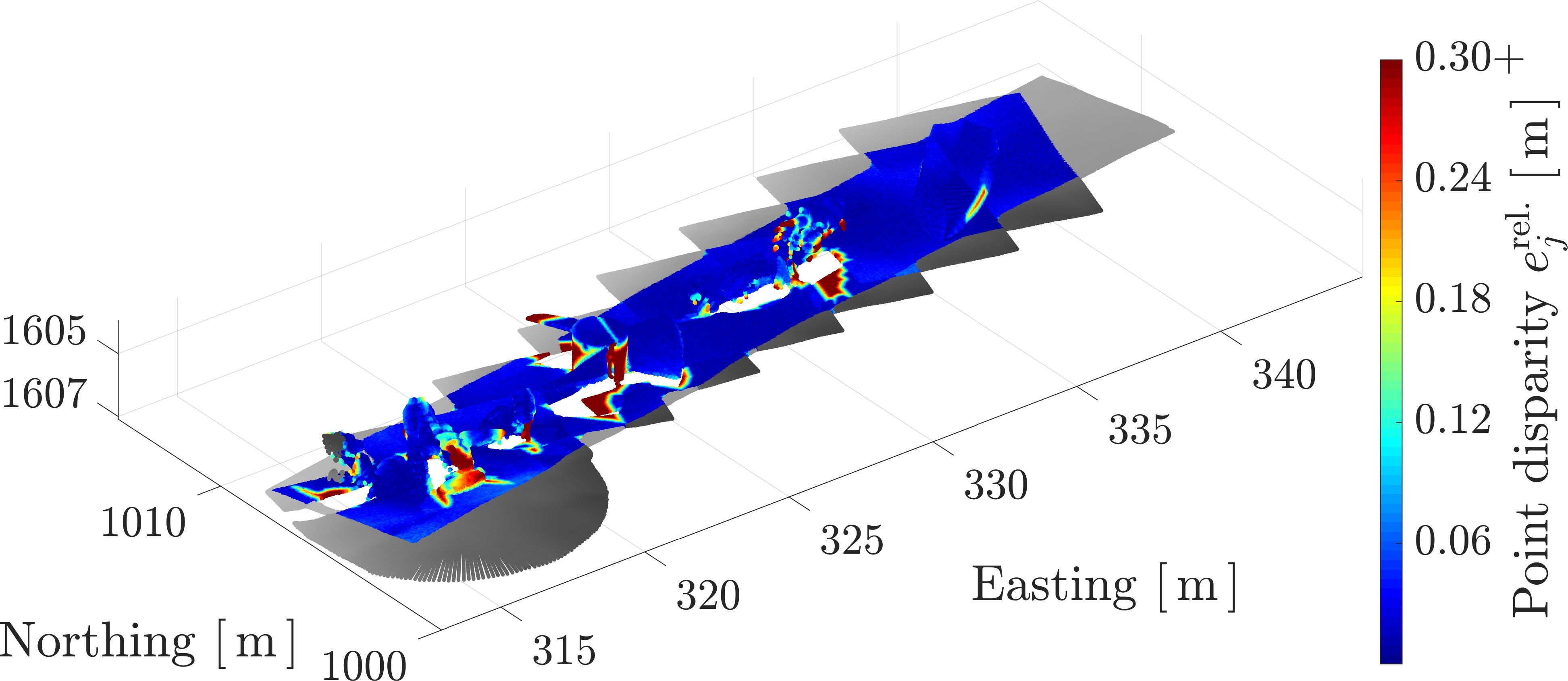}
    }
    \par\medskip
    \subcaptionbox{Ground-truth elevation (INS+GPS)
    \label{fig:sf_elevation_gt}}{%
    \includegraphics[height=\subfigheight]{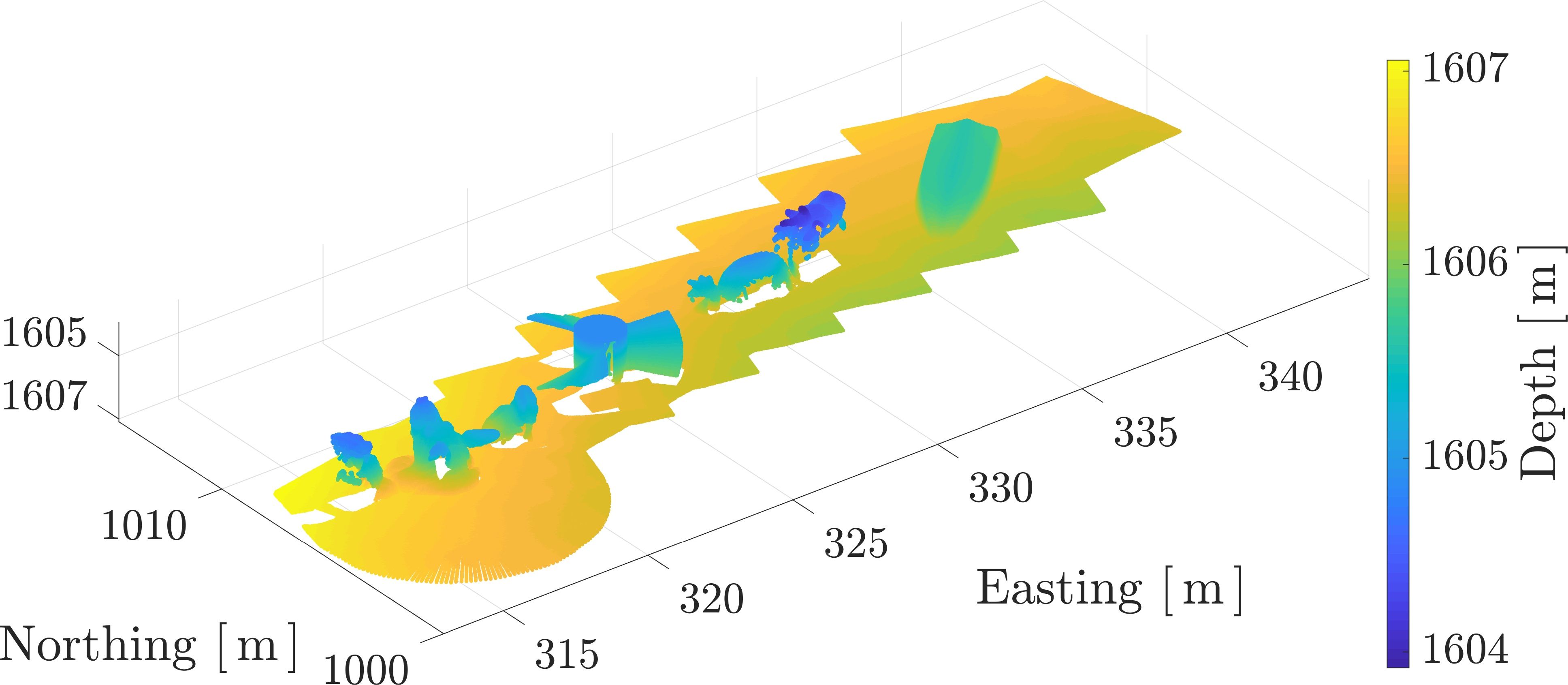}
    }
    \hspace{3pt}
    \subcaptionbox{Ground-truth disparity (INS+GPS)
    \label{fig:sf_disp_gt}}{%
        \includegraphics[height=\subfigheight]{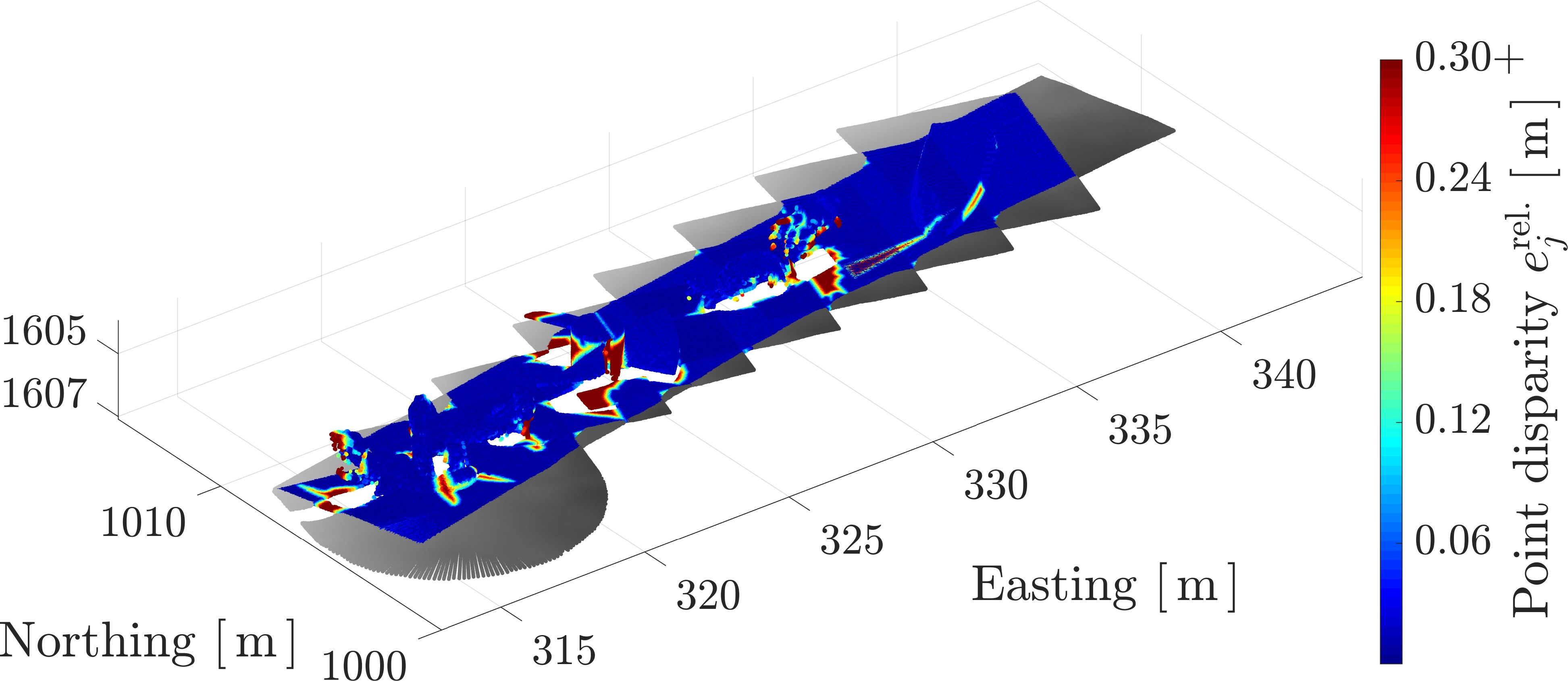}
    }
    \caption{Scanning a 3D scene in the Stonefish AUV simulator
    \cite{Cieslak2019} to evaluate the quality of the point cloud map.  The four
    models in \Cref{fig:sf_screengrab} are positioned along the tie-line near
    the loop-closure locations, and are scanned multiple times as the AUV
    completes the trajectory.  Errors in the trajectory estimate are easily seen
    in the point disparity maps in the right column.  Note the areas of high
    point disparity in the ground-truth map \Cref{fig:sf_disp_gt} are due to
    occlusion of the range-bearing scanner.  Scanning the 3D models at different
    orientations produces different occlusion patterns, leading to
    non-overlapping areas of the map.  This in turn leads to large
    nearest-neighbour distances to points in other scans, and thus a high
    point disparity error.}
	\label{fig:sf_elevationdisparitymaps}
\end{figure*}
The ground-truth trajectory is shown in \Cref{fig:sonardyne_sec}, where the
vehicle starts with a single tie-line followed by multiple planar ``lawnmower''
passes over a large inspection area.  Loop closures occur at the eight
intersections between the lawnmower passes and the tie-line.  The prior ``INS''
estimate is then conditioned on the loop-closure measurements using the
methodology from \Cref{sec:methodology} to produce a posterior ``INS+LC''
trajectory estimate.  Both estimates are then compared to the ground-truth
solution (``GT'') using the metrics from \Cref{sec:measuringselfconsistency}.
The application and propagation of loop-closure measurements within the WNOA
framework is expected to produce a more accurate navigation solution with a
correspondingly more self-consistent point cloud map.

To illustrate the improvements in accuracy and the smoothing effect from the
WNOA terms, \Cref{fig:heatmapfig} displays the relative displacement error
$e^{\textrm{rel.}}_k$ as a ``heatmap'' for the prior INS and posterior INS+LC
trajectories.  From \Cref{fig:sd_heat_prior_top} it is clear that the prior
estimate is not accurate, with relative displacement errors exceeding
\SI{0.5}{\meter} for much of the trajectory.  Incorporating loop-closure
measurements improves the accuracy of the trajectory estimate, as demonstrated
by the cooler colours throughout the heatmap in
\Cref{fig:sd_heat_posterior_top}.  The improvement is particularly noticeable
around the tie-line at the bottom of \Cref{fig:sd_heat_posterior_top}, however
for many of the passes the effects extend \textit{hundreds of meters beyond  the
loop-closure location}, increasing the accuracy across the entire inspection
area.  Relative displacement errors of \SI{0.5}{\meter} may seem inconsequential
at this scale, but may prove critical for certain subsea activities such as
jumper pipe installation. 

\vfill

\begin{figure}[h!tb]
	\centering
	\begin{subfigure}[t]{\columnwidth}
		\includegraphics[width=\linewidth]{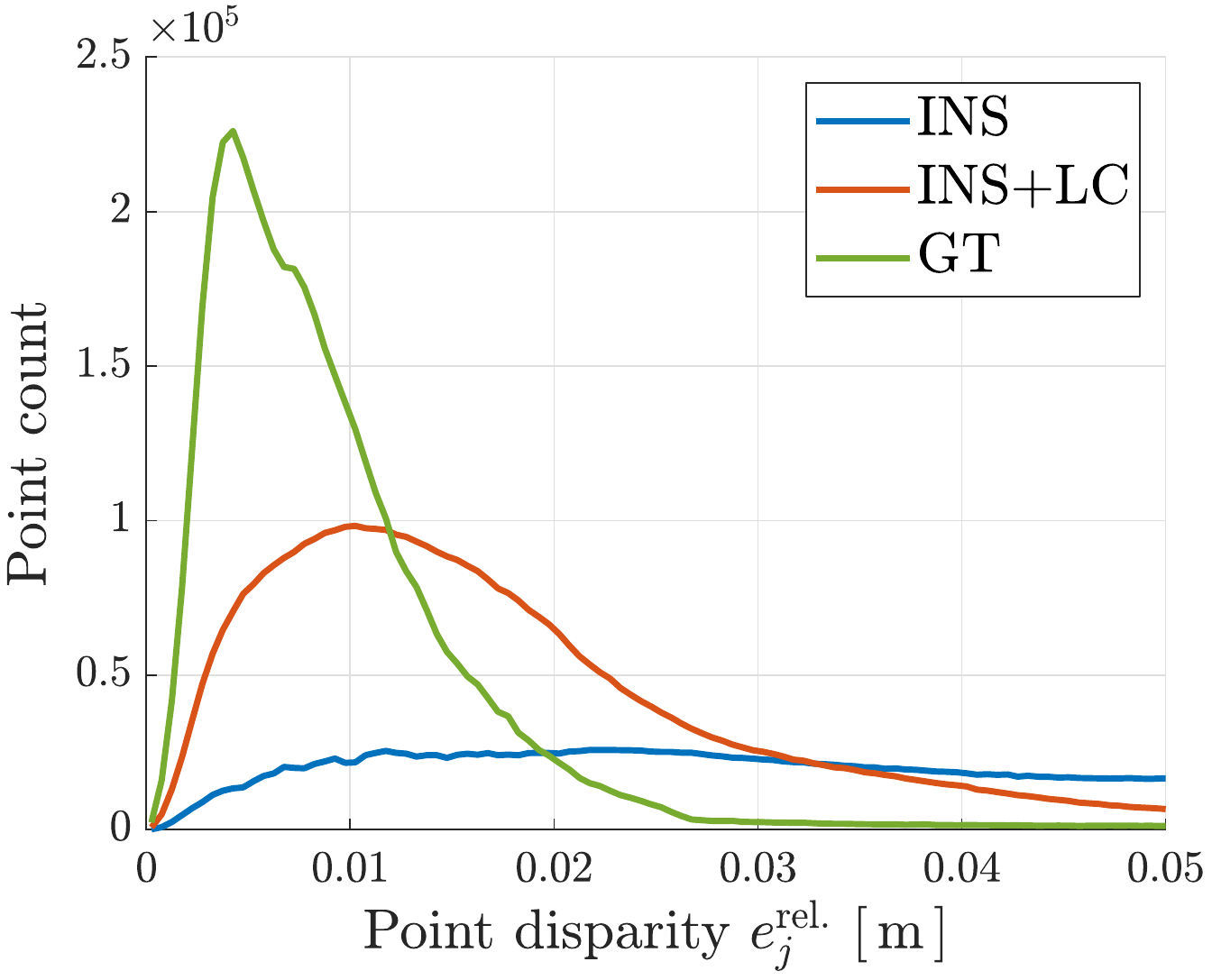}
		\caption{Empirical probability density functions (EPDF)}
        \vspace{8pt}
		\label{fig:sf_pointhist}
	\end{subfigure}
	\begin{subfigure}[t]{\columnwidth}
		\includegraphics[width=\linewidth]{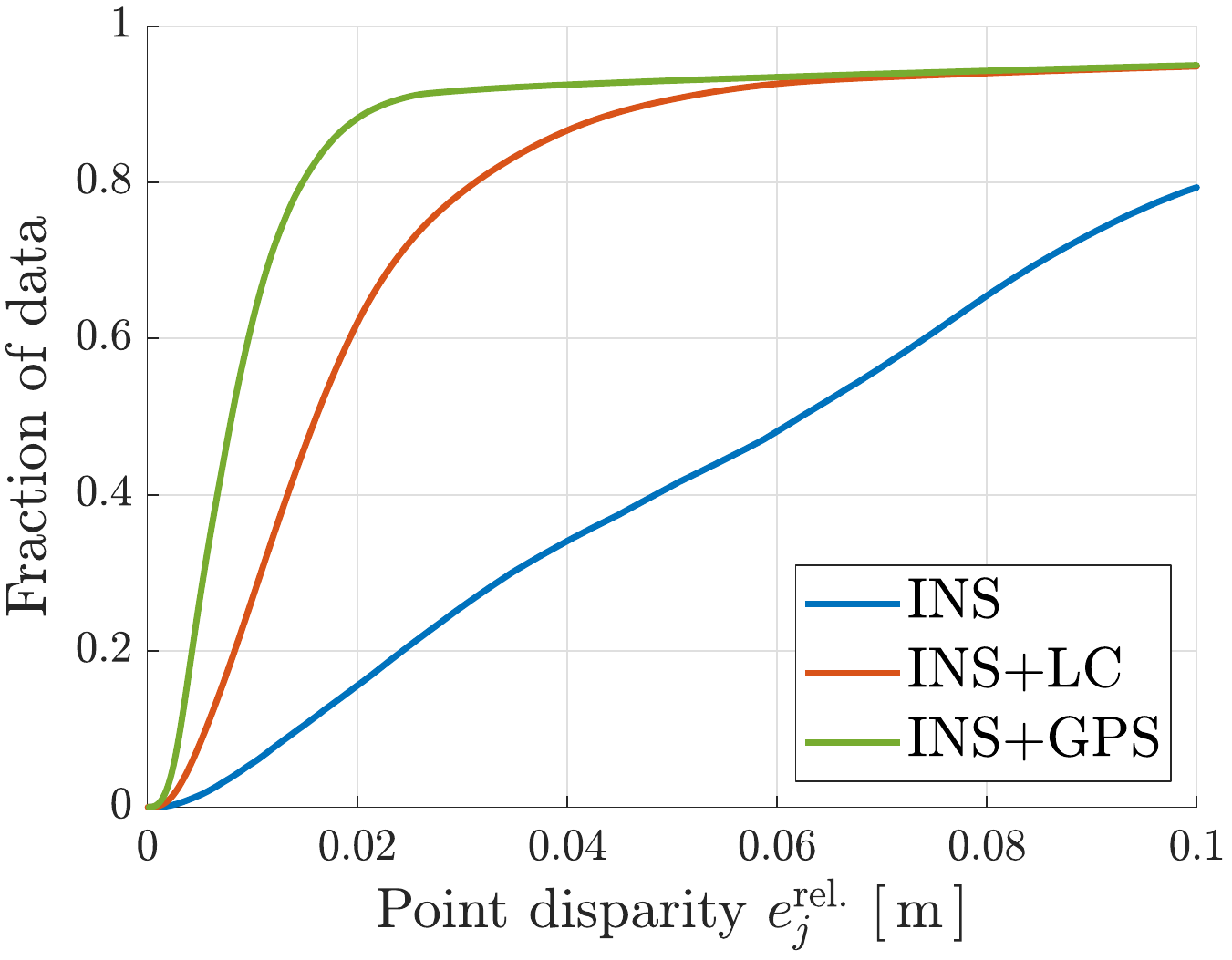}
		\caption{Empirical cumulative distribution functions (ECDF)}
		\label{fig:sf_pointecdf}
	\end{subfigure}
	\caption{Distributions on the point disparity error for each of the three
    navigation solutions for the simulated AUV area inspection dataset.  The
    INS+LC solution greatly improves on the prior INS solution, virtually
    eliminating errors beyond \SI{6}{\centi\meter}.  The relatively uniform
    error distribution for the INS solution is visible in the heatmap of
    \Cref{fig:sf_disp_prior}. }
	\label{fig:simpointerrordists}
    \vspace{-6pt}
\end{figure}

The smoothing effects from the WNOA motion prior are visible in
\Cref{fig:sd_heat_prior_iso,fig:sd_heat_posterior_iso}, which are, respectively,
isometric views of \Cref{fig:sd_heat_prior_top,fig:sd_heat_posterior_top}.
Here, the relative displacement error is represented both by the heatmap and the
plot elevation.  The prior INS estimate shows visible step changes in the
relative displacement error, which are characteristic of the correction step of
a filter and likely represent the effects of DVL measurements within the DVL-INS
estimation algorithm.  In contrast, the posterior INS+LC estimate has been
visibly smoothed due to the presence of the WNOA error terms.  Trajectory
smoothing is important in this context, as the point cloud map is generated by
registering individual laser profiles to the trajectory estimate.  A trajectory
with step changes will produce a map with step changes, which will surely impact
front-end activities such as feature detection and point cloud alignment.  A
smooth, self-consistent map is also visually appealing, and will improve the
accuracy of \update{subsea} metrology.

\update{Note the series of ``spikes'' in the relative displacement error of
\Cref{fig:sd_heat_posterior_iso}, corresponding to the posterior trajectory
estimate.  These spikes occur at the beginning and end of each low-radius turn,
suggesting that the smoothing effect of the WNOA terms may be erasing trajectory
information in high-curvature regions.  A geometry-based trajectory upsampling
approach, for example based on scale-invariant density \cite{Kurz2021}, is
expected to resolve this, and will be explored as part of future work.}

To evaluate the effects of the optimization on map quality, an underwater scene
was constructed and scanned using the open-source AUV simulator Stonefish
\cite{Cieslak2019}.  Raw laser profiles were collected along the ground-truth
trajectory, and were then registered to the INS and INS+LC trajectories to
produce, respectively, the prior and posterior point cloud maps.  The 3D scene
in Stonefish, as well as the resulting elevation and point disparity maps, are
shown throughout \Cref{fig:sf_elevationdisparitymaps}.  

\begin{table}[t]
    \centering
    \caption{\update{Critical values from the ECDF of \Cref{fig:sf_pointecdf}.
    For example, for the INS solution, \SI{95.45}{\percent} ($2\sigma$) of point
    disparity errors are below \SI{23.39}{\centi \meter}.  Occluded areas in the
    simulated point cloud scan are responsible for the large errors at the upper
    ends of the ECDF distributions.}}
    \renewcommand{\arraystretch}{1.2}
    \begin{tabularx}{\columnwidth}{XYYYY}
    \toprule
    Solution & \SI{50}{\percent} & $1\sigma$ & $2\sigma$ & $3\sigma$ \\
	\hline
		INS     & \SI{6.23}{\centi \meter} & \SI{8.32}{\centi \meter} &
		\SI{23.39}{\centi \meter} & \SI{56.15}{\centi \meter} \\
		INS+LC  & \SI{1.61}{\centi \meter} & \SI{2.27}{\centi \meter} &
		\SI{11.60}{\centi \meter} & \SI{44.70}{\centi \meter} \\
        INS+GPS & \SI{0.81}{\centi \meter} & \SI{1.13}{\centi \meter} &
		\SI{11.48}{\centi \meter} & \SI{45.64}{\centi \meter} \\
	\bottomrule
    \end{tabularx}
	\label{tab:sf_pointerrortable}
\end{table}

Compared to the ground-truth disparity map \Cref{fig:sf_disp_gt}, the prior map
\Cref{fig:sf_disp_prior} shows high point disparity errors throughout,
indicating a self-inconsistent map.  In contrast, the disparity errors are
largely resolved by the INS+LC solution, which incorporates both loop-closure
measurements and smoothing into the INS estimate.  The improvements are
quantified in \Cref{fig:simpointerrordists} by plotting an empirical probability
density function (EPDF) and an empirical cumulative distribution function (ECDF)
of the disparity error for each of the three solutions.  \update{For a
quantitative comparison, \Cref{tab:sf_pointerrortable} lists critical values
drawn from the ECDF curves.} The INS+LC solution improves on the INS solution by
producing a larger fraction of points with a lower point disparity error.  This
is especially evident in \Cref{fig:sf_pointecdf}, where the INS+LC solution
converges to the GT solution around \SI{6}{\centi\meter}.  Assuming the
remaining \SI{5}{\percent} of errors lie in occluded regions, as explained in
the caption of \Cref{fig:sf_elevationdisparitymaps}, the INS+LC solution has
effectively eliminated point disparity errors beyond \SI{6}{\centi\meter}.  In
contrast, \SI{20}{\percent} of disparity errors from the prior INS solution
exceed \SI{10}{\centi\meter}.  ``Double-vision'' effects of this magnitude
arising from poor scan alignment are sure to complicate inspection and metrology
tasks, even within the small domain of this simulation.  Following the
methodology from \Cref{sec:methodology}, the INS+LC solution has produced a
smooth, crisp, self-consistent point cloud map from which relative distance
measurements may accurately be drawn.

\subsection{Field Results: Wiarton Shipwreck}
\label{sec:fieldresults}

%

\begin{figure*}[h!tb]
	\sbox\subfigbox{%
	  \resizebox{\dimexpr0.94\textwidth-1em}{!}{%
		\includegraphics[height=3cm]{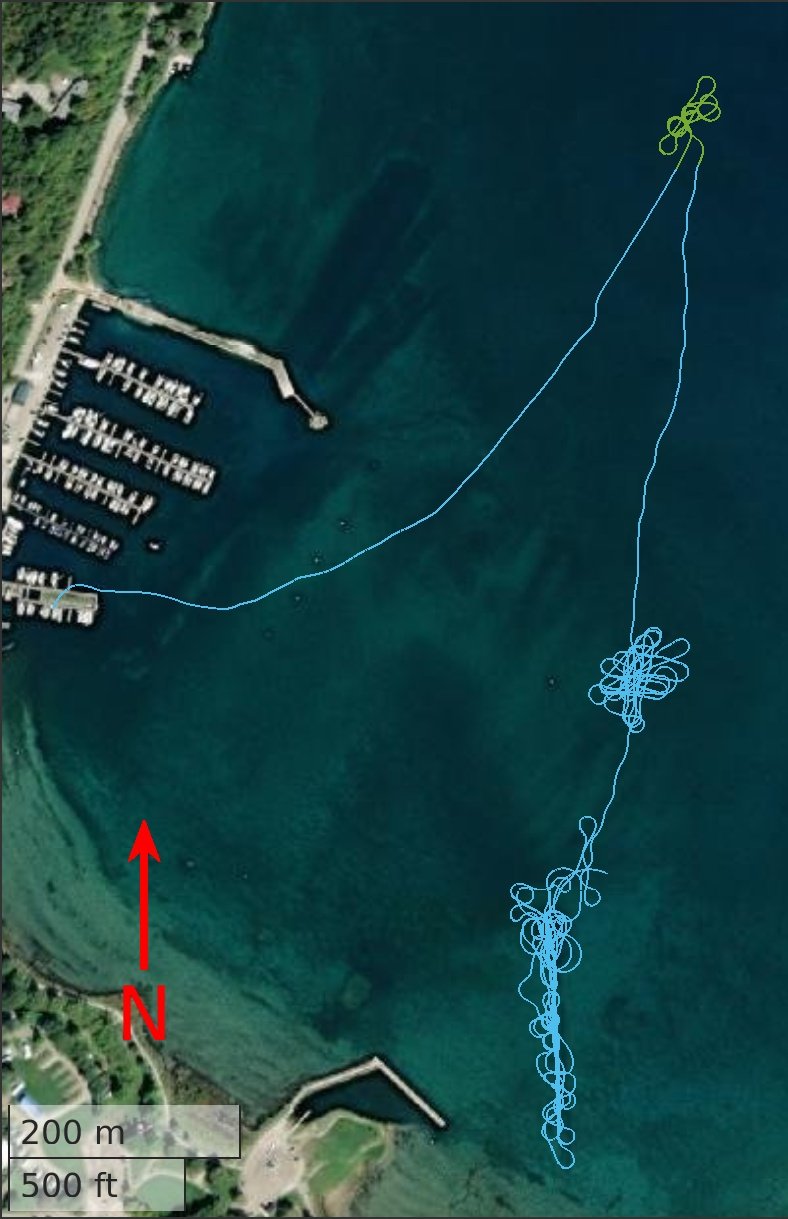}%
		\includegraphics[height=3cm]{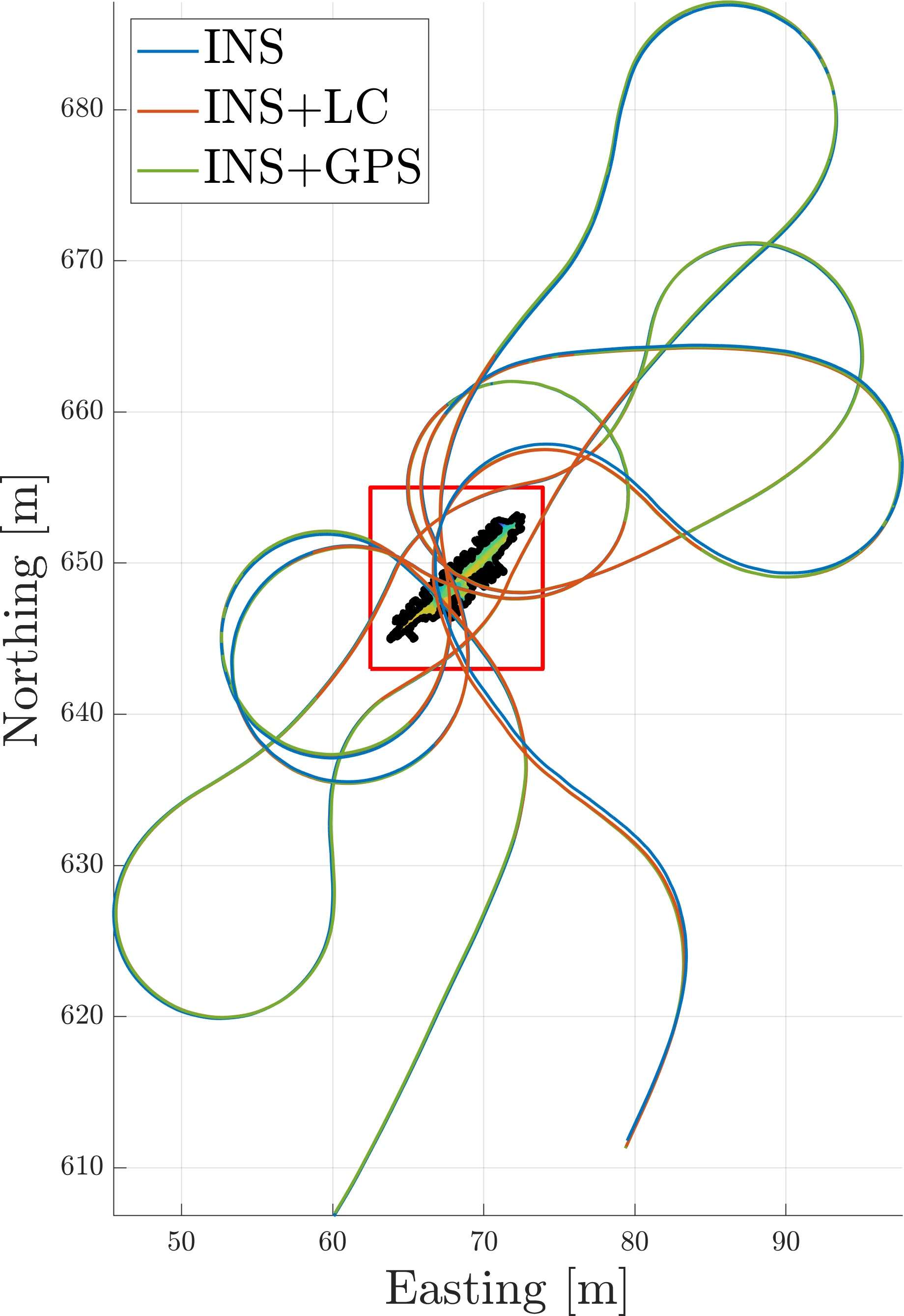}%
		\includegraphics[height=3cm]{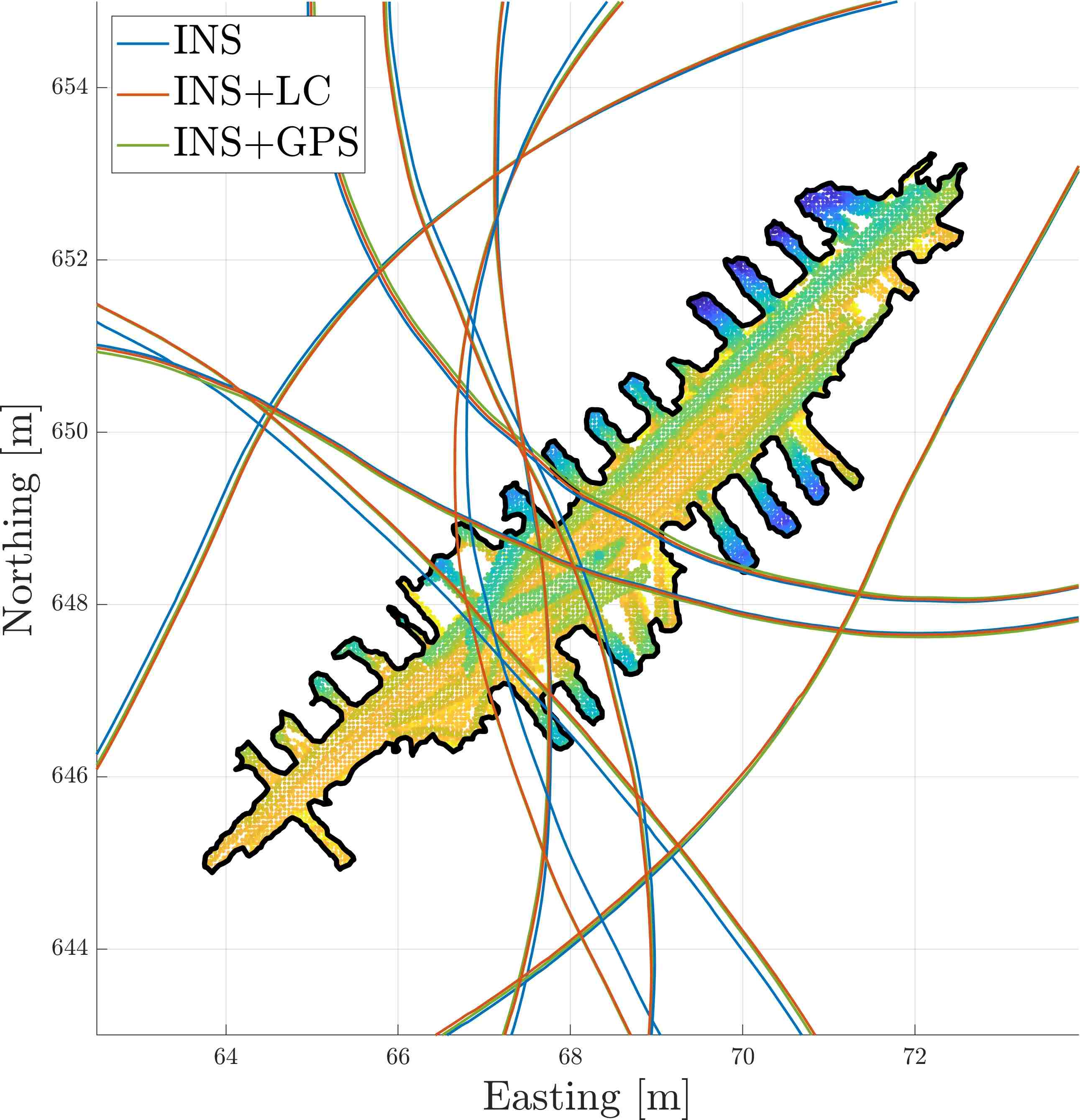}%
	  }%
	}
	\setlength{\subfigheight}{\ht\subfigbox}
	\centering
	\subcaptionbox{The test trajectory, with shipwreck section in green in
	northeast corner.  \label{fig:trajectory}}{%
	  \includegraphics[height=\subfigheight]{figs/trajectory_map_cropped.jpg}
	}
	\quad
	\subcaptionbox{Shipwreck section, with different trajectory estimates
	and shipwreck area. \label{fig:nav_solutions}}{%
	  \includegraphics[height=\subfigheight]{figs/trajectories_small.jpg}
	}
	\quad
	\subcaptionbox{The boxed region from \Cref{fig:nav_solutions}, showing the
	outline of the main shipwreck structure. \label{fig:nav_solutions_zoom}}{%
	  \includegraphics[height=\subfigheight]{figs/trajectories_zoom_small.jpg}
	} \caption{Field deployment in Colpoy's Bay, Wiarton, Ontario, Canada.  The
	full trajectory is shown in \Cref{fig:trajectory}.  The shipwreck
	section, shown in green in northeast corner of \Cref{fig:trajectory}, is
	\SI{0.58}{\kilo \meter} long and took approximately \SI{10.5}{\minute} to
	complete.  The trajectory makes eight passes over the shipwreck area.  The
	different navigation solutions are summarized in \cref{tab:solutiontable}. }
\end{figure*}

A field trial was conducted with Voyis Imaging Inc. in Colpoy's Bay, Wiarton,
Ontario, Canada.  The bay is shallow and contains multiple shipwrecks and other
manmade structures, making it an ideal test location for Voyis's surface vessel.
The full test trajectory is shown in blue in \Cref{fig:trajectory}.  A section
of the trajectory, highlighted in green in the northeast corner of
\Cref{fig:trajectory}, makes eight passes over a small shipwreck.
\Cref{fig:nav_solutions} shows this section in detail, and
\Cref{fig:nav_solutions_zoom} shows the main shipwreck structure segmented from
the lakebed.  This section of the trajectory, which is approximately
\SI{580}{\m} long and took \SI{10.5}{\minute} to complete, is the focus of the
field results. 

The surface vessel was equipped with a Sonardyne SPRINT-Nav 500 DVL-aided INS, a
u-blox ZED-F9P high precision GNSS module equipped with a u-blox ANN-MB series
high precision multi-band antenna, and a Voyis Insight Pro underwater laser
scanner.  GNSS data were post-processed using the Canadian Spacial Reference
System Precise Point Positioning (CSRS-PPP) application \cite{Tetreault2005},
which in a recent study was found to be capable of measuring 2D position with a
precision of \SI{2}{\centi \meter} ($1\sigma$) \cite{Alkan2020}.

Three navigation solutions were generated from these data.  The first solution is
a dead-reckoned DVL-INS trajectory (``INS''), where the positioning precision of
the SPRINT-Nav 500 has been manually degraded by Sonardyne from the nominal
value of \SI{0.02}{\percent} of distance traveled (CEP50)
\cite[Sec.~4.9.1.2]{Farrell2008}\cite{sprintnavdatasheet}.  The DVL-INS output
is available at \SI{10}{\Hz}.  \secondupdate{Note that this solution is
representative of the state-of-the-art for high-grade commercial systems, and
will be used to benchmark the proposed methodology.}  The second solution,
referred to as ``INS+LC,'' applies the methodology of \cref{sec:methodology} to
incorporate loop-closure measurements into the dead-reckoned DVL-INS estimate.
\secondupdate{Batch processing was performed offline in \textsc{MATLAB}, taking
approximately \SI{90}{\second} to converge on a laptop with an E3-1505M v5 CPU
and \SI{16}{\giga\byte} of RAM.}  It is important to note that this solution is
produced using the DVL-INS state \textit{estimate, without access to the raw
DVL-INS sensor measurements}.  The third solution fuses the DVL-INS output with
the GNSS data to form a ground-truth estimate (``INS+GPS'').  The three
navigation solutions are summarized in \cref{tab:solutiontable}, and the
trajectory estimates are overlaid on the shipwreck area in
\Cref{fig:nav_solutions,fig:nav_solutions_zoom}.

\begin{table}[tb]
    \centering
    \caption{Understanding the different navigation solutions for the Wiarton
    shipwreck field dataset.}
    \renewcommand{\arraystretch}{1.5}
    \begin{tabularx}{\columnwidth}{lX}
    \toprule
    Solution & Description \\
	\hline
		INS     & Dead-reckoned DVL-INS solution, with position precision
		manually degraded by Sonardyne.  \\
		INS+LC  & The dead-reckoned DVL-INS trajectory estimate conditioned on
		loop-closure measurements.  Raw sensor measurements from the DVL-INS are
		\textbf{inaccessible}, and GNSS data is \textbf{not} used as part of
		this solution.  \\
        INS+GPS & DVL-aided INS solution with GNSS correction.  \\
	\bottomrule
    \end{tabularx}
	\label{tab:solutiontable}
\end{table}

\begin{figure}[t]
    \centering
    \includegraphics[width=\columnwidth]{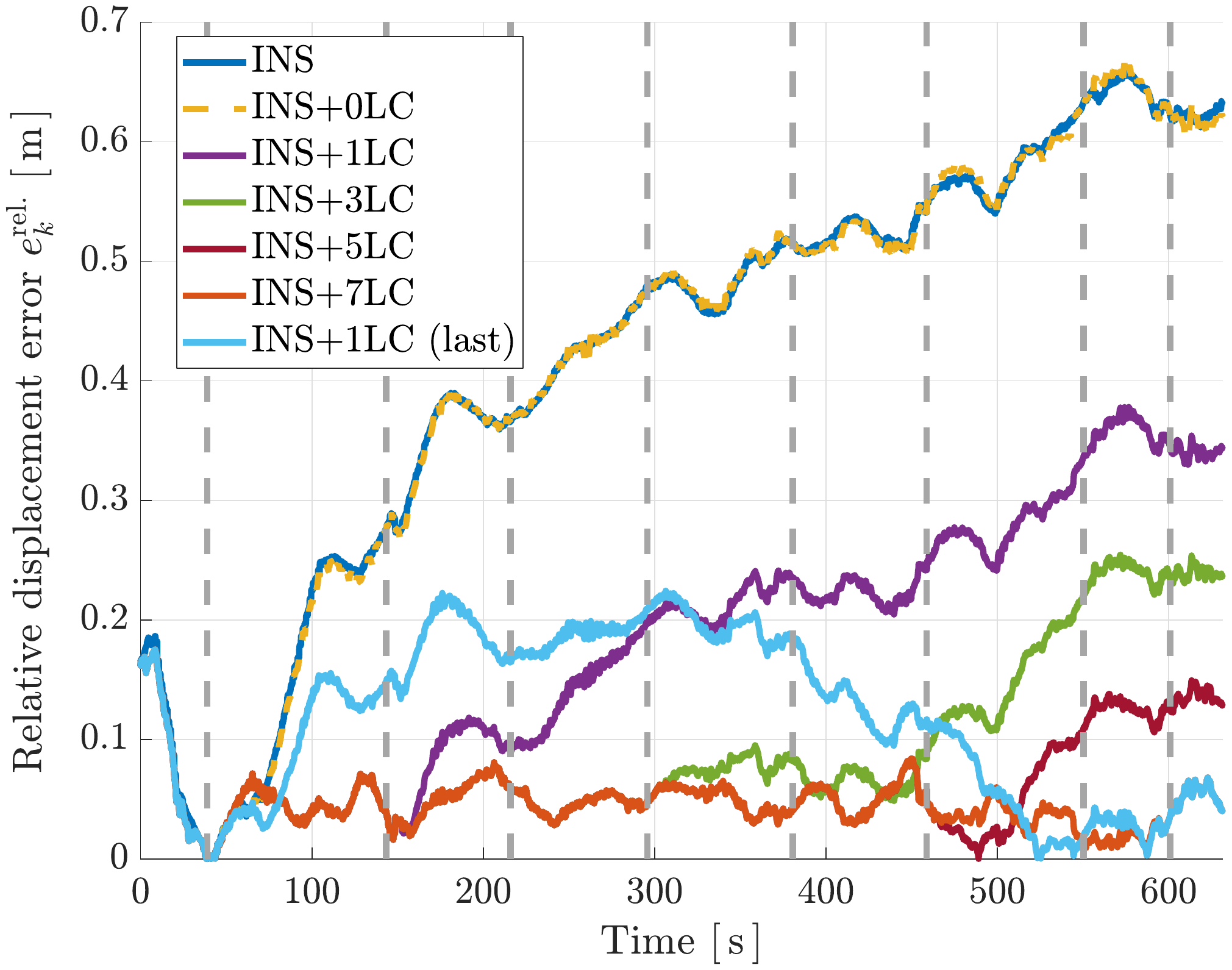}
    \caption{Relative displacement errors for trajectory estimates incorporating
    an increasing number of loop closures. Loop-closure locations are marked as
    vertical dashed lines, and ``INS+XLC'' indicates the first X loop closures
    were used in generating the estimate.  Incorporating loop closures bounds
    navigation drift over time.  For numerical results, see
    \Cref{tab:drifttable}.}
    \label{fig:drift_plot}
\end{figure}

\setcounter{table}{6}
\begin{table*}[b]
    \centering
    \caption{\secondupdate{Summary statistics on relative pose errors \eqref{eqn:posemetric}
    computed for the Wiarton field trial.  Relative attitude errors $| \delta
    \phi^\textrm{rel.}_i |$ have been broken down by component, while the
    rightmost column gives statistics on the norm of the relative body-centric
    position error.  Cumulative statistics for each error are reported in the
    format}
    \SI{50}{\percent}~$\cdot$~\colour{dark_grey}{\SI{75}{\percent}}~$\cdot$~\colour{light_grey}{\SI{90}{\percent}}.}
    \renewcommand{\arraystretch}{1.2}
    \begin{tabularx}{\textwidth}{lYYYY}
    \toprule
    Solution & $| \delta \phi^{\textrm{rel.}}_1 | \, \left[ \deg \right]$ & $| \delta \phi^{\textrm{rel.}}_2 |
    \, \left[ \deg \right]$ & $| \delta \phi^{\textrm{rel.}}_3 | \, \left[ \deg \right]$ & $\|
    \delta \mbs{\rho}^{\textrm{rel.}} \| \, \left[ \meter \right]$ \\
	\hline
        INS     & \tabentry{1.1e-3}{1.9e-3}{2.5e-3} & \tabentry{1.5e-3}{2.4e-3}{3.5e-3} & \tabentry{1.5e-2}{1.9e-2}{2.2e-2} & \tabentry{0.482}{0.562}{0.630} \\
        INS+LC  & \tabentry{1.1e-1}{2.7e-1}{5.3e-1} & \tabentry{1.2e-1}{3.8e-1}{5.7e-1} & \tabentry{1.5e-2}{2.0e-2}{2.2e-2} & \tabentry{0.077}{0.096}{0.108} \\
	\bottomrule
    \end{tabularx}
	\label{tab:wiartonposeerrortable}
\end{table*}

{Incorporating loop-closure measurements produces a more accurate trajectory
estimate, as measured by the relative displacement error
\eqref{eqn:displacementmetric}.  \Cref{fig:drift_plot} shows the relative
displacement error, measured against the ground-truth INS+GPS estimate, for the
dead-reckoned INS trajectory and the INS+LC trajectory with an increasing number
of loop closures.  The loop-closure locations are marked with vertical dashed
lines, with the first observation of the shipwreck occurring approximately
\SI{40}{\second} in to the trajectory.  The relative displacement drift in the
INS trajectory estimate increases without bound, while the maximum displacement
error decreases monotonically as more loop-closure measurements are applied.
Even a single loop-closure measurement at the end of the trajectory is effective
in bounding the relative displacement drift over time, as demonstrated by the
cyan line in \Cref{fig:drift_plot}.  From \cref{tab:drifttable}, which
summarizes the maximum error and final error as a percent of distance traveled
for the different solutions, the final drift error for the ``INS+1LC (last)''
solution is \SI{6.82e-3}{\percent} of distance traveled. This particular
solution suggests an order of magnitude improvement over state-of-the-art
DVL-INS systems \cite{sprintnavdatasheet}.  \secondupdate{Importantly, the
dashed yellow ``INS+0LC'' curve in \Cref{fig:drift_plot} indicates that the
posterior solution does not deviate far from the prior DVL-INS solution when
loop-closure measurements are absent.  This suggests that the proposed
methodology may still be used to smooth the DVL-INS solution in the absence of
loop-closure measurements, without sacrificing solution accuracy.  For example,
the maximum position drift error for the ``INS-0LC'' solution tabulated in
\Cref{tab:drifttable} is only \SI{9}{\milli \meter}}\unskip\parfillskip 0pt
\par}

\setcounter{table}{5}
\begin{table}[H]
    \centering
    \caption{Summary of drift errors from \Cref{fig:drift_plot}, with values
    drawn \textit{after} the first shipwreck observation at \SI{40}{\second}.}
    \renewcommand{\arraystretch}{1.2}
    \begin{tabularx}{\columnwidth}{lYY}
    \toprule
    Solution & Max drift [\SI{}{\meter}] & Final \%DT \\
	\hline
        INS & 0.658 & \SI{10.98e-2}{} \\
        \colour{black}{INS+0LC} & 0.667 & \SI{10.82e-2}{} \\
        INS+1LC & 0.378 & \SI{5.99e-2}{} \\
        INS+3LC & 0.255 & \SI{4.12e-2}{} \\
        INS+5LC & 0.150 & \SI{2.25e-2}{} \\
        INS+7LC & 0.084 & \SI{6.83e-3}{} \\ 
        INS+1LC (last) & 0.224 & \SI{6.82e-3}{} \\
	\bottomrule
    \end{tabularx}
	\label{tab:drifttable}
\end{table}

\secondupdate{\noindent greater than the maximum drift observed in the ``INS''
solution.  However, note the proposed methodology is intended to be used in a
targeted fashion in situations where at least one loop-closure measurement is
available.}

\secondupdate{In addition to the relative displacement errors summarized in
\Cref{fig:drift_plot}, relative pose errors \eqref{eqn:posemetric} are computed
across the trajectory for the prior ``INS'' and posterior ``INS+LC'' solutions,
with summary statistics given in \Cref{tab:wiartonposeerrortable}.  Relative
attitude errors in \Cref{tab:wiartonposeerrortable} are decomposed into
body-centric roll, pitch, and yaw, while the rightmost column gives statistics
on the Euclidean norm of the body-centric relative position errors.
Interestingly, the proposed methodology has produced an increase in the relative
body-centric pitch and roll errors, from median values of \SI{1.1e-3}{\deg} and
\SI{1.5e-5}{\deg}, respectively, to \SI{1.1e-1}{\deg} and \SI{1.2e-1}{\deg},
respectively.  Relative body-centric yaw errors remain largely unchanged by the
proposed methodology, while trends in the relative body-centric position error
generally follow the trend of the relative displacement error plotted in
\Cref{fig:drift_plot}.  For example, \SI{90}{\percent} of body-centric position
errors for the prior ``INS'' solution fall below \SI{0.630}{\meter}, while the
corresponding value for the posterior ``INS+LC'' solution is \SI{0.108}{\meter}.  

An increase in roll and pitch errors may seem concerning, however the posterior
errors remain low and bounded.  Such errors were likely introduced in this field
trial through a combination of small angular errors in the INS-laser extrinsics
estimate and the relatively weak pitch and roll prior used in the optimization
(see \eqref{eqn:obserror} and the value of hyperparameter $\sigma_{\textrm{rp}}$
in \Cref{tab:hypparamtable}).  The more important result is that relative
body-centric position errors remain low and bounded when multiple loop-closure
measurements are present.}

Barring measurement outliers, finding that loop closures improve trajectory
accuracy is not particularly surprising in a conventional state estimation
context.  However, these results have been achieved following the methodology of
\Cref{sec:methodology}, \textit{without} access to raw sensor measurements, a
vehicle process model, exteroceptive sensor models, or sensor noise and bias
specifications.  The loop-closure corrections have instead been smoothly
integrated into the DVL-INS estimate using the factor graph illustrated in
\Cref{fig:factorgraph}, improving the accuracy of the trajectory estimate.  

\begin{figure*}[h!b]
	\sbox\subfigbox{%
	  \resizebox{\dimexpr0.96\textwidth-1em}{!}{%
		\includegraphics[height=5cm]{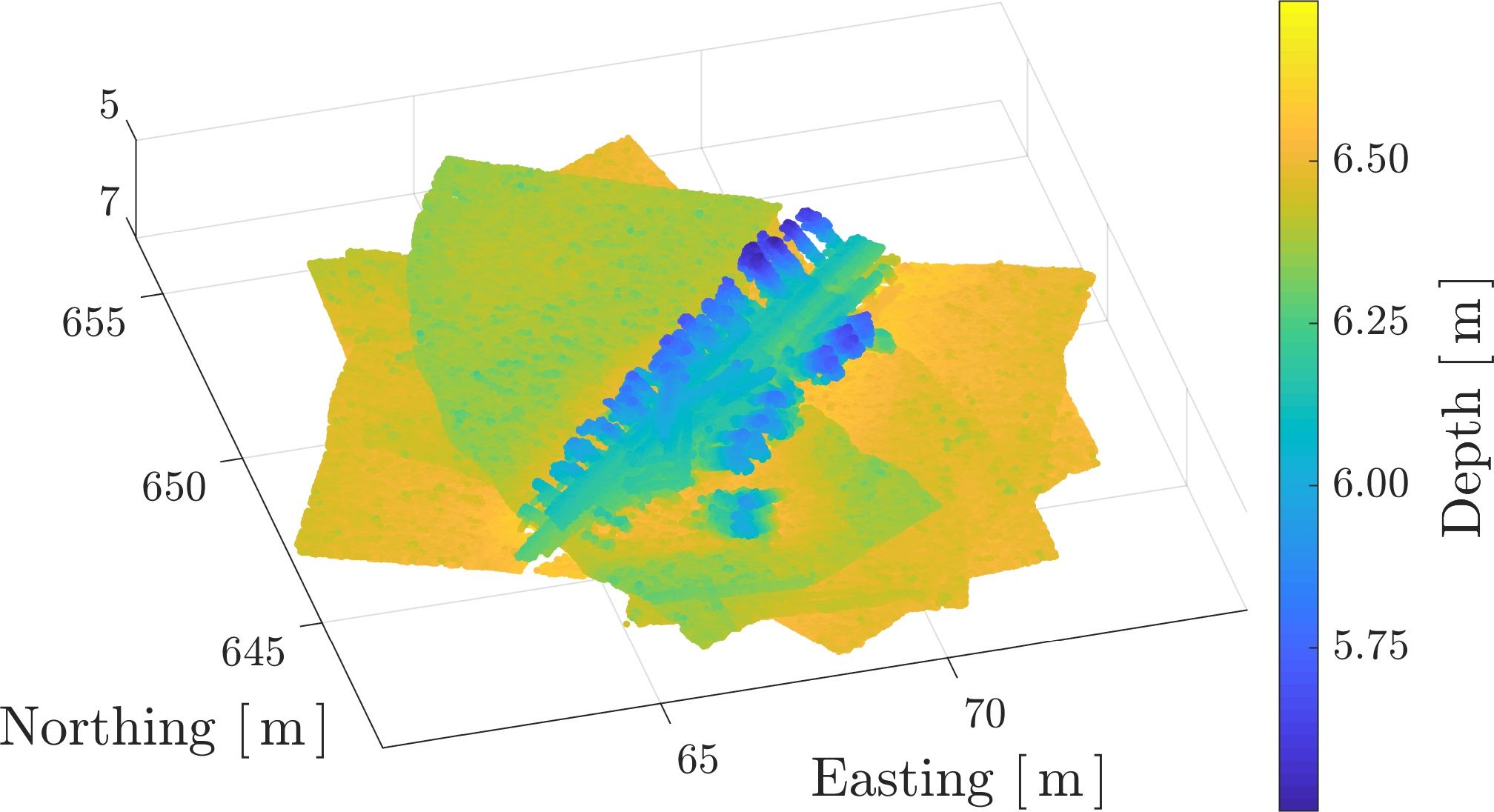}%
        \includegraphics[height=5cm]{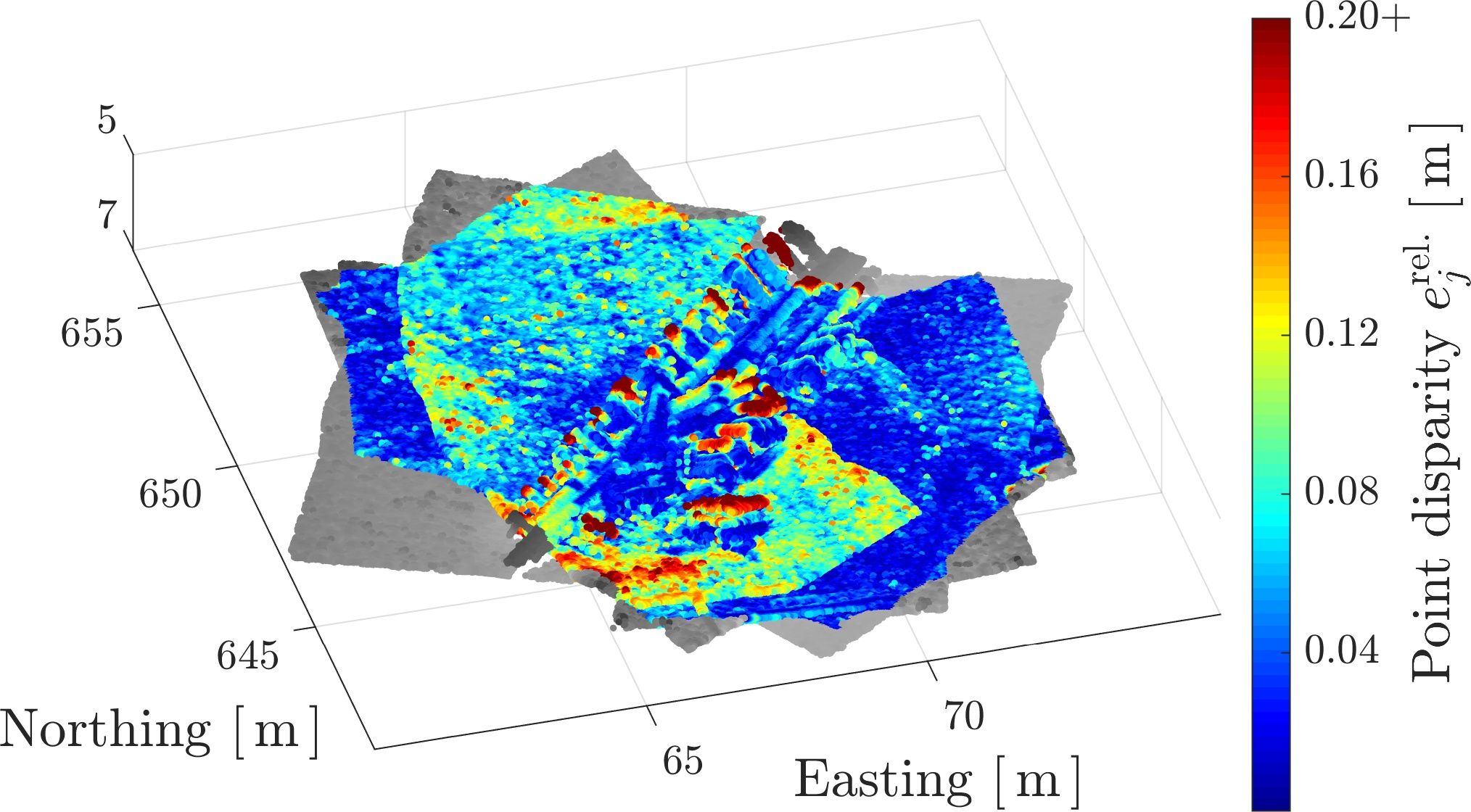}%
	  }%
	}
	\setlength{\subfigheight}{\ht\subfigbox}
	\centering
	\subcaptionbox{Prior elevation (INS) \label{fig:elevation_prior}}{%
		\includegraphics[height=\subfigheight]{figs/updated/small/elevation_prior.jpg}
	}	
    \hspace{3pt}
	\subcaptionbox{Prior disparity (INS) \label{fig:disp_prior}}{%
		\includegraphics[height=\subfigheight]{figs/updated/small/disp_prior.jpg}
	}
	\par\medskip
	\subcaptionbox{Posterior elevation (INS+LC) \label{fig:elevation_posterior}}{%
	    \includegraphics[height=\subfigheight]{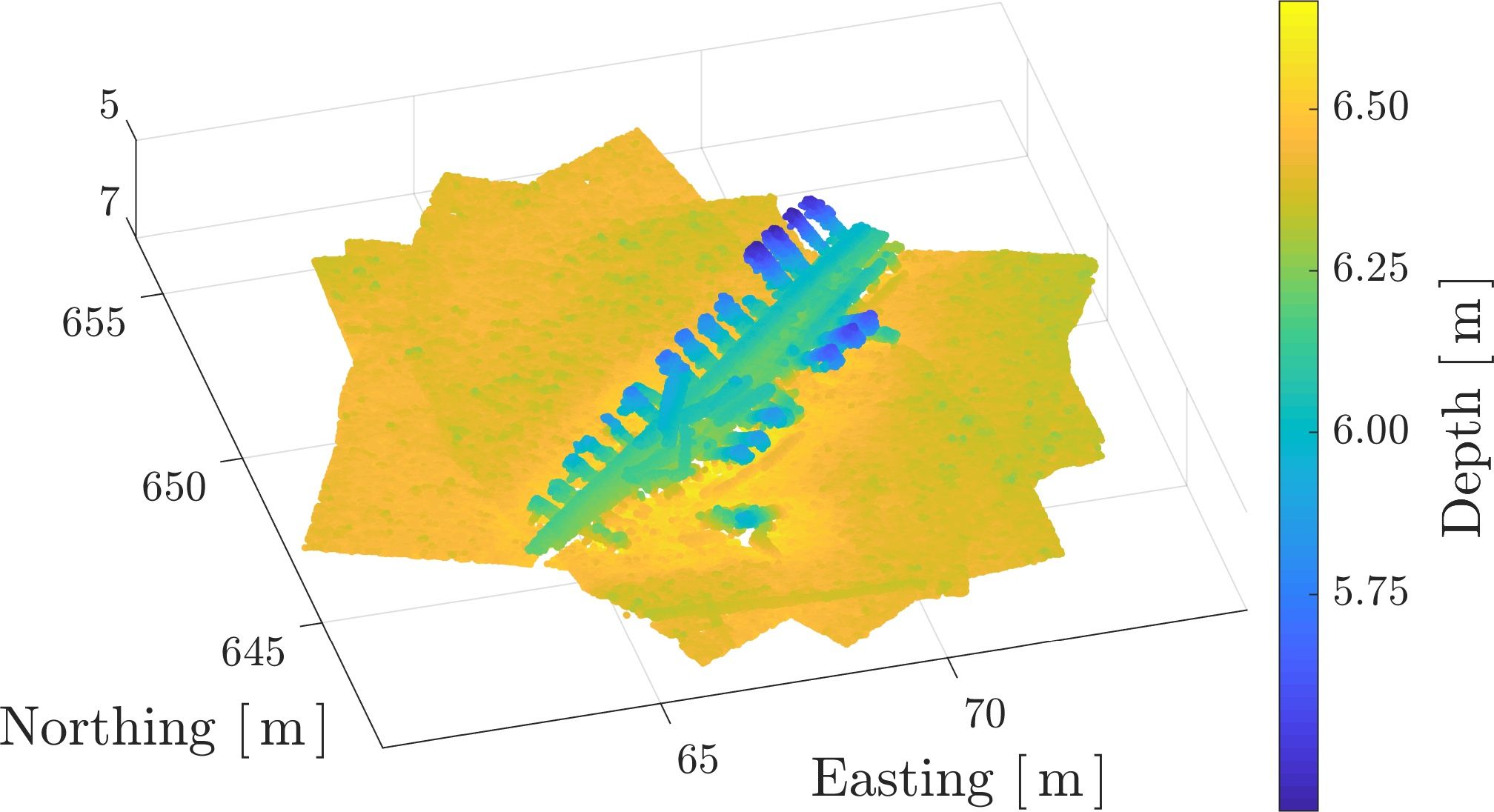}
	}
    \hspace{3pt}
    \subcaptionbox{Posterior disparity (INS+LC) \label{fig:disp_posterior}}{%
        \includegraphics[height=\subfigheight]{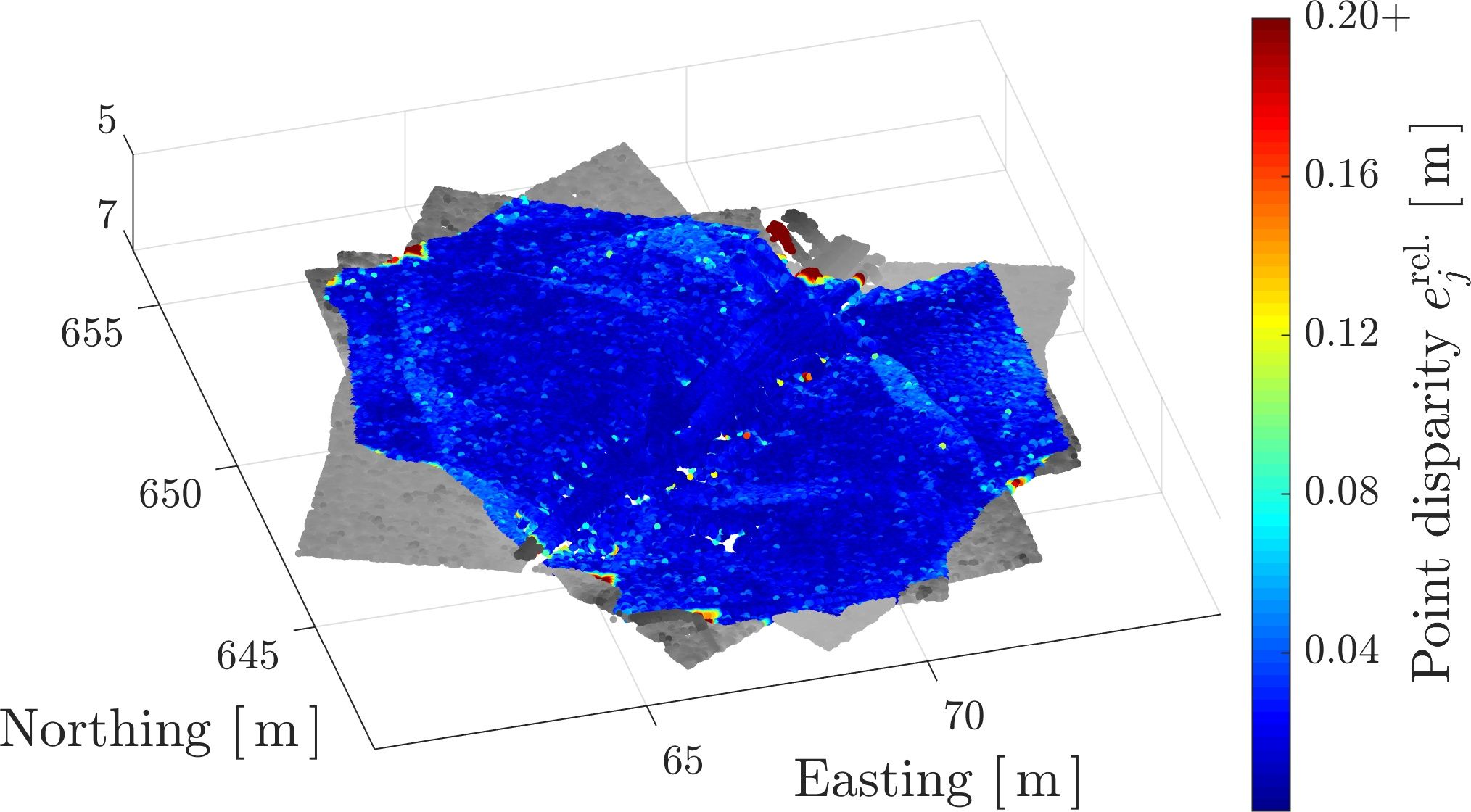}
    }
	\par\medskip
	\subcaptionbox{Ground-truth elevation (INS+GPS)
	\label{fig:elevation_gt}}{%
	    \includegraphics[height=\subfigheight]{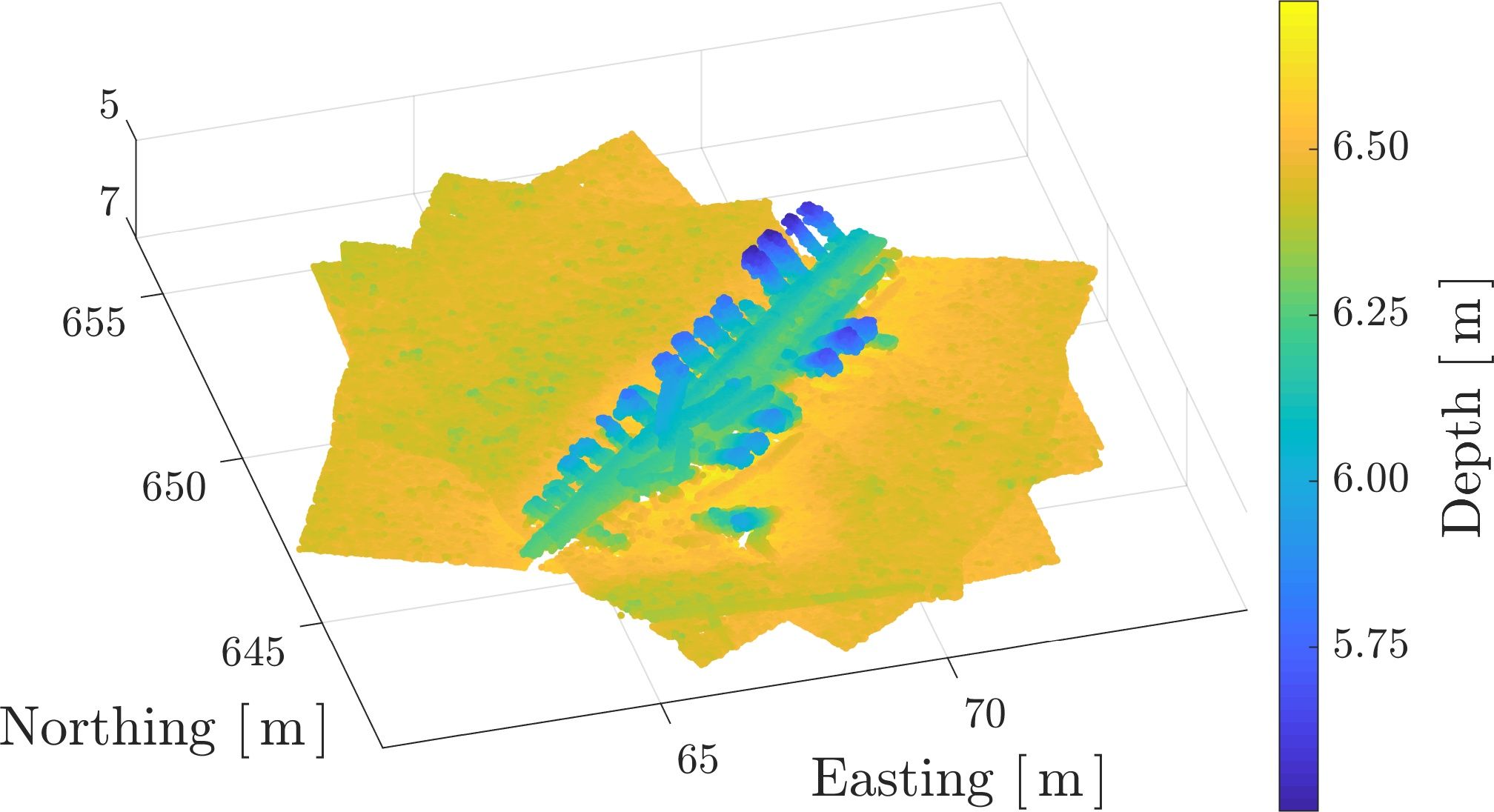}
	}
    \hspace{3pt}
    \subcaptionbox{Ground-truth disparity (INS+GPS)
	\label{fig:disp_gt}}{%
		\includegraphics[height=\subfigheight]{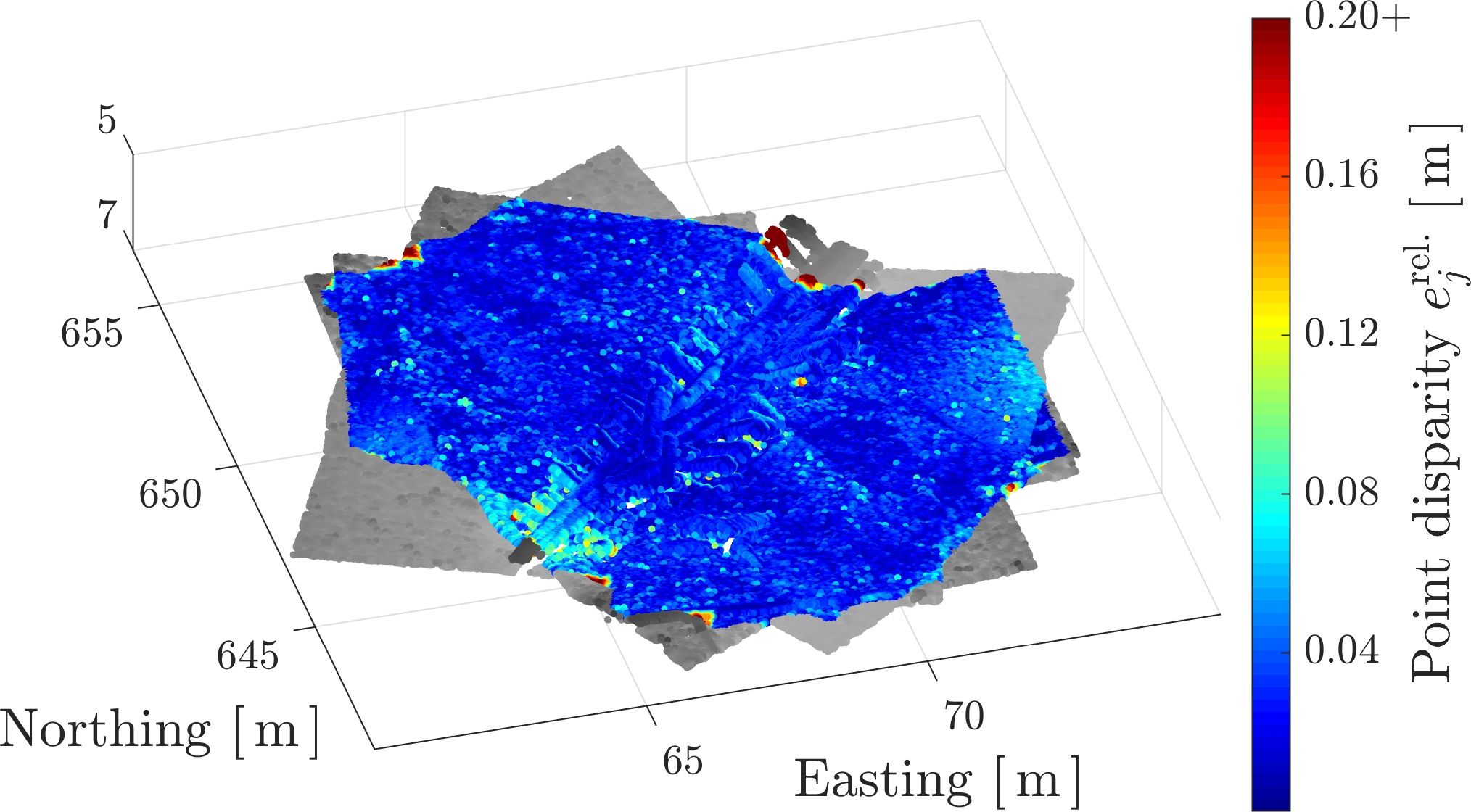}
	} \caption{Visualizing the point disparity error in the shipwreck area for
	the three navigation solutions.  Left column: colour map indicates depth,
	and has been included for context.  Right column: colour map indicates point
	disparity.}
	\label{fig:elevationdisparitymaps}
\end{figure*}

Incorporating loop-closure measurements produces a more self-consistent point
cloud map, as measured by the point disparity error
\eqref{eqn:pointconsistencymetric}.  \Cref{fig:elevationdisparitymaps} shows the
point disparity in the shipwreck area as a heatmap, for each of the three
navigation solutions.  The disparity is computed for each of the eight passes
over the wreck, and is the Euclidean distance from each point in one pass to its
nearest neighbour in the remaining seven passes.  A highly accurate trajectory
estimate is expected to produce a tightly overlapping, crisp point cloud map
from the composite scans, with a low point disparity error. 

From a qualitative evaluation of \Cref{fig:disp_prior}, the dead-reckoned INS
trajectory estimate has clearly produced a self-inconsistent point cloud map.
Areas around the ribs of the shipwreck have point disparity errors of around
\SI{20}{\centi\meter}, while one of the passes shows relatively large errors on
the seabed owing to drift in the depth dimension.  In contrast, both the
posterior and the ground-truth estimates have produced highly self-consistent
maps, with low point disparity errors throughout. 

\begin{figure}[tb]
	\centering
	\begin{subfigure}[t]{\columnwidth}
		\includegraphics[width=\linewidth]{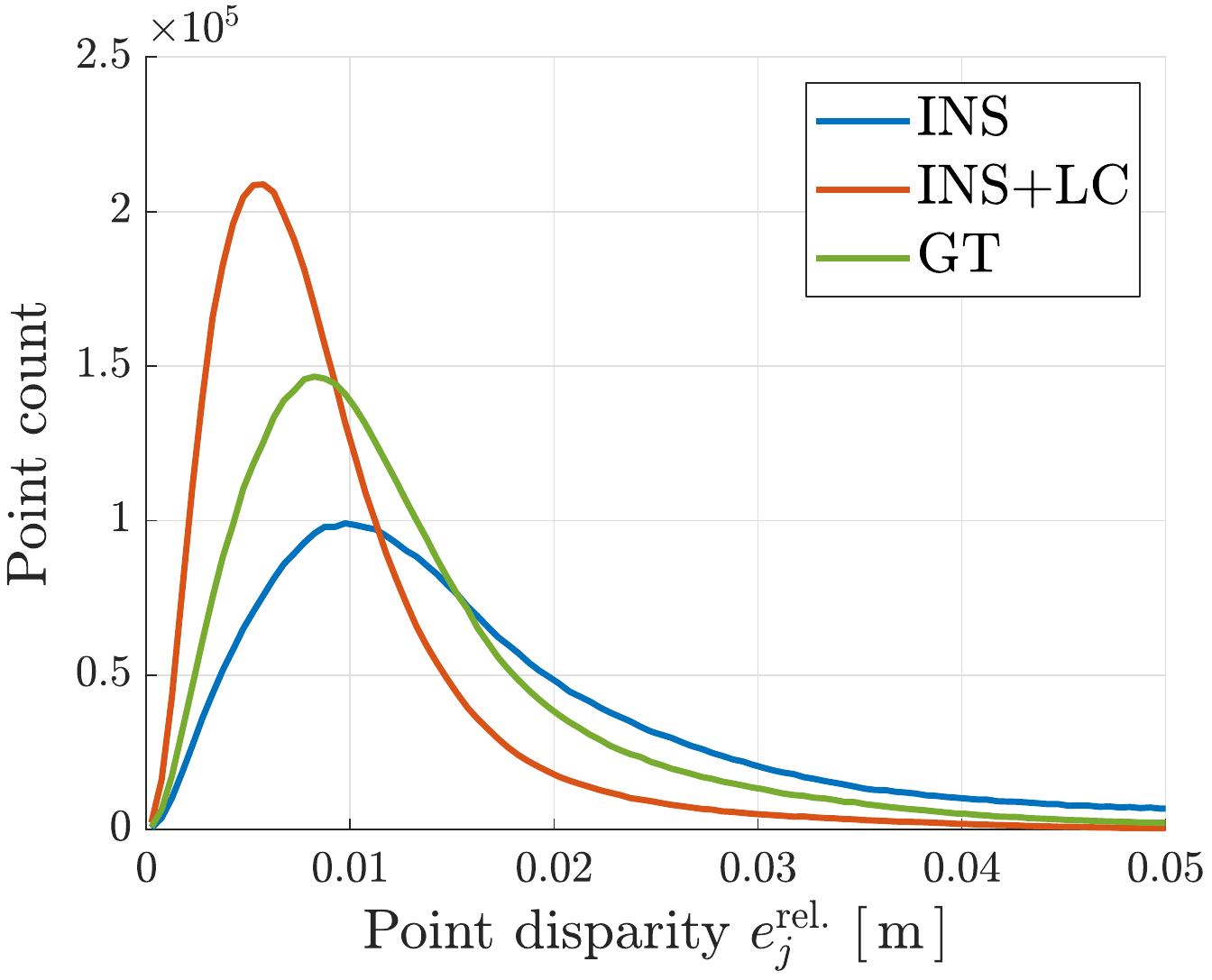}
		\caption{Empirical probability density functions (EPDF)}
        \vspace{8pt}
		\label{fig:pointhist}
	\end{subfigure}
	\begin{subfigure}[t]{\columnwidth}
		\includegraphics[width=\linewidth]{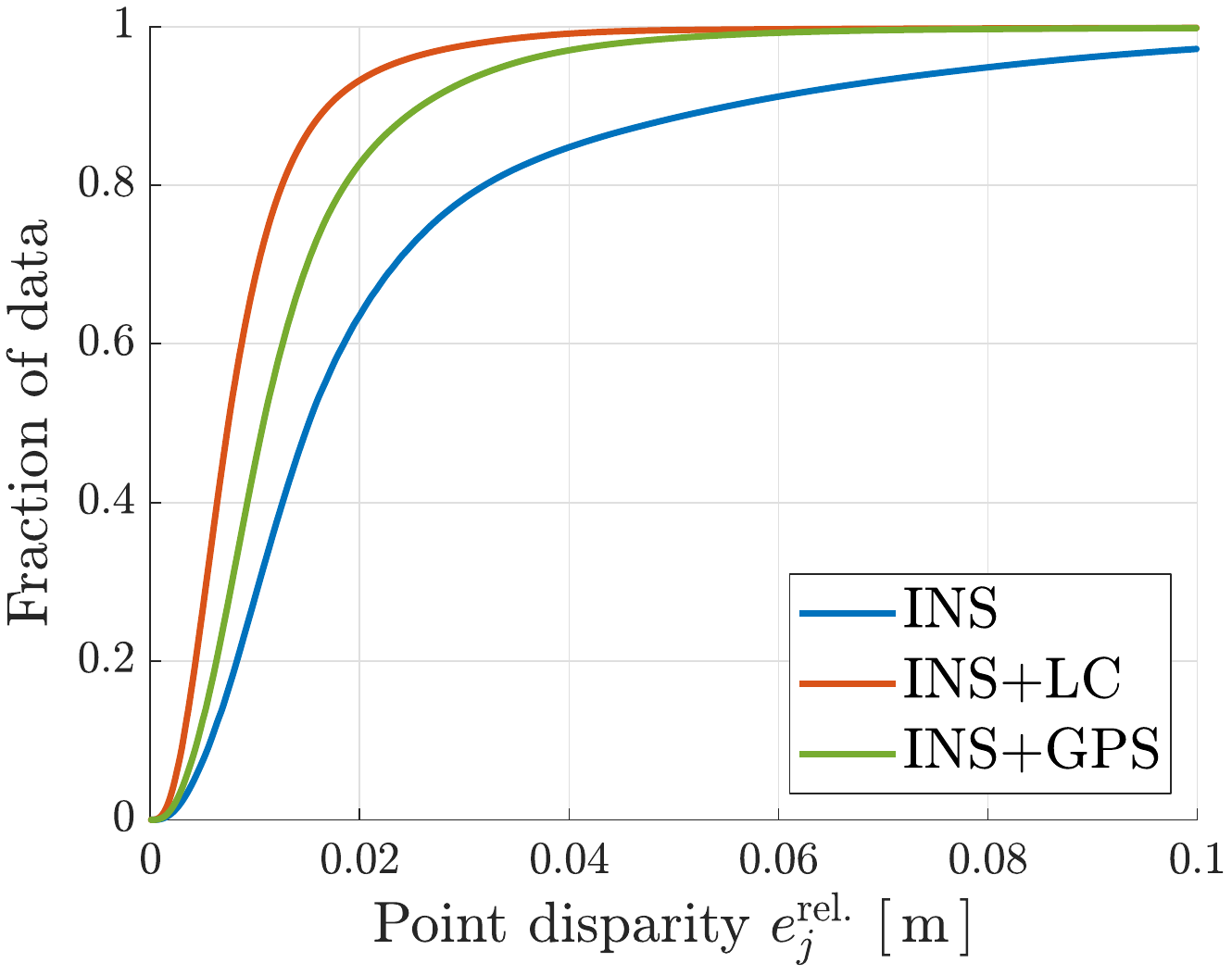}
		\caption{Empirical cumulative distribution functions (ECDF)}
		\label{fig:pointecdf}
	\end{subfigure}
	\caption{Distributions on the point disparity error for each
	of the three navigation solutions for the Wiarton shipwreck field dataset.
	The posterior INS+LC solution produces a more self-consistent point cloud
	map than the ground-truth INS+GPS solution, likely owing to a combination of
	residual navigation and extrinsics errors.  Critical values from the ECDF
	are summarized in \cref{tab:pointerrortable}.}
	\label{fig:fieldpointerrordists}
\end{figure}

Interestingly, the INS+LC solution produces a point cloud map that is
\textit{more self-consistent than the ground-truth estimate}.  This is difficult
to judge qualitatively from \Cref{fig:elevationdisparitymaps}, however
\Cref{fig:fieldpointerrordists} shows the EPDF and ECDF of the disparity error
for each of the three navigation solutions.  Critical values from the ECDF are
tabulated in \cref{tab:pointerrortable}.  In \Cref{fig:pointhist}, the INS+LC
curve peaks to the left of the INS+GPS curve, indicating a lower overall point
disparity error and thus a more self-consistent point cloud map
\cite{Barkby2011a}.  This is likely due to a combination of small estimation
errors in the ground-truth solution and small errors in the scanner extrinsics
estimate $\mbf{T}^{sz}_{b\ell}$ from \eqref{eqn:registerprofiles}.  It should
therefore come as no surprise that the INS+LC solution delivers a more
self-consistent map, as the point disparity error is precisely what is minimized
during point cloud alignment \eqref{eqn:pcalignment}.  For additional images of
the shipwreck area generated using the prior and posterior navigation solutions,
see \Cref{sec:additionalimages}.

\setcounter{table}{7}
\begin{table}[ht]
    \centering
    \caption{Critical values from the ECDF of \Cref{fig:pointecdf}.  For
    example, for the INS solution, \SI{95.45}{\percent} ($2\sigma$) of point
    disparity errors are below \SI{8.51}{\centi \meter}.  Note the improvement
    in the INS+LC solution over the INS+GPS solution.}
    \renewcommand{\arraystretch}{1.2}
    \begin{tabularx}{\columnwidth}{XYYYY}
    \toprule
    Solution & \SI{50}{\percent} & $1\sigma$ & $2\sigma$ & $3\sigma$ \\
	\hline
		INS     & \SI{1.51}{\centi \meter} & \SI{2.23}{\centi \meter} &
		\SI{8.42}{\centi \meter} & \SI{18.11}{\centi \meter} \\
		INS+LC  & \SI{0.75}{\centi \meter} & \SI{1.00}{\centi \meter} &
		\SI{2.35}{\centi \meter} & \SI{7.23}{\centi \meter} \\
        INS+GPS & \SI{1.08}{\centi \meter} & \SI{1.45}{\centi \meter} &
		\SI{3.49}{\centi \meter} & \SI{8.63}{\centi \meter} \\
	\bottomrule
    \end{tabularx}
	\label{tab:pointerrortable}
\end{table}

Again, this improvement in map self-consistency has been achieved
\textit{without access} to the standard ingredients available in typical state
estimation problems.  Visualizing the point cloud map and the resulting
disparity errors is a straightforward way to verify that the loop-closure
measurements have been successfully applied, and that the updates have been
smoothly propagated throughout the trajectory.

Compared to the GPS-aided solution, the improvement in map self-consistency that
comes from leveraging loop-closure measurements may appear modest.  However, an
improvement on the order of centimeters may be consequential for certain subsea
inspection tasks, such as measuring deformation in manmade structures.  In this
respect, the methodology of \Cref{sec:methodology} offers a valuable addition to
inspection and metrology work.  This is especially true for dead-reckoned
solutions, but remains true even when localizing measurements are available, for
example LBL, USBL, or GPS measurements.

\secondupdate{

A Monte Carlo experiment was conducted on the Wiarton field dataset to
test the effectiveness of the loop-closure measurement outlier rejection method
discussed in \Cref{sec:outlierrejection}.  To run the experiment, between one
and five of the seven loop-closure measurements were randomly replaced by
randomly generated measurements.  Thirty Monte Carlo trials were conducted for
each outlier corruption level, for a total of 150 trials.  The number of trials
at each corruption level was chosen to provide a representative statistical
sample.  Additionally, the experimental results obtained using 30 trials per
corruption level were very similar to results obtained when using 20 and 25
trials per level.  

The outlier measurements were generated to mimic the outliers experimentally
observed in the detector/descriptor study in \Cref{sec:pcalignment}.  Outlier
position measurements were uniformly sampled so that ${\| \mbf{r}^{\textrm{xy}}
\| \le \SI{5}{\meter}}$ and ${r^{\textrm{z}} \in \left[ \left[
\SI{-0.5}{\meter}, \SI{0.5}{\meter} \right] \cup \left[ \SI{13.5}{\meter},
\SI{14.5}{\meter} \right] \right]}$, with ${\mbf{r}^{\textrm{out}} =
\begin{bmatrix} (\mbf{r}^{\textrm{xy}})^\trans & r^\textrm{z}
\end{bmatrix}^\trans}$.  This reflects both the planar search bound used to
detect loop-closure candidates \eqref{eqn:findlc} as well as the range
``flipping'' effect discussed in \Cref{sec:pcalignment}.  Outlier attitude
measurements were uniformly sampled according to ${\phi^{\textrm{out}}_j \in
(-\pi, \pi] \ \SI{}{\radian}, j = 1,2,3}$.  An outlier measurement
$\mbs{\Xi}^\textrm{out}_{\ell_1 \ell_2}$ is then generated according
\begin{equation}
    \mbs{\Xi}^\textrm{out}_{\ell_1 \ell_2} = \begin{bmatrix}
        \mbf{C}^\textrm{out}(\mbs{\phi}^\textrm{out}) & \mbf{r}^{\textrm{out}} \\ \mbf{0} & 1
    \end{bmatrix}.
    \label{eqn:mc_outlier}
\end{equation}

Results from this experiment are summarized throughout \Cref{fig:mc_results}.
All 150 Monte Carlo trials are plotted in \Cref{fig:mc_trajectory}, along with
the ground-truth ``INS+GPS'' trajectory and the prior ``INS'' trajectory
estimate.  No visible navigation failures are seen in \Cref{fig:mc_trajectory},
implying the adaptive robust cost function is effective in rejecting false
loop-closure measurements.  A zoom of the shipwreck region in
\Cref{fig:mc_trajectories_zoom} shows the Monte Carlo trajectory samples
gracefully decaying from the ground-truth solution to the prior estimate as more
outliers are included.  This behaviour is also seen in the relative displacement
error plot of \Cref{fig:mc_drift}, where the mean relative $(x,y)$ navigation
drift \eqref{eqn:displacementmetric} is plotted for each outlier corruption
level.  The trend of higher outlier rates producing larger relative drift values
is reminiscent of the ablation study summarized in \Cref{fig:drift_plot}, in
which loop-closure measurements are simply removed from the solution.  This
provides sound evidence that the proposed outlier rejection algorithm is
successful in identifying and removing false loop-closure measurements.

Finally, the grey patch in \Cref{fig:mc_drift} shows the worst-case relative
displacement error at each time step across all 150 Monte Carlo trials.  Compared
to the prior ``INS'' estimate (blue curve), it is clear that, for this dataset,
the proposed methodology delivers worst-case posterior estimates that are, at
any given time, \textit{no worse than the prior estimate, even in instances with
extreme outlier rates.}

\begin{figure*}[htb]
	\sbox\subfigbox{%
	  \resizebox{\dimexpr0.94\textwidth-1em}{!}{%
		\includegraphics[height=3cm]{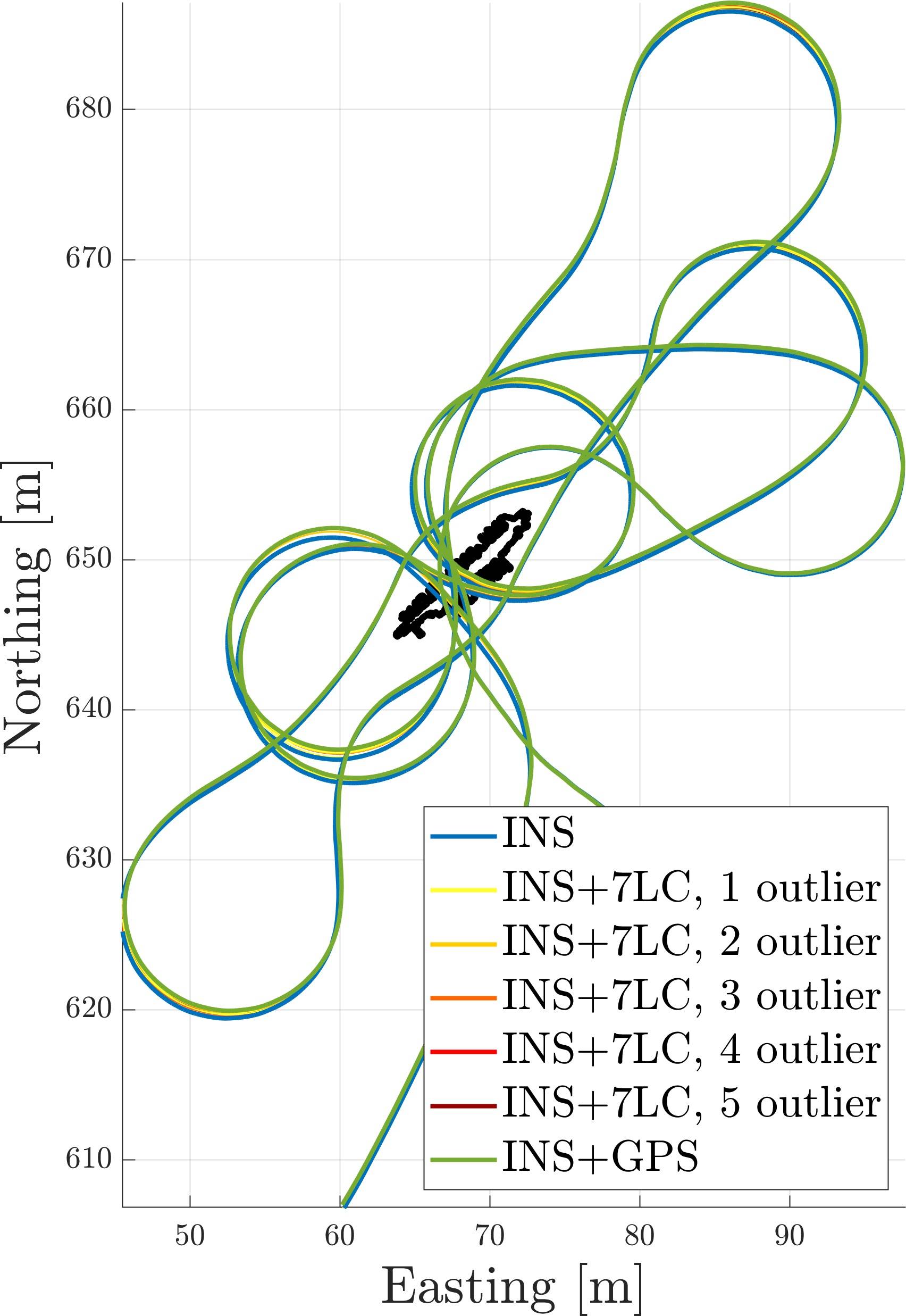}%
		\includegraphics[height=3cm]{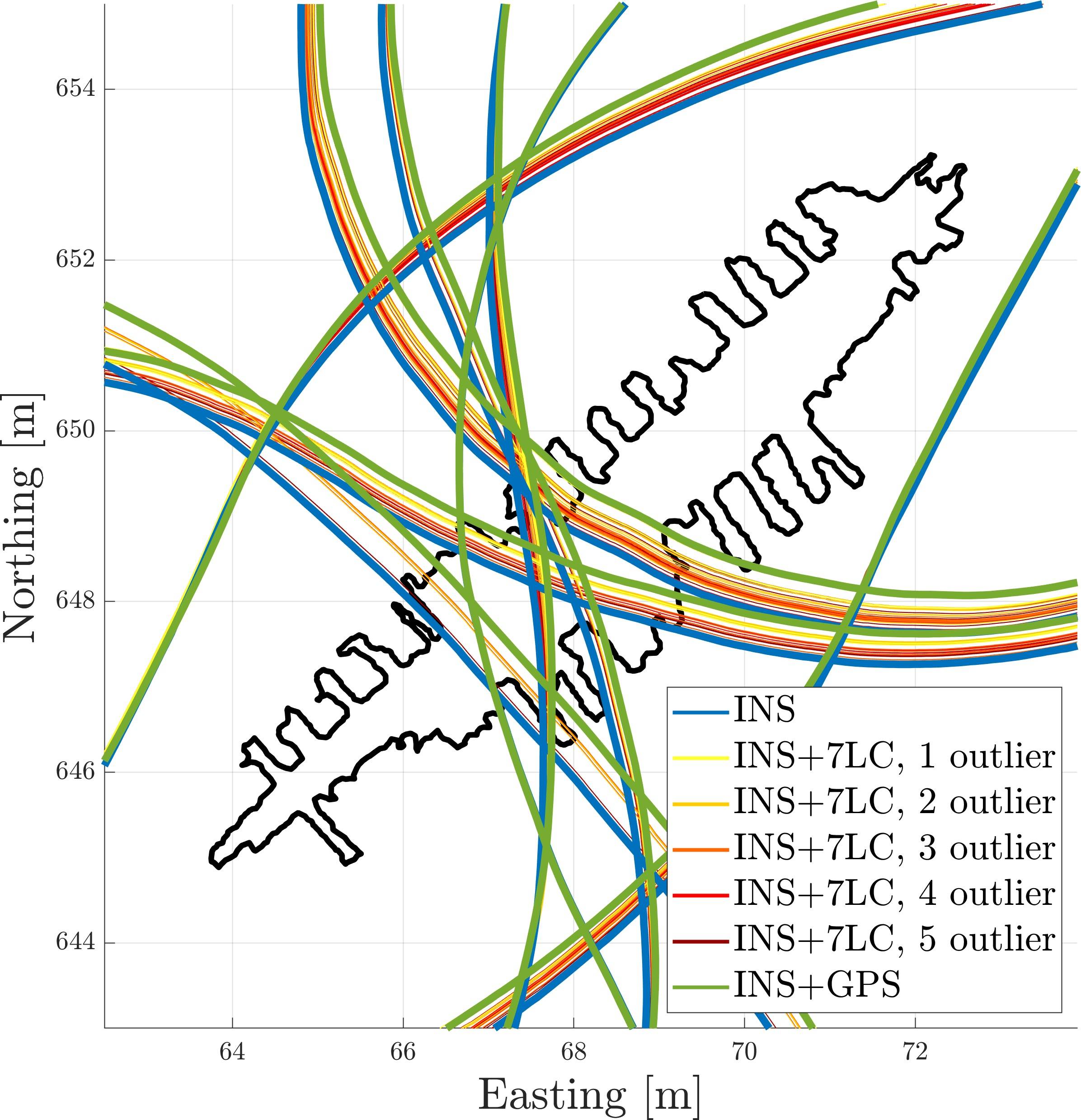}%
		\includegraphics[height=3cm]{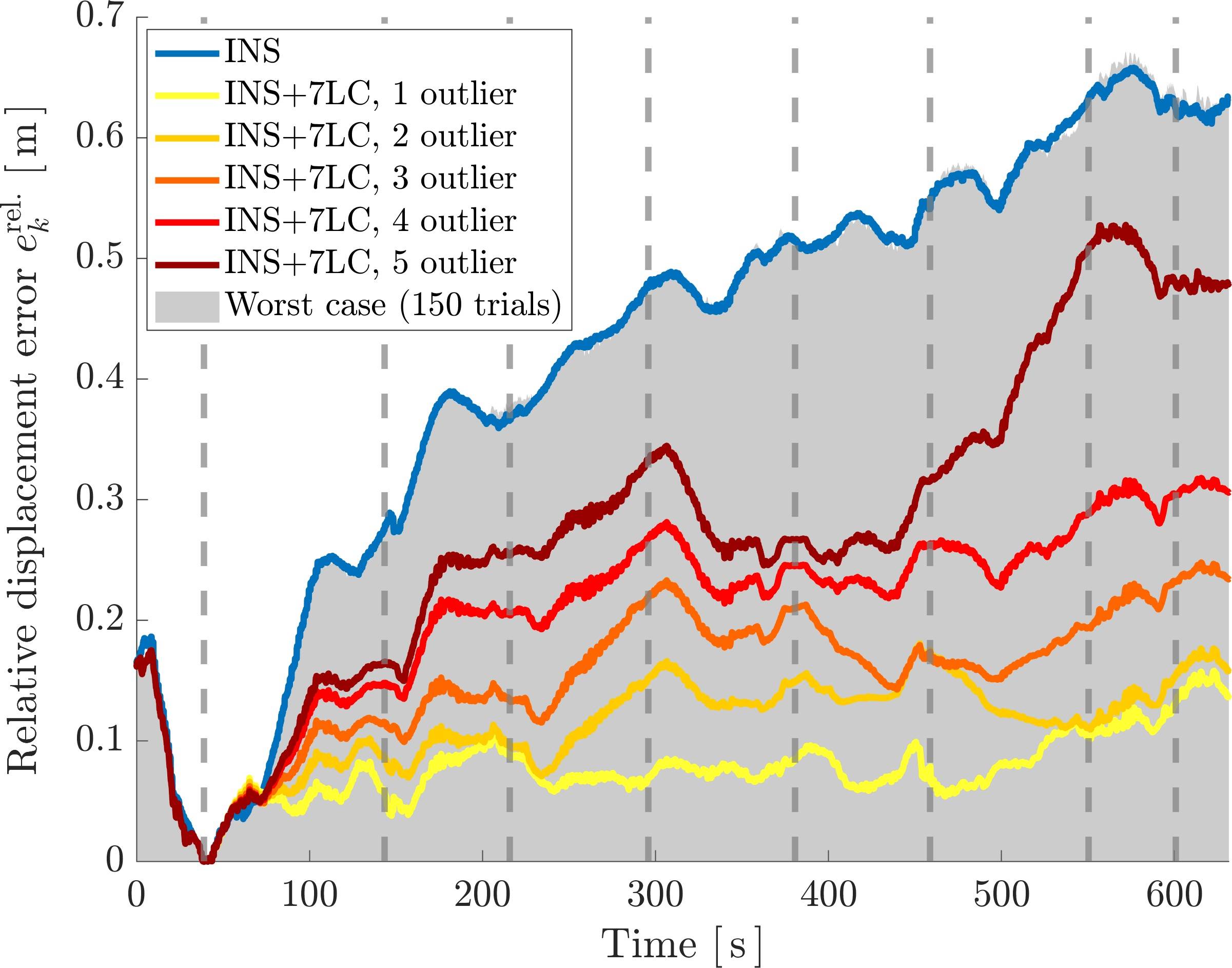}%
	  }%
	}
	\setlength{\subfigheight}{\ht\subfigbox}
	\centering
	\subcaptionbox{150 Monte Carlo trajectories  \label{fig:mc_trajectory}}{%
	  \includegraphics[height=\subfigheight]{figs/updated/outlier_rejection_study/outlier_traj_small.jpg}
	}
	\quad
	\subcaptionbox{Shipwreck section \label{fig:mc_trajectories_zoom}}{%
	  \includegraphics[height=\subfigheight]{figs/updated/outlier_rejection_study/outlier_traj_zoom_small.jpg}
	}
	\quad
	\subcaptionbox{Mean navigation drift by outlier corruption level \label{fig:mc_drift}}{%
	  \includegraphics[height=\subfigheight]{figs/updated/outlier_rejection_study/outlier_drift_small.jpg}
	} \caption{\secondupdate{Trajectory estimates and associated relative drift
	errors \eqref{eqn:displacementmetric} for 150 Monte Carlo trials.  For each
	trial, between 1 and 5 of the seven loop-closure measurements are replaced
	by outlier measurements \eqref{eqn:mc_outlier}, with 30 trials conducted per
	outlier corruption level.  The outlier rejection method discussed in
	\Cref{sec:outlierrejection} is effective in rejecting false loop-closure
	measurements, with no failures visible in \Cref{fig:mc_trajectory}.  The
	zoom in \Cref{fig:mc_trajectories_zoom} show a graceful decay from the
	ground-truth ``INS+GPS trajectory'' to the prior ``INS'' estimate as more
	outlier measurements are added.  This is confirmed by the relative
	displacement error plot in \Cref{fig:mc_drift}, which simply follows the
	trend of measurement removal first seen in the ablation study of
	\Cref{fig:drift_plot}.  The worst-case position drift at each time step
	measured across all 150 trials is shown as a grey patch in
	\Cref{fig:mc_drift}.  When compared against the relative displacement error
	from the prior ``INS'' estimate (blue line), the proposed methodology is
	seen to produce estimates that are, at any given time, \textit{no worse than
	the prior estimate, even in instances with extreme outlier rates.}}}
    \label{fig:mc_results}
\end{figure*}
}

\section{Conclusion}
\label{sec:conclusion}

\secondupdate{This works presents a novel and comprehensive method for
systematically conditioning the output of a COTS DVL-INS navigation system on
loop-closure measurements} for the purpose of improving the
\textit{self-consistency} \cite{Roman2007,Kuemmerle2009} of the resulting
bathymetric map.  The method relies on a combination of relative pose and
white-noise-on-acceleration \cite{Anderson2015} error terms to smoothly
integrate the measurements in a batch state estimation framework.  

\secondupdate{The first contribution of this work is the development of a robust
front-end algorithm for computing high-precision loop-closure measurements from
3D scans of challenging underwater environments.  Second, loop-closure
measurements are cleanly incorporated into an existing state estimate via a
factor graph optimization framework, \textit{without access} to raw sensor
measurements, sensor models, or other information typically required in
conventional state estimation problems.}

\secondupdate{The effectiveness of the proposed} method was demonstrated for
both simulated and field datasets using loop-closure measurements from an
underwater laser scanner.  The same simple hyperparameter structure was used for
both studies, with good results.  For the field results, conditioning the
dead-reckoned DVL-INS estimate on loop-closure measurements produced a markedly
more self-consistent point cloud map of an underwater shipwreck.  Incorporating
all seven loop-closure measurements resulted in a maximum relative position
drift of \SI{8.4}{\centi \meter} over a \SI{576}{\meter} trajectory, with a
final position error of \SI{6.83e-3}{\percent} of distance traveled.  This
represents an order of magnitude improvement over unaided commercial DVL-INS
systems.  \secondupdate{Additionally, the proposed methodolgy was demonstrated
to be robust to false loop-closure measurements.}

Future work will \secondupdate{primarily} focus on hyperparameter training
through an expectation-maximization framework, for example \cite{Wong2020} and
\cite{Barfoot2020}.  The algorithm will be tested over longer trajectories with
more varied terrain, including open seabed \cite{Hitchcox2020}.  Finally, future
work may also incorporate image information, in the form of conventional image
descriptors and textured point cloud maps.

\section*{Acknowledgment}

The authors would like to thank Ryan Wicks of Voyis for providing experimental
data and guidance, and Martin J{\o}rgensen of Sonardyne International for
providing simulation data and helpful feedback.

\printbibliography
\newpage

\begin{IEEEbiography}[{\includegraphics[width=1in,height=1.25in,clip,keepaspectratio]{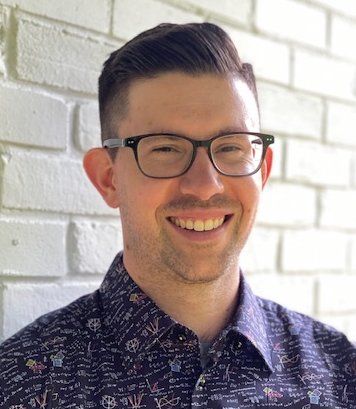}}]
    {Thomas Hitchcox} received his B.Eng. and M.Eng. degrees in mechanical
    engineering in 2015 and 2018, respectively, from McGill University,
    Montreal, QC, Canada.  He is currently a Ph.D. Candidate with the Department
    of Mechanical Engineering at McGill.  His research interests include state
    estimation, computer vision, and robust algorithms for point cloud filtering
    and alignment. 
\end{IEEEbiography}
\begin{IEEEbiography}[{\includegraphics[width=1in,height=1.25in,clip,keepaspectratio]{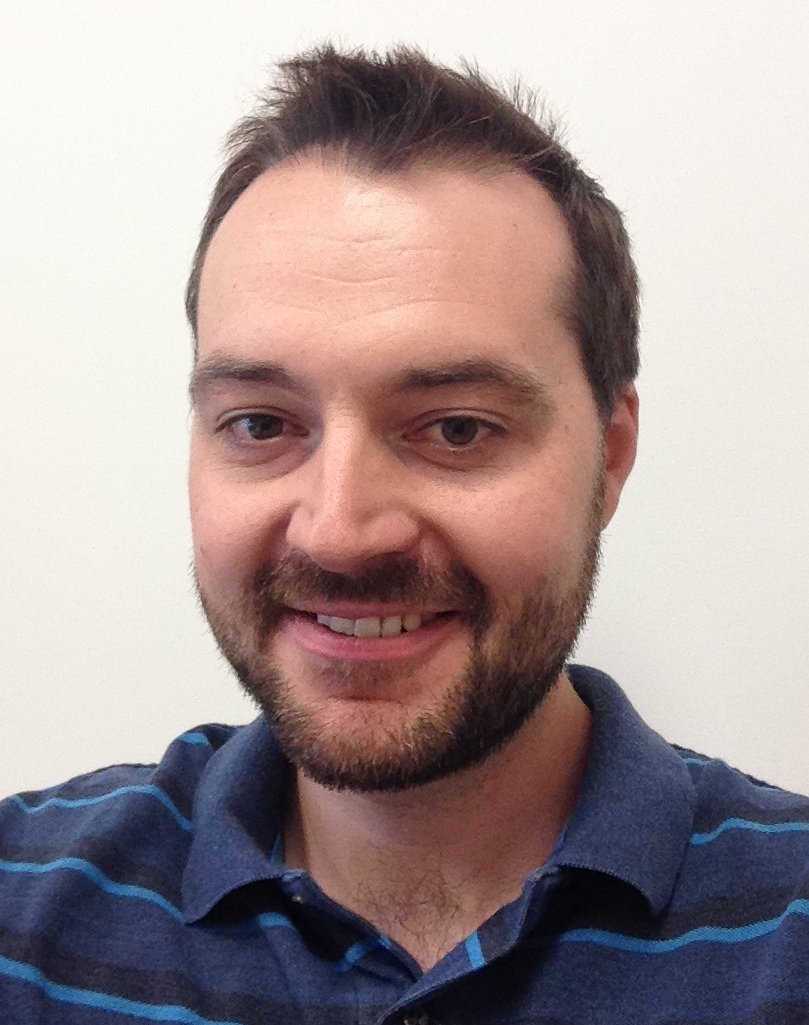}}]
    {James Richard Forbes} James Richard Forbes received the B.A.Sc. degree in
    Mechanical Engineering (Honours, Co-op) from the University of Waterloo,
    Waterloo, ON, Canada in 2006, and the M.A.Sc. and Ph.D. degrees in Aerospace
    Science and Engineering from the University of Toronto Institute for
    Aerospace Studies (UTIAS), Toronto, ON, Canada in 2008 and 2011,
    respectively. James is currently an Associate Professor and William Dawson
    Scholar in the Department of Mechanical Engineering at McGill University,
    Montreal, QC, Canada. James is a Member of the Centre for Intelligent
    Machines (CIM), and a Member of the Group for Research in Decision Analysis
    (GERAD). James was awarded the McGill Association of Mechanical Engineers
    (MAME) Professor of the Year Award in 2016, the Engineering Class of 1944
    Outstanding Teaching Award in 2018, and the Carrie M. Derick Award for
    Graduate Supervision and Teaching in 2020. The focus of James’ research is
    navigation, guidance, and control of robotic systems.
\end{IEEEbiography}
\vfill

\onecolumn  
\appendices
\crefalias{section}{appendix}

\section{Supporting Derivations}
\label{apx:derivations}

\begin{refsection}
\newrefcontext[labelprefix=\thesection]
\renewcommand{\theequation}{\thesection.\arabic{equation}}
\setcounter{equation}{0}
\renewcommand{\thefigure}{\thesection.\arabic{figure}}
\setcounter{figure}{0}
\renewcommand{\thetable}{\thesection.\arabic{table}}
\setcounter{table}{0}

\subsection{Introduction}
\label{apx:intro}

The purpose of this appendix is to derive in detail the prior, process, and loop
closure Jacobians appearing in \Cref{sec:minimizeJ} of ``Improving
Self-Consistency in Underwater Mapping through Laser-Based Loop Closure.''  Key
identities from matrix Lie group theory are reviewed in
\Cref{apx:preliminaries}, and the white-noise-on-acceleration (WNOA) motion
prior \cite{Anderson2015} is reviewed in \Cref{apx:wnoaprior}.
\Cref{apx:errorkin} examines the WNOA error kinematics.  Finally, the necessary
Jacobians are derived in \Cref{apx:jacobians}.  The intent of this appendix is
to make these derivations accessible, with key steps and identities indicated
throughout.  For a more detailed treatment of matrix Lie group theory, please
consult the references cited throughout, particularly \cite{Barfoot2017} and
\cite{Sola2018}.

\subsection{Preliminaries}
\label{apx:preliminaries}

\subsubsection{Matrix Lie groups}
\label{apx:liegroups}

A matrix Lie group $G$ is a set of real, invertible ${n\times n}$ matrices that
is closed under matrix multiplication.  Associated with every matrix Lie group
is a matrix Lie algebra $\mathfrak{g}$, defined as the tangent space at the
group identity, ${\mathfrak{g} \triangleq T_\eye G}$.  The matrix Lie algebra is
a vector space closed under the operation of the Lie bracket
\cite[Sec.~10.2.6]{Chirikjian2011}.  It is often more convenient to work with
isometric representations of matrix Lie algebra elements, namely ${\mbs{\xi} \in
\rnums^d}$, where ${\mbs{\xi}^\wedge \in \mathfrak{g}}$.
 
\begin{figure}[htb]
	\centering
	\includegraphics[width=0.7\textwidth]{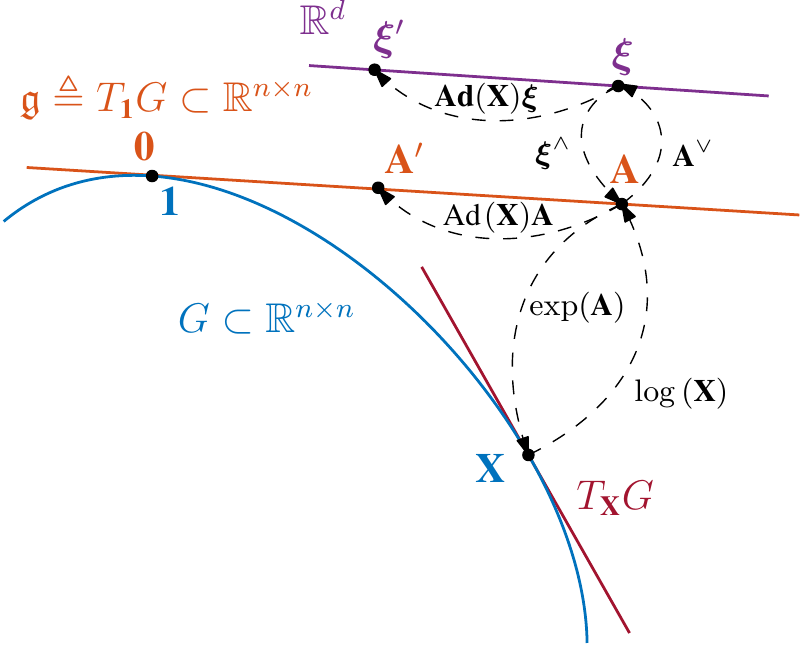}
	\caption{Matrix Lie groups, modified from \cite{Bourmaud2015a}.  Mappings
	are shown between the matrix Lie group element $\mbf{X}$, its matrix Lie
	algebra representation ${\mbf{A} = \mbs{\xi}^\wedge}$, and its $\rnums^d$
	representation $\mbs{\xi}$.  Note the difference between the adjoint
	\textit{operator} $\Adjop(\cdot)$ and the adjoint \textit{matrix},
	$\Adj(\cdot)$, discussed in \Cref{apx:adjoint}.  All perturbations are
	modeled in the matrix Lie algebra $\mathfrak{g}$ (orange), defined as the
	tangent space at the group identity. }
	\label{fig:liegroups}
\end{figure}

A Lie group and its corresponding Lie algebra are related through the
exponential map.  For matrix Lie groups this is simply the matrix exponential
\cite[Sec.~7.1.3]{Barfoot2017}.  For ${\mbf{X} \in G}$, this leads to
expressions of the form 
\begin{equation}
	\mbf{X} = \exp(\mbs{\xi}^\wedge), 
\end{equation}
where $\mbs{\xi}^\wedge$ is the representation of $\mbf{X}$ in $\mathfrak{g}$,
and $\mbs{\xi}$ is the representation of $\mbf{X}$ in $\rnums^d$.  Finally, the
matrix logarithm is used to move from the matrix Lie group to the matrix Lie
algebra, as in 
\begin{equation}
	\mbs{\xi}^\wedge = \log \left( \mbf{X} \right).
\end{equation}
The relationship between a matrix Lie group, its associated matrix Lie algebra,
and the isometric space $\rnums^d$, as well as several other quantities
discussed throughout this document, is illustrated in \Cref{fig:liegroups},
which is inspired by, but modified from, \cite{Bourmaud2015a}.

\subsubsection{Errors and perturbations on matrix Lie groups}
\label{apx:perturbations}

There are four ways to define a matrix Lie group error \cite{Arsenault2019}.
These are summarized in \Cref{tab:errordefinitions}, along with their
corresponding perturbation schemes.  The error definition and perturbation
scheme are linked.  For example the selection of a left-invariant error
definition necessitates the use of a left-invariant perturbation scheme.  To
see this, consider
\begin{subequations}
	\begin{align*}
		\delta \mbf{X} =& \ \mbf{X}\inv \mbfbar{X}, \numberthis \label{eqn:lefterrorapx} \\
		\exp(\delta \mbs{\xi}^\wedge) =& \ \mbf{X}\inv \mbfbar{X}, \\
		\mbf{X} \exp(\delta \mbs{\xi}^\wedge) =& \ \mbf{X} \mbf{X}\inv \mbfbar{X}, \\
		\mbf{X} =& \ \mbfbar{X} \exp( -\delta \mbs{\xi}^\wedge), \numberthis \label{eqn:leftpert}
	\end{align*}
\end{subequations}
where \eqref{eqn:lefterrorapx} is the left-invariant matrix Lie group error and
\eqref{eqn:leftpert} is the left-invariant perturbation scheme.  This work uses
a left-invariant error definition. 

\newcolumntype{Y}{>{\centering\arraybackslash}X}
\begin{table}[b]
    \centering
    \caption{Matrix Lie group error definitions and corresponding perturbation schemes.}
    \renewcommand{\arraystretch}{1.2}
    \begin{tabularx}{0.7\textwidth}{YYYY}
    \toprule
    Error definition & Matrix Lie group error & Perturbation scheme \\
    \hline
		Right invariant & $\delta \mbf{X} = \mbfbar{X} \mbf{X}\inv$ & $\mbf{X} =
		\exp(-\delta \mbs{\xi}^\wedge) \mbfbar{X}$ \\
		Right perturbation & $\delta \mbf{X} = \mbf{X} \mbfbar{X}\inv$ & $\mbf{X} =
		\exp(\delta \mbs{\xi}^\wedge) \mbfbar{X} \hphantom{-}$ \\
		Left invariant & $\delta \mbf{X} = \mbf{X}\inv \mbfbar{X}$ & $\mbf{X} =
		\mbfbar{X} \exp(-\delta \mbs{\xi}^\wedge)$ \\
		Left perturbation & $\delta \mbf{X} = \mbfbar{X}\inv \mbf{X}$ & $\mbf{X} =
		\mbfbar{X} \exp(\delta \mbs{\xi}^\wedge) \hphantom{-}$ \\
	\bottomrule
    \end{tabularx}
	\label{tab:errordefinitions}
\end{table}

\subsubsection{The Baker-Campbell-Hausdorff (BCH) equation}
\label{apx:bch}

The BCH equation describes how to combine elements of the matrix Lie algebra
\cite[Sec.~10.2.7]{Chirikjian2011},
\vspace{-3pt}
\begin{equation}
	\mbf{c}^\wedge = \log \left( \exp(\mbf{a}^\wedge) \exp(\mbf{b}^\wedge) \right),
	\label{eqn:bch}
    \vspace{-3pt}
\end{equation}
where ${\mbf{a}^\wedge, \mbf{b}^\wedge, \mbf{c}^\wedge \in \mathfrak{g}}$.
Elements of $\mathfrak{g}$ are therefore correctly combined on the group $G$,
through application of the exponential map.  However, the following
approximation, 
\vspace{-3pt}
\begin{equation}
	\mbf{c} \approx \mbf{a} + \mbf{J}^{\textrm{r}}(\mbf{a})\inv \mbf{b}, 
	\label{eqn:bchsmallb}
    \vspace{-3pt}
\end{equation}
is valid if $\mbf{b}$ is small \cite[Sec.~7.1.5]{Barfoot2017}, where
$\mbf{J}^{\textrm{r}}(\mbs{\xi})$ is the right Jacobian of $G$.  The following
approximation,
\vspace{-3pt}
\begin{equation}
	\mbf{c} \approx \mbf{a} + \mbf{b},
    \vspace{-3pt}
\end{equation}
is valid if both $\mbf{a}$ and $\mbf{b}$ are small.  This leads to the following
three useful identities related to the BCH equation, 
\begin{subequations}
	\begin{align}
		\exp(\mbs{\xi}^\wedge) \exp(\delta \mbs{\xi}^\wedge) \approx& \ \exp \big( ( \mbs{\xi} + \mbf{J}^{\textrm{r}}(\mbs{\xi})\inv \delta \mbs{\xi} )^\wedge \big),
		\label{eqn:bchapprox1} \\
		\exp((\mbs{\xi} + \delta \mbs{\xi})^\wedge) \approx& \ \exp (\mbs{\xi}^\wedge) \exp \big( ( \mbf{J}^{\textrm{r}}(\mbs{\xi}) \delta \mbs{\xi} )^\wedge \big),
		\label{eqn:bchapprox2} \\
		\exp(\delta \mbs{\xi}^\wedge_1) \exp(\delta \mbs{\xi}^\wedge_2) \approx& \ \exp \left( \left( \delta \mbs{\xi}_1 + \delta \mbs{\xi}_2 \right)^\wedge \right).
		\label{eqn:bchapprox3}
	\end{align}
    \label{eqn:bchapprox}
\end{subequations}
%

\subsubsection{The adjoint operator and the adjoint matrix}
\label{apx:adjoint}

The adjoint operator maps the effects of perturbations about the group identity
to other group elements.  For ${\mbf{X} \in G}$ and ${\mbs{\xi}^\wedge \in
\mathfrak{g}}$, the adjoint \textit{operator} $\Adjop: \mathfrak{g} \to
\mathfrak{g}$ is defined as \cite[Sec.~2.5]{Eade2014}
\begin{equation}
	\Adjop (\mbf{X}) \mbs{\xi}^\wedge \triangleq \mbf{X} \mbs{\xi^\wedge} \mbf{X}\inv.
\end{equation}
The adjoint \textit{matrix} $\Adj: \rnums^d \to \rnums^d$ encodes the effects of
the adjoint operator directly on $\rnums^d$ \cite{Sola2018},
\vspace{-3pt}
\begin{equation}
	\Adj(\mbf{X}) \mbs{\xi} \triangleq \left( \mbf{X} \mbs{\xi}^\wedge \mbf{X}\inv \right)^\vee.  
    \label{eqn:adjmatrix}
    \vspace{-3pt}
\end{equation}
The adjoint matrix may also be defined in terms of the left and right group
Jacobians \cite[Sec.~7.1.5]{Barfoot2017},
\begin{equation}
	\Adj(\mbf{X}) \triangleq \mbf{J}^\ell(\mbs{\xi}) \mbf{J}^\textrm{r}(\mbs{\xi})\inv,
	\label{eqn:adjjacidentity}
\end{equation}
where ${\mbf{J}^\ell (\mbs{\xi}) = \mbf{J}^{\textrm{r}}(-\mbs{\xi}) }$.
Finally, the adjoint matrix exists in the matrix Lie algebra as 
\begin{equation}
	\left( \adj(\mbs{\xi}_1^\wedge) \mbs{\xi}_2 \right)^\wedge \triangleq \left[ \mbs{\xi}^\wedge_1, \mbs{\xi}^\wedge_2 \right] = \mbs{\xi}^\wedge_1 \mbs{\xi}^\wedge_2 - \mbs{\xi}^\wedge_2 \mbs{\xi}^\wedge_1,
\end{equation}
where ${\mbs{\xi}^\wedge_1, \mbs{\xi}^\wedge_2 \in \mathfrak{g}}$ and ${[\cdot,
\cdot]}$ is the Lie bracket \cite[Sec.~10.2.6]{Chirikjian2011}.  $\Adj$ and
$\adj$ are related through the exponential map, 
\begin{equation}
	\Adj(\mbf{X}) = \exp \left( \adj( \mbs{\xi}^\wedge ) \right), 
    \vspace{-3pt}
\end{equation}
where ${\mbf{X} = \exp(\mbs{\xi}^\wedge)}$.

\subsection{The white-noise-on-acceleration motion prior}
\label{apx:wnoaprior}

The white-noise-on-acceleration (WNOA) motion prior, modified slightly from
\cite{Anderson2015}, may be summarized by the following set of nonlinear
stochastic differential equations (SDEs), 
\begin{subequations}
	\begin{align}
		\mbfdot{T}(t) =& \ \mbf{T}(t) \mbs{\varpi}_b(t)^\wedge, 
		\label{eqn:se3kin} \\
		\mbsdot{\varpi}_b(t) =& \ \mbf{w}_b(t), \\
		\mbf{w}_b(t) \sim& \ \mathcal{GP}(\mbf{0}, \mbc{Q} \delta (t - t^\prime)),
	\end{align}
	\label{eqn:sdes}%
\end{subequations}
where the time argument is included to emphasize that \eqref{eqn:sdes} evolves
in continuous time, and the subscript $(\cdot)_b$ is included to emphasize that
the generalized velocity $\mbs{\varpi}_b$ is a body-frame quantity.  The WNOA
prior promotes constant body-centric velocity (smoothing) throughout in the
trajectory.  The navigation state is defined as the ordered pair 
\begin{equation}
	\mbf{X} = \left( \mbf{T}, \mbs{\varpi} \right) \in SE(3) \times \rnums^6,
\end{equation}
with ${\mbf{T} \in SE(3)}$ and ${T \mbs{\varpi}^\wedge \in \mathfrak{g}}$, where
$T$ is a time increment.  Following a left-invariant perturbation scheme for the
pose, the navigation state is perturbed as 
\begin{subequations}
	\begin{align}
		\mbf{T} =& \ \mbfbar{T} \exp(-\delta \mbs{\xi}^\wedge), 
		\label{eqn:perturbationschemepose} \\
		\mbs{\varpi} =& \ \mbsbar{\varpi} + \delta \mbs{\varpi}.
		\label{eqn:perturbationschemevarpi}
	\end{align}
	\label{eqn:perturbationscheme}
\end{subequations}
\Cref{eqn:sdes} may be divided into a set of deterministic mean equations, 
\begin{subequations}
	\begin{align}
		\dot{\mbfbar{T}} =& \ \mbfbar{T} \mbsbar{\varpi}^\wedge, \\
		\dot{\mbsbar{\varpi}} =& \ \mbf{0},
	\end{align}
\end{subequations}
and a separate SDE describing the perturbations \cite{Anderson2015}, 
\begin{equation}
	\begin{bmatrix}
		\delta \mbsdot{\xi}(t) \\ \delta \mbsdot{\varpi}(t)
	\end{bmatrix} =
	\mbf{A}
	\begin{bmatrix}
		\delta \mbs{\xi}(t) \\ \delta \mbs{\varpi}(t)
	\end{bmatrix} + \mbf{L} \, \delta \mbf{w}(t),
	\label{eqn:wnoaerrorkingeneral}
\end{equation}
with $\mbf{L} = \begin{bmatrix}	\mbf{0} & \eye \end{bmatrix}^\trans$, and where 
\begin{equation}
	\delta \mbf{w}(t) \sim \mathcal{GP}(\mbf{0}, \mbc{Q} \delta(t-t^\prime)).
\end{equation}
To formulate a batch estimation problem, the continuous-time error kinematics
$\mbf{A}$ must be derived and discretized.

\subsection{Deriving the WNOA state error kinematics on $SE(3) \times \rnums^6$}
\label{apx:errorkin}

The WNOA state error kinematics are derived in this section.  The
continuous-time state error kinematics are first obtained by linearizing the
navigation state kinematics, and are then discretized exactly via the matrix
exponential. 

Following the perturbation scheme \eqref{eqn:perturbationscheme}, approximating
${\exp(-\delta \mbs{\xi}^\wedge) \approx (\eye - \delta \mbs{\xi}^\wedge)}$, and
ignoring higher-order terms, the continuous-time pose kinematics
\eqref{eqn:se3kin} are perturbed as
\begin{align*}
	\mbfdot{T} =& \ \mbf{T} \mbs{\varpi}^\wedge, \\
	\frac{\dee}{\dee t} \big(\mbfbar{T} \exp(-\delta \mbs{\xi}^\wedge) \big) =& \ \mbfbar{T} \exp(-\delta \mbs{\xi}^\wedge) (\mbsbar{\varpi} + \delta \mbs{\varpi})^\wedge, \\
	\dot{\mbfbar{T}} - \dot{\mbfbar{T}} \delta \mbs{\xi}^\wedge - \mbfbar{T} \delta \mbsdot{\xi}^\wedge \approx& \ \mbfbar{T} \mbsbar{\varpi}^\wedge + \mbfbar{T} \delta \mbs{\varpi}^\wedge - \mbfbar{T} \delta \mbs{\xi}^\wedge \mbsbar{\varpi}^\wedge, \\
	\mbfbar{T} \delta \mbsdot{\xi}^\wedge =& \ -\mbfbar{T} \delta \mbs{\varpi}^\wedge + \mbfbar{T} \delta \mbs{\xi}^\wedge \mbsbar{\varpi}^\wedge - \mbfbar{T} \mbsbar{\varpi}^\wedge \delta \mbs{\xi}^\wedge, \\
	\delta \mbsdot{\xi}^\wedge =& \ - \delta \mbs{\varpi}^\wedge + \delta \mbs{\xi}^\wedge \mbsbar{\varpi}^\wedge - \mbsbar{\varpi}^\wedge \delta \mbs{\xi}^\wedge, \\
	\delta \mbsdot{\xi} =& \ -\adj(\mbsbar{\varpi}^\wedge) \delta \mbs{\xi} - \delta \mbs{\varpi}. 
	\label{eqn:ctperturbleft} \numberthis
\end{align*}
\Cref{eqn:ctperturbleft} describes the continuous-time pose error kinematics.
Inserting \eqref{eqn:ctperturbleft} into \eqref{eqn:wnoaerrorkingeneral} yields
\begin{equation}
	\underbrace{\begin{bmatrix}
		\delta \mbsdot{\xi}(t) \\ \delta \mbsdot{\varpi}(t)
	\end{bmatrix}}_{\delta \mbfdot{x}(t)} =
	\underbrace{\begin{bmatrix}
		-\adj(\mbsbar{\varpi}^\wedge) & -\eye \\
		\mbf{0} & \mbf{0}
	\end{bmatrix}}_{\mbf{A}(t)}
	\underbrace{\begin{bmatrix}
		\delta \mbs{\xi}(t) \\ \delta \mbs{\varpi}(t)
	\end{bmatrix}}_{\delta \mbf{x}(t)} + 
	\underbrace{\begin{bmatrix}
		\mbf{0} \\ \eye
	\end{bmatrix}}_{\mbf{L}(t)} \delta \mbf{w}(t),
	\label{eqn:wnoacterrors}
\end{equation}
which describes the continuous-time state error kinematics.  

The continuous-time state error kinematics will now be discretized, to provide a
check on the solution when deriving the discrete-time batch Jacobians in
\Cref{apx:jacodom}.  The matrix $\mbf{A}(t)$ is discretized exactly via the
matrix exponential \cite[Sec.~3.5.4]{Farrell2008}.  Considering ${\mbf{A}_{k-1}
= \exp(T \mbf{A})}$, where ${T = t_k - t_{k-1}}$, the first few powers of
$\mbf{A}^n$ are
\begin{subequations}
	\begin{align}
		\mbf{A}^2 =& \begin{bmatrix}
			\adj(\mbsbar{\varpi}^\wedge)^2 & \adj(\mbsbar{\varpi}^\wedge) \\ \mbf{0} & \mbf{0}
		\end{bmatrix}, \\
		\mbf{A}^3 =& \begin{bmatrix}
			-\adj(\mbsbar{\varpi}^\wedge)^3 & -\adj(\mbsbar{\varpi}^\wedge)^2 \\ \mbf{0} & \mbf{0}
		\end{bmatrix}.
	\end{align}
\end{subequations}
The matrix $\mbf{A}(t)$ is unfortunately not nilpotent, but may be written in
closed form by noting 
\begin{subequations}
	\begin{align}
		-\adj(\mbs{\xi}^\wedge) =& \ \adj(-\mbs{\xi}^\wedge), \\
		\exp(\adj(\mbs{\xi}^\wedge)) =& \ \Adj(\exp(\mbs{\xi}^\wedge)).
	\end{align}
\end{subequations}
Writing out the first few terms of ${\mbf{A}_{k-1} = \exp(T\mbf{A})}$
component-wise,
\begin{align*}
	\exp(T\mbf{A}) =& \begin{bmatrix}
		\mbf{A}^{11}_{k-1} & \mbf{A}^{12}_{k-1} \\ \mbf{0} & \eye
	\end{bmatrix},\\
	\mbf{A}^{11}_{k-1} =& \ \eye + \adj(-T\mbsbar{\varpi}^\wedge) + \frac{1}{2} \adj(-T\mbsbar{\varpi}^\wedge)^2 + \frac{1}{6} \adj(-T\mbsbar{\varpi}^\wedge)^3 + \cdots, \\
	\mbf{A}^{12}_{k-1} =& \ -T\eye - \frac{T}{2} \adj(-T \mbsbar{\varpi}^\wedge) - \frac{T}{6} \adj(-T \mbsbar{\varpi}^\wedge)^2 + \cdots, \\
	\exp(T\mbf{A}) =& \begin{bmatrix}
		\sum^{\infty}_{n = 0} \frac{1}{n!} \adj(-T \mbsbar{\varpi}^\wedge)^n & -T \sum^\infty_{n=0} \frac{1}{(1+n)!} \adj(-T \mbsbar{\varpi}^\wedge)^n \\
		\mbf{0} & \eye
	\end{bmatrix}, \\
	\mbf{A}_{k-1} =& \begin{bmatrix}
		\Adj(\exp(-T \mbsbar{\varpi}_{k-1}^\wedge)) & -T \mbf{J}^{\textrm{r}}(T \mbsbar{\varpi}_{k-1}) \\
		\mbf{0} & \eye
	\end{bmatrix}.  \numberthis \label{eqn:wnoaperturbleftdtjac}
\end{align*}
Matrix $\mbf{A}_{k-1}$ describes the discrete-time state error kinematics for
the WNOA motion prior.

\subsection{Deriving the batch Jacobians}
\label{apx:jacobians}

The prior, process, and loop closure Jacobians are derived in this section.
Each derivation starts with the respective discrete-time error definitions, and
uses the identities given throughout \Cref{apx:preliminaries} to arrive at the
final result.  Note throughout that $\bar{(\cdot)}$ is used to denote a mean
state estimate, and $\tilde{(\cdot)}$ is used to denote a state estimate
generated from sensor measurements or prior information.

\subsubsection{Deriving the Jacobians on the prior error}
\label{apx:jacprior}

With a left-invariant pose error definition, the prior navigation state error is
\begin{equation}
	\mbf{e}_0 = \begin{bmatrix}
		\mbf{e}^\xi_0 \\ \mbf{e}^\varpi_0
	\end{bmatrix} = \begin{bmatrix}
		\log \left( \mbf{T}_0\inv \mbf{Y}_0 \right)^\vee \\
		\mbs{\varpi}_0 - \mbs{\psi}_0
	\end{bmatrix},
	\label{eqn:priorerrorright}
\end{equation}
where ${ ( \mbf{Y}_0, \mbs{\psi}_0 ) }$ is the prior estimate on the first
navigation state.  \secondupdate{The objective is to perturb
\eqref{eqn:priorerrorright} to first order with respect to the design variables
$\mbf{T}_0$ and $\mbs{\varpi}_0$ to recover the Jacobian matrices.  In the case
of the prior pose error, the BCH identities \eqref{eqn:bchapprox} given in
\Cref{apx:bch} will be used to manipulate the resulting expression into a form
that matches the left-invariant error definition introduced in
\Cref{tab:errordefinitions}.} With ${\delta \mbf{T}_0 =
\exp((\mbf{e}^\xi_0)^\wedge)}$, ${\mbftilde{T}_0 = \mbf{Y}_0}$, and following
the \secondupdate{left-invariant} perturbation scheme
\eqref{eqn:perturbationschemepose}, the prior pose error is linearized as 
\begin{subequations}
\begin{align*}
	\delta \mbf{T}_0 =& \ \mbf{T}_0\inv \mbftilde{T}_0 \\
	=& \ \mbf{T}_0\inv \mbf{Y}_0 \\
	=& \ \exp(\delta \mbs{\xi}^\wedge_0) \mbfbar{T}\inv_0 \mbfbar{Y}_0 \exp(-\delta \mbs{\eta}^\wedge_0) \\
	=& \ \exp(\delta \mbs{\xi}_0^\wedge) \delta \mbfbar{T}_0 \exp(-\delta \mbs{\eta}^\wedge_0) \\
	=& \ \delta \mbfbar{T}_0 \, \delta \mbfbar{T}_0\inv \exp(\delta \mbs{\xi}_0^\wedge) \delta \mbfbar{T}_0 \exp(-\delta \mbs{\eta}^\wedge_0) \\
	=& \ \delta \mbfbar{T}_0 \exp( ( \Adj(\delta \mbfbar{T}_0\inv) \delta \mbs{\xi}_0)^\wedge) \exp(-\delta \mbs{\eta}^\wedge_0).
	\numberthis
	\label{eqn:priortemp_a}
\end{align*}
\secondupdate{Note that the mean prior pose error $\delta \mbfbar{T}_0$ in
\eqref{eqn:priortemp_a} has been relocated to the far left through use of the
adjoint matrix \eqref{eqn:adjmatrix}, matching the form of the left-invariant
error.  However, \eqref{eqn:priortemp_a} contains two $\exp(\cdot)$ terms, which
must be combined to match the left-invariant error definition.} \update{Using
BCH identity \eqref{eqn:bchapprox3} to combine the perturbations in
\eqref{eqn:priortemp_a} and continuing,}
\begin{align*}
	\delta \mbf{T}_0 \approx& \ \delta \mbfbar{T}_0 \exp( ( \Adj(\delta \mbfbar{T}_0\inv) \delta \mbs{\xi}_0 - \delta \mbs{\eta}_0)^\wedge), \\
	\exp((\mbf{e}^\xi_0)^\wedge) =& \ \exp((\mbfbar{e}^\xi_0)^\wedge) \exp( ( \Adj(\delta \mbfbar{T}_0\inv) \delta \mbs{\xi}_0 - \delta \mbs{\eta}_0)^\wedge). 
	\numberthis
	\label{eqn:priortemp1}
\end{align*}
\secondupdate{Finally, to obtain a linear expression, use} \update{BCH identity
\eqref{eqn:bchapprox1} to combine all terms on the matrix Lie algebra in
\eqref{eqn:priortemp1},}
\begin{align*}
	\exp((\mbf{e}^\xi_0)^\wedge) \approx& \ \exp( ( \mbfbar{e}^\xi_0 + \mbf{J}^\textrm{r}(\mbfbar{e}^\xi_0)\inv ( \Adj(\delta \mbfbar{T}_0\inv) \delta \mbs{\xi}_0 - \delta \mbs{\eta}_0) )^\wedge), \\
	\mbf{e}^\xi_0 =& \ \mbfbar{e}^\xi_0 + \mbf{J}^\textrm{r}(\mbfbar{e}^\xi_0)\inv ( \Adj(\delta \mbfbar{T}_0\inv) \delta \mbs{\xi}_0 - \delta \mbs{\eta}_0).
	\numberthis
	\label{eqn:priorint1}
\end{align*}
\end{subequations}
To simplify \secondupdate{the Jacobian associated with prior pose perturbation
$\delta \mbs{\xi}_0$ in} \eqref{eqn:priorint1}, identity
\eqref{eqn:adjjacidentity} is used to produce
\begin{align*}
	\mbf{J}^\textrm{r}(\mbs{\xi})\inv \Adj(\mbf{X}\inv) =& \ \mbf{J}^\textrm{r}(\mbs{\xi})\inv \Adj((\exp(\mbs{\xi}^\wedge))\inv) \\
	=& \ \mbf{J}^\textrm{r}(\mbs{\xi})\inv \Adj(\exp(-\mbs{\xi}^\wedge)) \\
	=& \ \mbf{J}^\ell(-\mbs{\xi})\inv \Adj(\exp(-\mbs{\xi}^\wedge)) \\
	=& \ \mbf{J}^\textrm{r}(-\mbs{\xi})\inv \\
	=& \ \mbf{J}^\ell(\mbs{\xi})\inv.
	\numberthis
	\label{eqn:priorint2}
\end{align*}
Applying this to \eqref{eqn:priorint1}, the linearized prior pose error becomes
\begin{equation}
	\mbf{e}^\xi_0 = \mbfbar{e}^\xi_0 + \mbf{J}^\ell(\mbfbar{e}^\xi_0)\inv \delta \mbs{\xi}_0 - \mbf{J}^{\textrm{r}}(\mbfbar{e}^\xi_0)\inv \delta \mbs{\eta}_0.
	\label{eqn:priorposeerrorlin}
\end{equation}
The prior generalized velocity error is linearized as
\begin{align*}
	\mbf{e}^\varpi_0 =& \ (\mbsbar{\varpi}_0 + \delta \mbs{\varpi}_0) - (\mbsbar{\psi}_0 + \delta \mbs{\psi}_0) \\
	=& \ \mbfbar{e}^\varpi_0 + \delta \mbs{\varpi}_0 - \delta \mbs{\psi}_0.
	\numberthis
	\label{eqn:priorvelocityerrorlin}
\end{align*}
Combining \eqref{eqn:priorposeerrorlin} and \eqref{eqn:priorvelocityerrorlin}
yields the prior Jacobian,
\begin{equation}
	\underbrace{\begin{bmatrix} \delta \mbf{e}^\xi_0 \\ \delta \mbf{e}^\varpi_0 \end{bmatrix}}_{\delta \mbf{e}_0}
	= \underbrace{\begin{bmatrix}
		\mbf{J}^\ell(\mbfbar{e}^\xi_0)\inv & \mbf{0} \\ 
		\mbf{0} & \eye
	\end{bmatrix}}_{\mbf{F}_0^0} \underbrace{\begin{bmatrix}
		\delta \mbs{\xi}_0 \\ \delta \mbs{\varpi}_0
	\end{bmatrix}}_{\delta \mbf{x}_0} + \underbrace{\begin{bmatrix} 
		- \mbf{J}^{\textrm{r}}(\mbfbar{e}^\xi_0)\inv & \mbf{0} \\
		\mbf{0} & -\eye
	\end{bmatrix}}_{\mbf{M}_0} \underbrace{\begin{bmatrix}
		\delta \mbs{\eta}_0 \\ \delta \mbs{\psi}_0
	\end{bmatrix}}_{\delta \mbf{y}_0}.
\end{equation}

\subsubsection{Deriving the Jacobians on the WNOA error}
\label{apx:jacodom}

With a left-invariant pose error definition, the WNOA process (constant body
velocity) navigation state errors are
\begin{equation}
	\mbf{e}_k = \begin{bmatrix}
		\mbf{e}^\xi_k \\ \mbf{e}^\varpi_k
	\end{bmatrix} =
	 \begin{bmatrix}
		\log \left( \mbf{T}\inv_k \mbftilde{T}_k \right)^\vee \\
		\mbs{\varpi}_k - \mbs{\varpi}_{k-1}
	\end{bmatrix}. 
	\label{eqn:wnoaerror}
\end{equation}
\secondupdate{The objective is to linearize \eqref{eqn:wnoaerror}, using the BCH
identities \eqref{eqn:bchapprox} to produce an expression that looks like a
left-invariant error.}  With ${\delta \mbf{T}_k = \exp((\mbf{e}^\xi_k)^\wedge)}$
and ${\mbftilde{T}_k = \mbf{T}_{k-1} \exp(T\mbs{\varpi}^\wedge_{k-1})}$, ${T =
t_k - t_{k-1}}$, the WNOA pose error is linearized as
\begin{subequations}
\begin{align*}
	\delta \mbf{T}_k =& \ \mbf{T}\inv_k \mbftilde{T}_k \\
	=& \ \mbf{T}\inv_k \mbf{T}_{k-1} \exp(T \mbs{\varpi}^\wedge_{k-1}) \\
	=& \ \exp(\delta \mbs{\xi}^\wedge_k) \mbfbar{T}\inv_k \mbfbar{T}_{k-1} \exp(-\delta \mbs{\xi}^\wedge_{k-1}) \exp(T (\mbsbar{\varpi}_{k-1} + \delta \mbs{\varpi}_{k-1})^\wedge).
	\numberthis
	\label{eqn:posetemp1}
\end{align*}
\update{Using BCH identity \eqref{eqn:bchapprox2} to separate the terms in the
last exponential and continuing from \eqref{eqn:posetemp1},}
\begin{align*}
	\delta \mbf{T}_k \approx& \ \exp(\delta \mbs{\xi}^\wedge_k) \mbfbar{T}\inv_k \mbfbar{T}_{k-1} \exp(-\delta \mbs{\xi}^\wedge_{k-1}) \exp(T \mbsbar{\varpi}_{k-1}^\wedge) \exp( T (\mbf{J}^\textrm{r}(T \mbsbar{\varpi}_{k-1}) \delta \mbs{\varpi}_{k-1})^\wedge) \\
	=& \ \exp(\delta \mbs{\xi}^\wedge_k) \mbfbar{T}\inv_k \mbfbar{T}_{k-1}  \exp(T \mbsbar{\varpi}_{k-1}^\wedge) \exp(- T \mbsbar{\varpi}_{k-1}^\wedge) \exp(-\delta \mbs{\xi}^\wedge_{k-1}) \exp(T \mbsbar{\varpi}_{k-1}^\wedge) \\
		& \ \quad \times \exp( T (\mbf{J}^\textrm{r}(T \mbsbar{\varpi}_{k-1}) \delta \mbs{\varpi}_{k-1})^\wedge) \\
	=& \ \exp(\delta \mbs{\xi}^\wedge_k) \mbfbar{T}\inv_k \mbfbar{T}_{k-1}  \exp(T \mbsbar{\varpi}_{k-1}^\wedge) \exp(- ( \Adj( \exp(-T \mbsbar{\varpi}_{k-1}^\wedge)) \delta \mbs{\xi}_{k-1})^\wedge) \\
		& \ \quad \times \exp( T (\mbf{J}^\textrm{r}(T \mbsbar{\varpi}_{k-1}) \delta \mbs{\varpi}_{k-1})^\wedge) \\
	=& \ \exp(\delta \mbs{\xi}^\wedge_k) \delta \mbfbar{T}_k \exp(- ( \Adj( \exp(-T \mbsbar{\varpi}_{k-1}^\wedge)) \delta \mbs{\xi}_{k-1})^\wedge) \exp( T (\mbf{J}^\textrm{r}(T \mbsbar{\varpi}_{k-1}) \delta \mbs{\varpi}_{k-1})^\wedge) \\
	=& \ \delta \mbfbar{T}_k \delta \mbfbar{T}_k\inv \exp(\delta \mbs{\xi}^\wedge_k) \delta \mbfbar{T} \exp(- ( \Adj( \exp(-T \mbsbar{\varpi}_{k-1}^\wedge)) \delta \mbs{\xi}_{k-1})^\wedge) \exp( T (\mbf{J}^\textrm{r}(T \mbsbar{\varpi}_{k-1}) \delta \mbs{\varpi}_{k-1})^\wedge) \\
	=& \ \delta \mbfbar{T}_k \exp( ( \Adj(\delta \mbfbar{T}_k\inv) \delta \mbs{\xi}_k)^\wedge) \exp(- ( \Adj( \exp(-T \mbsbar{\varpi}_{k-1}^\wedge)) \delta \mbs{\xi}_{k-1})^\wedge) \exp( T (\mbf{J}^\textrm{r}(T \mbsbar{\varpi}_{k-1}) \delta \mbs{\varpi}_{k-1})^\wedge), \\
	\begin{split}
		\exp((\mbf{e}^\xi_k)^\wedge) =& \ \exp((\mbfbar{e}^\xi_k)^\wedge) \exp( ( \Adj(\delta \mbfbar{T}_k\inv) \delta \mbs{\xi}_k)^\wedge) \exp(- ( \Adj( \exp(-T \mbsbar{\varpi}_{k-1}^\wedge)) \delta \mbs{\xi}_{k-1})^\wedge) \\ 
		& \ \quad \times \exp( T (\mbf{J}^\textrm{r}(T \mbsbar{\varpi}_{k-1}) \delta \mbs{\varpi}_{k-1})^\wedge).
	\end{split}
	\numberthis
	\label{eqn:posetemp3}
\end{align*}
\update{Using BCH identity \eqref{eqn:bchapprox3} to combine all perturbation terms in
\eqref{eqn:posetemp3} and continuing,}
\begin{equation}
	\exp((\mbf{e}^\xi_k)^\wedge) \approx \exp((\mbfbar{e}^\xi_k)^\wedge) \exp( ( \Adj(\delta \mbfbar{T}_k\inv) \delta \mbs{\xi}_k - \Adj( \exp(-T \mbsbar{\varpi}_{k-1}^\wedge)) \delta \mbs{\xi}_{k-1} + T \mbf{J}^\textrm{r}(T \mbsbar{\varpi}_{k-1}) \delta \mbs{\varpi}_{k-1} )^\wedge ).
	\label{eqn:posetemp4}
\end{equation}
\secondupdate{\Cref{eqn:posetemp4} now resembles a left-invariant error, as
required.}  \update{Combining all terms on the matrix Lie algebra via BCH
identity \eqref{eqn:bchapprox1} \secondupdate{in order to produce a linear
expression,} and continuing from \eqref{eqn:posetemp4},}
\begin{align*}
	\exp((\mbf{e}^\xi_k)^\wedge) \approx& \ \exp( ( \mbfbar{e}^\xi_k + \mbf{J}^\textrm{r}(\mbfbar{e}^\xi_k)\inv ( \Adj(\delta \mbfbar{T}_k\inv) \delta \mbs{\xi}_k - \Adj( \exp(-T \mbsbar{\varpi}_{k-1}^\wedge)) \delta \mbs{\xi}_{k-1} \! + \! T \mbf{J}^\textrm{r}(T \mbsbar{\varpi}_{k-1}) \delta \mbs{\varpi}_{k-1} ) )^\wedge ), 
	\numberthis \label{eqn:posetemp5} \\
	\mbf{e}^\xi_k =& \ \mbfbar{e}^\xi_k + \mbf{J}^\textrm{r}(\mbfbar{e}^\xi_k)\inv ( \Adj(\delta \mbfbar{T}_k\inv) \delta \mbs{\xi}_k - \Adj( \exp(-T \mbsbar{\varpi}_{k-1}^\wedge)) \delta \mbs{\xi}_{k-1} + T \mbf{J}^\textrm{r}(T \mbsbar{\varpi}_{k-1}) \delta \mbs{\varpi}_{k-1} ) \\
	=& \ \mbfbar{e}^\xi_k - \mbf{J}^\textrm{r}(\mbfbar{e}^\xi_k)\inv \Adj( \exp(-T \mbsbar{\varpi}_{k-1}^\wedge)) \delta \mbs{\xi}_{k-1} + T \mbf{J}^\textrm{r}(\mbfbar{e}^\xi_k)\inv \mbf{J}^\textrm{r}(T \mbsbar{\varpi}_{k-1}) \delta \mbs{\varpi}_{k-1} + \mbf{J}^\ell(\mbfbar{e}^\xi_k)\inv \delta \mbs{\xi}_k. 
	\numberthis
	\label{eqn:linearizewnoapose}
\end{align*}
\end{subequations}
Note BCH identity \eqref{eqn:adjjacidentity} was used to simplify
\eqref{eqn:linearizewnoapose}, for details see \eqref{eqn:priorint2}.  The
generalized velocity error is linearized as 
\begin{align*}
	\mbf{e}^\varpi_k =& \ (\mbsbar{\varpi}_k + \delta \mbs{\varpi}_k) - (\mbsbar{\varpi}_{k-1} + \delta \mbs{\varpi}_{k-1}) \\
	=& \ \mbfbar{e}^\varpi_k + \delta \mbs{\varpi}_k - \delta \mbs{\varpi}_{k-1}.
	\numberthis 
	\label{eqn:linearizewnoavarpi}
\end{align*}
Collecting \eqref{eqn:linearizewnoapose} and \eqref{eqn:linearizewnoavarpi}, the
WNOA error Jacobians are given by 
\begin{equation}
	\underbrace{\begin{bmatrix}
		\delta \mbf{e}^\xi_k \\ \delta \mbf{e}^\varpi_k
	\end{bmatrix}}_{\delta \mbf{e}_k} = \underbrace{\begin{bmatrix}
		- \mbf{J}^\textrm{r}(\mbfbar{e}^\xi_k)\inv \Adj( \exp(-T \mbsbar{\varpi}_{k-1}^\wedge)) & T \mbf{J}^\textrm{r}(\mbfbar{e}^\xi_k)\inv \mbf{J}^\textrm{r}(T \mbsbar{\varpi}_{k-1}) \\ \mbf{0} & -\eye
	\end{bmatrix}}_{\mbf{F}^k_{k-1}} \underbrace{\begin{bmatrix}
		\delta \mbs{\xi}_{k-1} \\ \delta \mbs{\varpi}_{k-1}
	\end{bmatrix}}_{\delta \mbf{x}_{k-1}} + 
	\underbrace{\begin{bmatrix}
		\mbf{J}^\ell(\mbfbar{e}^\xi_k)\inv & \mbf{0} \\ \mbf{0} & \eye
	\end{bmatrix}}_{\mbf{F}^k_k} \underbrace{\begin{bmatrix}
		\delta \mbs{\xi}_k \\ \delta \mbs{\varpi}_k
	\end{bmatrix}}_{\delta \mbf{x}_k}. \label{eqn:wnoaerrorjacsleft}
\end{equation}
Note that $\mbf{F}^k_{k-1}$ is the negative of $\mbf{A}_{k-1}$ from
\eqref{eqn:wnoaperturbleftdtjac}, with the exception of the
$\mbf{J}^{\textrm{r}}(\mbfbar{e}^\xi_k)\inv$ terms in the top row owing to the
application of BCH identity \eqref{eqn:bchapprox1} when moving from
\eqref{eqn:posetemp4} to \eqref{eqn:posetemp5}.  The discrete-time state error
kinematics $\mbf{A}_{k-1}$, also known as the \textit{transition matrix}
\cite[Sec.~3.1.1]{Barfoot2017}, are expected to appear at this location of the
batch problem \cite[Sec.~3.1.2]{Barfoot2017}, and therefore
\eqref{eqn:wnoaperturbleftdtjac} provides a useful check on the solution.

\subsubsection{Deriving the Jacobians on the loop closure error}
\label{apx:jaclc}

Again using a left-invariant pose error definition, the loop closure error is 
\begin{subequations}
\begin{equation}
	\mbf{e}_\ell = \log \left( \mbf{T}_{\ell_2}\inv \mbftilde{T}_{\ell_2} \right)^\vee.
\end{equation}
With ${\delta \mbf{T}_\ell = \exp(\mbf{e}_\ell^\wedge)}$ and
${\mbftilde{T}_{\ell_2} = \mbf{T}_{\ell_1} \mbs{\Xi}_{\ell_1\ell_2} }$, the
loop close error is linearized as 
\begin{align*}
	\delta \mbf{T}_\ell =& \mbf{T}\inv_{\ell_2} \mbftilde{T}_{\ell_2} \\
	=& \ \mbf{T}\inv_{\ell_2} \mbf{T}_{\ell_1} \mbs{\Xi}_{\ell_1\ell_2} \\
	=& \ \exp(\delta \mbs{\xi}_{\ell_2}^\wedge) \mbfbar{T}_{\ell_2}^{-1} \mbfbar{T}_{\ell_1} \exp(-\delta \mbs{\xi}_{\ell_1}^\wedge) \mbsbar{\Xi}_{\ell_1\ell_2} \exp(-\delta \mbs{\xi}^\wedge_\Xi) \\
	=& \ \exp(\delta \mbs{\xi}_{\ell_2}^\wedge) \mbfbar{T}_{\ell_2}\inv \mbfbar{T}_{\ell_1} \mbsbar{\Xi}_{\ell_1\ell_2} \mbsbar{\Xi}_{\ell_1\ell_2}\inv \exp(-\delta \mbs{\xi}_{\ell_1}^\wedge) \mbsbar{\Xi}_{\ell_1\ell_2} \exp(-\delta \mbs{\xi}^\wedge_\Xi) \\
	=& \ \exp(\delta \mbs{\xi}_{\ell_2}^\wedge) \delta \mbfbar{T}_\ell \exp(-( \Adj(\mbsbar{\Xi}_{\ell_1\ell_2}\inv) \delta \mbs{\xi}_{\ell_1})^\wedge) \exp(-\delta \mbs{\xi}^\wedge_\Xi) \\
	=& \ \delta \mbfbar{T}_\ell \, \delta \mbfbar{T}_\ell\inv \exp(\delta \mbs{\xi}_{\ell_2}^\wedge) \delta \mbfbar{T}_\ell \exp(-( \Adj(\mbsbar{\Xi}_{\ell_1\ell_2}\inv) \delta \mbs{\xi}_{\ell_1})^\wedge) \exp(-\delta \mbs{\xi}^\wedge_\Xi) \\
	=& \ \delta \mbfbar{T}_\ell \exp( ( \Adj(\delta \mbfbar{T}_\ell\inv) \delta \mbs{\xi}_{\ell_2})^\wedge) \exp(-( \Adj(\mbsbar{\Xi}_{\ell_1\ell_2}\inv) \delta \mbs{\xi}_{\ell_1})^\wedge) \exp(-\delta \mbs{\xi}^\wedge_\Xi), \\
	\exp(\mbf{e}_\ell^\wedge) =& \ \exp(\mbfbar{e}_\ell^\wedge) \exp( ( \Adj(\delta \mbfbar{T}_\ell\inv) \delta \mbs{\xi}_{\ell_2})^\wedge) \exp(-( \Adj(\mbsbar{\Xi}_{\ell_1\ell_2}\inv) \delta \mbs{\xi}_{\ell_1})^\wedge) \exp(-\delta \mbs{\xi}^\wedge_\Xi)
	\numberthis \label{eqn:lctemp1} \\ 
	\approx& \ \exp(\mbfbar{e}_\ell^\wedge) \exp( ( \Adj(\delta \mbfbar{T}_\ell\inv) \delta \mbs{\xi}_{\ell_2} - \Adj(\mbsbar{\Xi}_{\ell_1\ell_2}\inv) \delta \mbs{\xi}_{\ell_1} -\delta \mbs{\xi}_\Xi )^\wedge).
	\numberthis \label{eqn:lctemp2}
\end{align*}
\update{BCH identity \eqref{eqn:bchapprox3} was applied to combine all
perturbation terms in \eqref{eqn:lctemp1}.  Continuing from \eqref{eqn:lctemp2}
and applying BCH identify \eqref{eqn:bchapprox1} to combine all terms on the
matrix Lie algebra, }
\begin{align*}
	\exp(\mbf{e}_\ell^\wedge) \approx& \ \exp( ( \mbfbar{e}_\ell + \mbf{J}^\textrm{r}(\mbfbar{e}_\ell)\inv ( \Adj( \delta \mbfbar{T}_\ell\inv) \delta \mbs{\xi}_{\ell_2} - \Adj(\mbsbar{\Xi}_{\ell_1\ell_2}\inv) \delta \mbs{\xi}_{\ell_1} -\delta \mbs{\xi}_\Xi ) )^\wedge), \\
	\mbf{e}_\ell =& \ \mbfbar{e}_\ell + \mbf{J}^\textrm{r}(\mbfbar{e}_\ell)\inv ( \Adj( \delta \mbfbar{T}_\ell \inv) \delta \mbs{\xi}_{\ell_2} - \Adj(\mbsbar{\Xi}_{\ell_1\ell_2}\inv) \delta \mbs{\xi}_{\ell_1} -\delta \mbs{\xi}_\Xi ) \\
	=& \ \mbfbar{e}_\ell - \mbf{J}^\textrm{r}(\mbfbar{e}_\ell)\inv \Adj(\mbsbar{\Xi}_{\ell_1\ell_2}\inv) \delta \mbs{\xi}_{\ell_1} + \mbf{J}^\ell(\mbfbar{e}_\ell)\inv \delta \mbs{\xi}_{\ell_2} - \mbf{J}^\textrm{r}(\mbfbar{e}_\ell)\inv \delta \mbs{\xi}_\Xi.
	\numberthis
	\label{eqn:lcfinal}
\end{align*}	
\end{subequations}
Again, identity \eqref{eqn:adjjacidentity} was used to simplify
\eqref{eqn:lcfinal}, for details see \eqref{eqn:priorint2}.  The loop closure
Jacobians are therefore
\begin{equation}
	\delta \mbf{e}_\ell = \underbrace{- \mbf{J}^\textrm{r}(\mbfbar{e}_\ell)\inv \Adj(\mbsbar{\Xi}_{\ell_1\ell_2}\inv)}_{\mbf{H}^\ell_{\ell_1}} \delta \mbs{\xi}_{\ell_1} + \underbrace{\mbf{J}^\ell(\mbfbar{e}_\ell)\inv}_{\mbf{H}^\ell_{\ell_2}} \delta \mbs{\xi}_{\ell_2} \underbrace{- \mbf{J}^\textrm{r}(\mbfbar{e}_\ell)\inv}_{\mbf{M}_\ell} \delta \mbs{\xi}_{\Xi}.
\end{equation}

\newpage
\printbibliography[heading=subbibliography,title=Appendix \thesection~References]
\end{refsection}

\section{Additional alignment images}
\label{sec:additionalimages}

\renewcommand{\thefigure}{\thesection.\arabic{figure}}
\setcounter{figure}{0}

\begin{figure*}[htb]
	\sbox\subfigbox{%
	  \resizebox{\dimexpr0.96\textwidth-1em}{!}{%
	    \includegraphics[height=4cm]{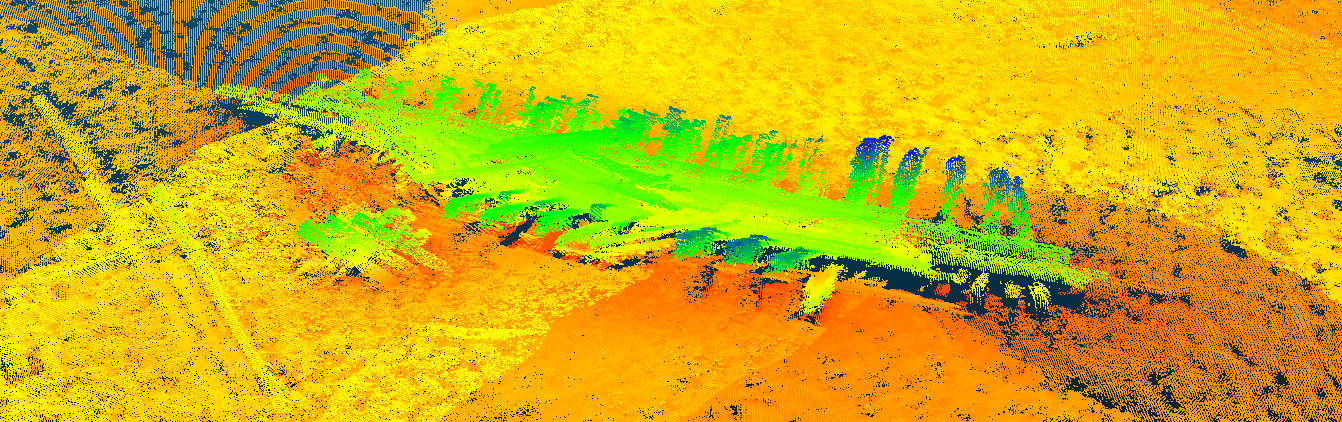}%
		\includegraphics[height=4cm]{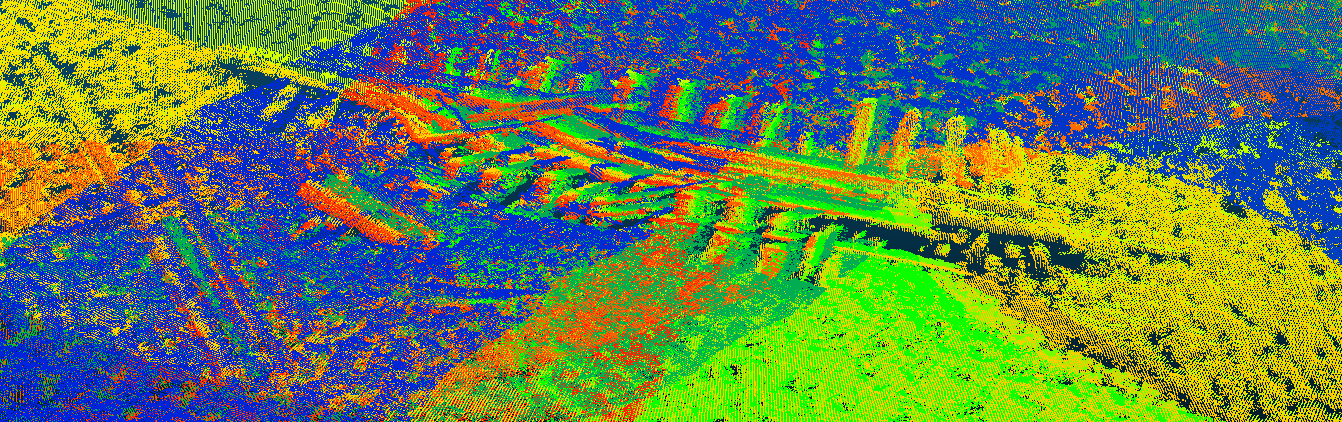}%
	  }%
	}
	\setlength{\subfigheight}{\ht\subfigbox}
	\centering
	\subcaptionbox{Prior elevation map (INS) \label{fig:cc_elevation_prior}}{%
	    \includegraphics[height=\subfigheight]{figs/cc_prior_elevation_small.png}
	}
	\hspace{3pt}
    \subcaptionbox{Prior time stamp map (INS)
	\label{fig:cc_time_prior}}{%
		\includegraphics[height=\subfigheight]{figs/cc_prior_time_small.png}
	}
    \par\medskip
	\subcaptionbox{Posterior elevation map (INS+LC)
	\label{fig:cc_elevation_posterior}}{%
		\includegraphics[height=\subfigheight]{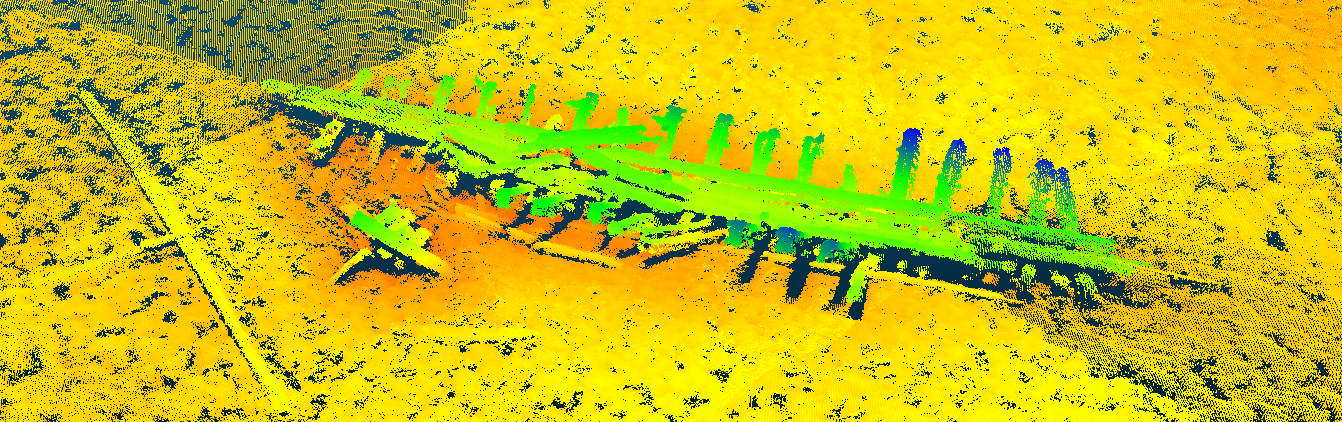}
	}
	\hspace{3pt}
  	\subcaptionbox{Posterior time stamp map (INS+LC)
  	\label{fig:cc_time_posterior}}{%
		\includegraphics[height=\subfigheight]{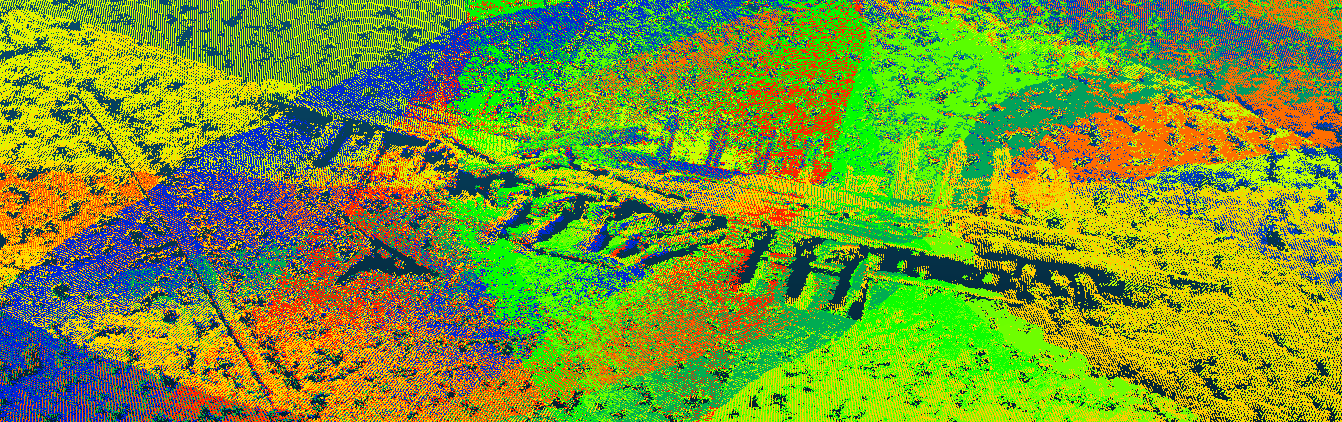}
	} 
    \caption{Images of the Wiarton shipwreck area, comparing the prior INS point
    cloud map in the top row to the posterior INS+LC point cloud map in the
    bottom row.  In the right column, colour denotes relative trajectory time,
    from low (blue) to high (red).  The time stamp plots are included to
    highlight the many passes involved in generating the final point cloud map.}
	\label{fig:shipwreck_passes}
\end{figure*}

%
\vspace{12pt}
%
\begin{figure*}[htb]
	\sbox\subfigbox{%
	  \resizebox{\dimexpr0.96\textwidth-1em}{!}{%
	    \includegraphics[height=4cm]{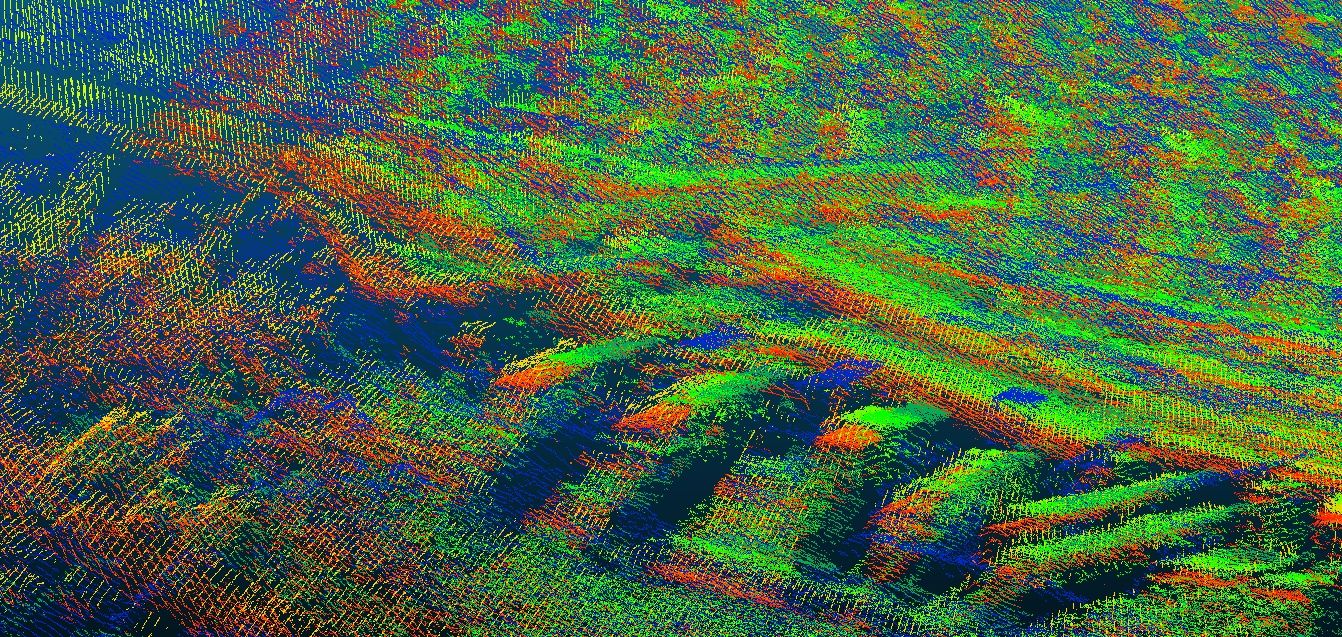}%
		\includegraphics[height=4cm]{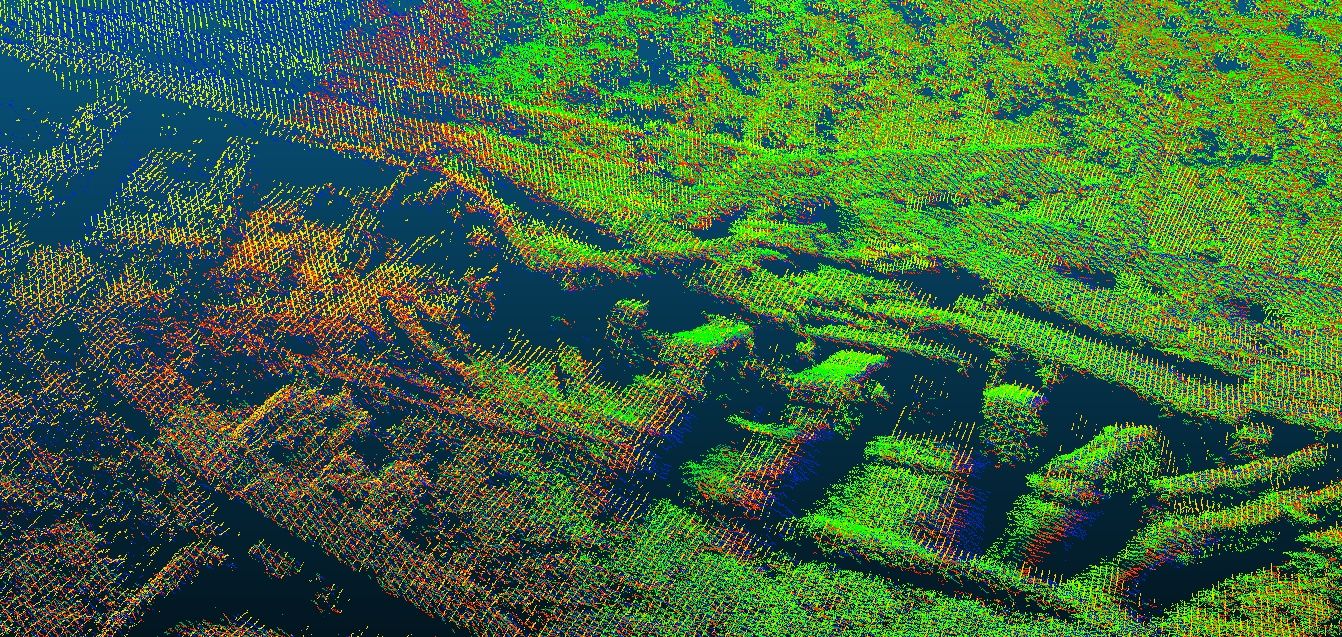}%
	  }%
	}
	\setlength{\subfigheight}{\ht\subfigbox}
	\centering
	\subcaptionbox{Prior time stamp map (INS), zoom
	\label{fig:cc_time_prior_zoom}}{%
	    \includegraphics[height=\subfigheight]{figs/cc_prior_time_zoom_crop_small.jpg}
	}
	\hspace{3pt}
    \subcaptionbox{Posterior time stamp map (INS+LC), zoom
	\label{fig:cc_time_posterior_zoom}}{%
		\includegraphics[height=\subfigheight]{figs/cc_posterior_time_zoom_crop_small.jpg}
	}
    \caption{A zoom of the shipwreck area, corresponding to the images in
    \Cref{fig:summary}.  The INS+LC solution on the right delivers a markedly
    more crisp, self-consistent point cloud map, suitable for inspection and
    metrology work.}
	\label{fig:shipwreck_passes_zoom}
\end{figure*}


\end{document}